\documentclass[acmsmall,table,xcdraw]{acmart}

\newlength{\bibitemsep}
\newlength{\bibparskip}
\let\oldthebibliography\thebibliography
\renewcommand\thebibliography[1]{%
  \oldthebibliography{#1}%
  \setlength{\parskip}{\bibitemsep}%
  \setlength{\itemsep}{\bibparskip}%
}
\usepackage{amsmath,amsfonts}
\usepackage{algorithmic}
\usepackage{graphicx}
\usepackage{textcomp}
\usepackage{xcolor}
\usepackage{enumitem}
\usepackage{verbatim}
\usepackage{booktabs}
\usepackage{multirow}
\usepackage{footmisc}
\usepackage{hyperref}
\usepackage{setspace}
\hypersetup{colorlinks=true,linkcolor=black,anchorcolor=black,citecolor=black,filecolor=black,menucolor=black,runcolor=black,urlcolor=black}

\usepackage{array}
\usepackage{color}
\usepackage{colortbl}
\usepackage{hhline}
\usepackage{makecell}
\usepackage[all=normal,floats=tight,paragraphs=tight]{savetrees}

\setcopyright{acmcopyright}
\copyrightyear{2021}
\acmYear{2021}
\acmDOI{XX.XXXX/XXXXXXX.XXXXXXX}

\acmJournal{CSUR}
\begin{document}


\title{The Threat of Offensive AI to Organizations}

	\author{Yisroel Mirsky}\authornote{Corresponding Author}
	\email{yisroel@post.bgu.ac.il}
	\affiliation{%
	\institution{Ben-Gurion University}
	\streetaddress{P.O.B. 653}
	\city{Beer-Sheva}
	\country{Israel}
	\postcode{8410501}
	}
	\author{Ambra Demontis}
	\email{ambrademontis@gmail.com}
	\affiliation{%
		\institution{University of Cagliari}
		\streetaddress{Via Università, 40}
		\city{Cagliari}
		\country{Italy}
		\postcode{09124}
	}
		\author{Jaidip Kotak}
	\email{jaidip@post.bgu.ac.il}
	\affiliation{%
	\institution{Ben-Gurion University}
	\streetaddress{P.O.B. 653}
	\city{Beer-Sheva}
	\country{Israel}
	\postcode{8410501}
	}
		\author{Ram Shankar}
	\email{ramk@microsoft.com}
	\affiliation{%
		\institution{Microsoft}
		\streetaddress{1 Microsoft Way}
		\city{Redmond}
		\state{Washington}
		\country{USA}
		\postcode{98052}
	}
			\author{Deng Gelei}
	\email{gelei.deng@ntu.edu.sg}
	\affiliation{%
		\institution{Nanyang Technological University}
		\streetaddress{50 Nanyang Ave}
		\country{Singapore}
		\postcode{639798}
	}
			\author{Liu Yang}
	\email{yangliu@ntu.edu.sg}
	\affiliation{%
		\institution{Nanyang Technological University}
		\streetaddress{50 Nanyang Ave}
		\country{Singapore}
		\postcode{639798}
	}
			\author{Xiangyu Zhang }
	\email{xyzhang@cs.purdue.edu}
	\affiliation{%
		\institution{Purdue}
		\streetaddress{610 Purdue Mall}
		\city{West Lafayette}
		\state{Indiana}
		\country{USA}
		\postcode{47907}
	}
			\author{Wenke Lee}
	\email{wenke@cc.gatech.edu}
	\affiliation{%
		\institution{Georgia Institute of Technology}
		\streetaddress{756 W Peachtree St NW}
		\city{Atlanta}
		\state{Georgia}
		\country{USA}
		\postcode{30308}
	}
			\author{Yuval Elovici }
	\email{elovici@bgu.ac.il}
	\affiliation{%
	\institution{Ben-Gurion University}
	\streetaddress{P.O.B. 653}
	\city{Beer-Sheva}
	\country{Israel}
	\postcode{8410501}
	}
	            
	\author{Battista Biggio}
	\email{battista.biggio@diee.unica.it}
	\affiliation{%
		\institution{University of Cagliari}
		\streetaddress{Via Università, 40}
		\city{Cagliari}
		\country{Italy}
		\postcode{09124}
	}
		\affiliation{%
		\institution{Pluribus One}
		\streetaddress{Via Bellini 9}
		\city{Cagliari}
		\country{Italy}
		\postcode{09128}
	}

	\renewcommand{\shortauthors}{Mirsky, et al.}
	

\begin{abstract}
AI has provided us with the ability to automate tasks, extract information from vast amounts of data, and synthesize media that is nearly indistinguishable from the real thing. However, positive tools can also be used for negative purposes. In particular, cyber adversaries can use AI to enhance their attacks and expand their campaigns. 

Although offensive AI has been discussed in the past, there is a need to analyze and understand the threat in the context of organizations. For example, how does an AI-capable adversary impact the cyber kill chain?  Does AI benefit the attacker more than the defender? What are the most significant AI threats facing organizations today and what will be their impact on the future?

In this survey, we explore the threat of offensive AI on organizations. First, we present the background and discuss how AI changes the adversary's methods, strategies, goals, and overall attack model. Then, through a literature review, we identify 33 offensive AI capabilities which adversaries can use to enhance their attacks. Finally, through a user study spanning industry and academia, we rank the AI threats and provide insights on the adversaries.
\end{abstract}



\maketitle

\section{Introduction}
For decades, organizations, including government agencies, hospitals, and financial institutions, have been the target of cyber attacks \cite{Knake17,mattei17,Tariq2018IMPACTOC}. These cyber attacks have been carried out by experienced hackers that has involved  manual effort. In recent years there has been a boom in the development of artificial intelligence (AI), which has enabled the creation of software tools that have helped to automate tasks such as prediction, information retrieval, and media synthesis. Throughout this period, members of academia and industry have utilized AI\footnote{In this paper, we consider machine learning to be a subset of AI technologies.} in the context of improving the state of cyber defense \cite{mirsky2018kitsune,liu2019machine,mahadi2018survey} and threat analysis \cite{272248,ucci2019survey,cohen2020dante}. However, AI is a double edged sword, and attackers can utilize it to improve their malicious campaigns. 

Recently, there has been a lot of work done to identify and mitigate attacks on AI-based systems (adversarial machine learning) ~\cite{barreno10,huang11,joseph18-advml-book,biggio18-pr,chakraborty18-cs,Papernot18-eurosp}. However, an AI-capable adversary can do much more than poison or fool a machine learning model. Adversaries can improve their tactics to launch attacks that were not possible before. For example, with deep learning one can perform highly effective spear phishing attacks by impersonating a superior's face and voice \cite{mirsky2021creation,fraudsters_mimic:online}. It is also possible to improve stealth capabilities by using automation to perform lateral movement through a network, limiting command and control (C\&C) communication \cite{zelinka2018swarm,truong2019neural}. Other capabilities include the use of AI to find zero-day vulnerabilities in software, automate reverse engineering, exploit side channels efficiently, build realistic fake personas, and to perform many more malicious activities with improved efficacy (more examples are presented later in section \ref{sec:ai_vs_ckc}).

\subsection{Goal}
In this work, we provide a survey of knowledge on offensive AI in the context of enterprise security. The goal of this paper is to help the community (1) better understand the current impact of offensive AI on organizations, (2) prioritize research and development of defensive solutions, and (3) identify trends that may emerge in the near future. This work isn't the first to raise awareness of offensive AI. In \cite{brundage2018malicious} the authors warned the community that AI can be used for unethical and criminal purposes with examples taken from various domains. In \cite{caldwell2020ai} a  workshop was held that attempted to identify the potential top threats of AI in criminology. However, these works relate to the threat of AI on society overall and are not specific to organizations and their networks. 

\subsection{Methodology}
Our survey was performed in the following way. First, we reviewed literature to identify and organize the potential threats of AI to organizations. Then, we surveyed experts from academia, industry, and government to understand which of these threats are actual concerns and why. Finally, using our survey responses, we ranked these threats to gain insights and to help identify the areas which require further attention.
The survey participants were from a wide profile of organizations such as MITRE, IBM, Microsoft, Airbus, Bosch, Fujitsu, Hitachi, and Huawei.

To perform our literature review, we used the MITRE ATT\&CK\footnote{\url{https://attack.mitre.org/matrices/enterprise/}} matrix as a guide. This matrix lists the common tactics (or attack steps) which an adversary performs when attacking an organization, from planning and reconnaissance leading to the final goal of exploitation. We divided the tactics among five different academic workgroups from different international institutions based on expertise. For each tactic in the MITRE ATT\&CK matrix, a workgroup surveyed related works to see how AI has and can be used by an attacker to improve their tactics and techniques. Finally, each workgroup cross inspected each other's content to ensure correctness and completeness.

\subsection{Main Findings}

\noindent\textbf{From the Literature Survey.}
\begin{itemize}
    \item There are three primary motivations for an adversary to use AI: coverage, speed, and success.
    \item AI introduces new threats to organizations. A few examples include the poisoning of machine learning models, theft of credentials through side channel analysis, and the targeting of proprietary training datasets.
    \item Adversaries can employ 33 offensive AI capabilities against organizations. These are categorized into seven groups: (1) automation, (2) campaign resilience, (3) credential theft, (4) exploit development, (5) information gathering, (6) social engineering, and (7) stealth.
    \item Defense solutions, such as AI methods for vulnerability detection \cite{lin2020software}, pen-testing \cite{zerofoxo16:online}, and credential leakage detection \cite{calzavara15-tw} can be weaponized by adversaries for malicious purposes.
\end{itemize}
\noindent\textbf{From the User Study.}
\begin{itemize}
    \item The top three most threatening categories of offensive AI capabilities against organizations are (1) exploit development, (2) social engineering, and (3) information gathering.
    \item 24 of the 33 offensive AI capabilities pose significant threats to organizations.
    \item For the most part, industry and academia are not aligned on the top threats of offensive AI against organizations. Industry is most concerned with AI being used for reverse engineering, with a focus on the loss of intellectual property. Academics, on the other hand, are most concerned about AI being used to perform biometric spoofing (e.g., evading fingerprint and facial recognition).
    \item Both industry and academia ranked the threat of using AI for impersonation (e.g., real-time deepfakes to perpetrate phishing and other social engineering attacks) as their second highest threat. Jointly, industry and academia feel that impersonation is the biggest threat of all.
    \item Evasion of intrusion detection systems (e.g., with adversarial machine learning) is considered to be the least threatening capability of the 24 significant threats, likely due to the adversary's inaccessibility to training data. 
    \item AI impacts the cyber kill chain the most during the initial attack steps. This is because the adversary has access to the environment for training and testing of their AI models. 
    \item Because of an AI's ability to automate processes, adversaries may shift from having a few slow covert campaigns to having numerous fast-paced campaigns to overwhelm defenders and increase their chances of success.
\end{itemize}

\subsection{Contributions}
In this survey, we make the following contributions:
\begin{itemize}
    \item An overview of how AI can be used to attack organizations and its influence on the cyber kill chain (section \ref{sec:attack_model}).
    \item An enumeration and description of the 33 offensive AI capabilities which threaten organizations, based on literature and current events (section \ref{sec:ai_vs_ckc}).
    \item A threat ranking and insights on how offensive AI impacts organizations, based on a user study with members from academia, industry, and government (section \ref{sec:user_study}).
    \item A forecast of the AI threat horizon and the resulting shifts in attack strategies (section \ref{sec:discussion}).
\end{itemize}


\section{Background on Offensive AI}\label{sec:background}
AI is intelligence demonstrated by a machine. It is often associated as a tool for automating some task which requires some level of intelligence. Early AI models were rule based systems designed using an expert's knowledge \cite{Yager84}, followed by search algorithms for selecting optimal decisions (e.g., finding paths or playing games \cite{zheng09}). Today, the most popular type of AI is machine learning (ML) where the machine can gain its intelligence by learning from examples. Deep learning (DL) is a type of ML where an extensive artificial neural network is used as the predictive model. Breakthroughs in DL have led to its ubiquity in applications such as automation, forecasting, and planning due to its ability to reason upon and generate complex data.

\subsection{Training and Execution} 
In general, a machine learning model can be trained on data with an explicit ground-truth (supervised), with no ground-truth (unsupervised), or with a mix of both (semi-supervised). The trade-off between supervised and non-supervised approaches is that supervised methods often have much better performance at a given task, but require labeled data which can be expensive or impractical to collect. Moreover, unsupervised techniques are open-world, meaning that they can identify novel patterns that may have been overlooked.
Another training method is reinforcement learning where a model is trained based on reward for good performance. Lastly, for generating content, a popular framework is adversarial learning. This was first popularised in \cite{goodfellow14-nips} where the generative adversarial network (GAN) was proposed. A GAN uses a discriminator model to `help' a generator model produce realistic content by giving feedback on how the content fits a target distribution.

Where a model is trained or executed depends on the attacker's task and strategy. For example, the training and execution of models for reconnaissance tasks will likely take place offsite from the organization. However, the training and execution of models for attacks may take place onsite, offsite, or both. Another possibility is where the adversary uses few-shot learning \cite{wang20-acmcs} by training on general data offsite and then fine tuning on the target data onsite. 
In all cases, the adversary will first design and evaluate their model offsite prior to its usage on the organization to ensure its success and to avoid detection. 

For onsite execution, an attacker runs the risk of detection if the model is complex (e.g. a DL model). For example when the model is transferred over to the organization's network or when the attacker's model begins to utilize resources, it may trigger the organization's anomaly detection system. To mitigate this issue, the adversary must consider a trade-off between stealth and effectiveness. For example the adversary may (1) execute the model during off hours or on non-essential devices, (2) leverage an insider to transfer the model, or (3) transfer the observations off-site for execution.

\begin{table}[t]
\centering
\setlength\tabcolsep{3pt}
\caption{Examples of where a model can be trained and executed in an attack on an organization. Onsite refers to being within the premisis or network of the organization.}
\label{tab:onoffsite}
\resizebox{0.55\columnwidth}{!}{%
\begin{tabular}{@{}cc|cc|c@{}}
\multicolumn{2}{c|}{\textbf{Training}} & \multicolumn{2}{c|}{\textbf{Execution}} &  \\
Offsite & \cellcolor[HTML]{EFEFEF}Onsite & Offsite & \cellcolor[HTML]{EFEFEF}Onsite & \multirow{-2}{*}{\textbf{Example}} \\ \midrule
$\bullet$ & \cellcolor[HTML]{EFEFEF} & $\bullet$ & \cellcolor[HTML]{EFEFEF} & Vulnerability detection \\
$\bullet$ & \cellcolor[HTML]{EFEFEF} &  & \cellcolor[HTML]{EFEFEF}$\bullet$ & Side channel keylogging \\
 & \cellcolor[HTML]{EFEFEF}$\bullet$ & $\bullet$ & \cellcolor[HTML]{EFEFEF} & Channel compression for exfiltration \\
 & \cellcolor[HTML]{EFEFEF}$\bullet$ &  & \cellcolor[HTML]{EFEFEF}$\bullet$ & Traffic shaping for evasion \\
$\bullet$ & \cellcolor[HTML]{EFEFEF}$\bullet$ &  & \cellcolor[HTML]{EFEFEF}$\bullet$ & Few-shot learning for record tampering \\ \bottomrule
\end{tabular}%
}
\vspace{-2em}
\end{table}

There are two forms of offensive AI: Attacks using AI and attacks against AI. For example, an adversary can (1) use AI to improve the efficiency of an attack (e.g., information gathering, attack automation, and vulnerability discovery) or (2) use knowledge of AI to exploit the defender's AI products and solutions (e.g., to evade a defense or to plant a trojan in a product). The latter form of offensive AI is commonly referred to as adversarial machine learning.

\subsection{Attacks Using AI}
Although there are a wide variety of AI tasks which can be used in attacks, we found the following to be the most common:
\begin{description}[leftmargin=.5cm]
    \item[Prediction] This is the task of making a prediction based on previously observed data. Common examples are classification, anomaly detection, and regression. Examples of prediction for an offensive purpose includes the identification of keystrokes on a smartphone based on motion \cite{hussain2016rise,javed2020alphalogger,marquardt2011sp}, the selection of the weakest link in the chain to attack \cite{Abid18-EDBT-ICDT}, and the localization of software vulnerabilities for exploitation \cite{lin2020software,jiang2019survey,NLP_static}. 
    
    \item[Generation] This is the task of creating content that fits a target distribution which, in some cases, requires realism in the eyes of a human. Examples of generation for offensive uses include the tampering of media evidence \cite{236284,schreyer2019adversarial}, intelligent password guessing  \cite{hitaj2019passgan, garg2019password}, and traffic shaping to avoid detection \cite{novo20-aisec,han2020practical}. Deepfakes are another instance of offensive AI in this category. A deepfake is a believable media created by a DL model. The technology can be used to impersonate a victim by puppeting their voice or face to perpetrate a phishing attack \cite{mirsky2021creation}.
    
    \item[Analysis] This is the task of mining or extracting useful insights from data or a model. Some examples of analysis for offense are the use of explainable AI techniques \cite{ribeiro16-kdd} to identify how to better hide artifacts (e.g., in malware) and the clustering or embedding of information on an organization to identify assets or targets for social engineering.
    
    \item[Retrieval] This is the task of finding content that matches or that is semantically similar to to a given query. For example, in offense, retrieval algorithms can be used to track an object or an individual in a compromised surveillance system \cite{Rahman19-avss,zhu18-tip}, to find a disgruntled employee (as a potential insider) using semantic analysis on social media posts, and to summarize lengthy documents \cite{zhang16-iecon} during open source intelligence (OSINT) gathering in the reconnaissance phase.
    
    \item[Decision Making] The task of producing a strategic plan or coordinating an operation. Examples of this in offensive AI are the use of swarm intelligence to operate an autonomous botnet \cite{castiglione14-jnca} and the use of heuristic attack graphs to plan optimal attacks on networks \cite{bland20-cs}.
\end{description}

\subsection{Attacks Against AI - Adversarial Machine Learning}\label{subsec:advml}
An attacker can use its AI knowledge to exploit ML model vulnerabilities violating its confidentiality, integrity, or availability. Attacks can be staged at either training (development) or test time (deployment) through one of the following attack vectors:
\begin{description}[leftmargin=.5cm]
    \item[Modify the Training Data.] Here the attacker modifies the training data to harm the integrity or availability of the model. Denial of service (DoS) poisoning attacks~\cite{biggio12-icml, munoz-gonzalez17-aisec,koh17-icml} are when the attacker decreases the model's performance until it is unusable. A backdoor poisoning attack~\cite{gu17-nips,chen17-arxiv} or trojaning attack~\cite{liu2017trojaning}, is where the attacker teaches the model to recognize an unusual pattern that triggers a behavior (e.g., classify a sample as safe). A triggerless version of this attack causes the model to misclassify a test sample without adding a trigger pattern to the sample itself~\cite{Shafahi18-nips,aghakhani20-bullseye}
    \item[Modify the Test Data.] In this case, the attacker modifies test samples to have them misclassified~\cite{biggio13-ecml, szegedy14-iclr, goodfellow15-iclr}. For example, altering the letters of a malicious email to have it misclassified as legitimate, or changing a few pixels in an image to evade facial recognition \cite{sharif2016accessorize}. Therefore, these types of attacks are often referred to as evasion attacks. By modifying test samples ad-hoc to increase the model's resource consumption, the attacker can also slow down the model performances.~\cite{shumailov2020sponge}.
    
    \item[Analyze the Model's Responses.] Here, the attacker sends a number of crafted queries to the model and observes the responses to infer information about the model's parameters or training data. To learn about the training data, there are membership inference~\cite{shokri17-sp}, deanonymization~\cite{Narayanan08-sp}, and model inversion~\cite{Model_Inversion} attacks. For learning about the model's parameters there are model stealing/extraction~\cite{Juuti19-sp,jia20}, and blind-spot detection~\cite{zhang19-iclr}, state prediction~\cite{woh18-scis}. 
    \item[Modify the Training Code.] This is where the attacker performs a supply chain attack by modifying a library used to train ML models (e.g., via an open source project). For example, a compromised loss (training) function that inserts a backdoor~\cite{Bagdasaryan20}.
    \item[Modify the Model's Parameters.] In this attack vector, the attacker accesses a trained model (e.g., via a model zoo or security breach) and tamper its parameters to insert a latent behavior. These attacks can be performed at the software~\cite{yao19-ccs,wang20,wang20} or hardware~\cite{breier2018practical} levels (a.k.a. fault attacks).
\end{description}
Depending on the scenario, an attacker may not have full knowledge or access to the target model:
\begin{itemize}
\item{\textbf{White-Box  (Perfect-Knowledge) Attacks:}} The attacker knows everything about the target system. This is the worst case for the system defender. Although it is not very likely to happen in practice, this setting is interesting as it provides an empirical upper bound on the attacker's performance.
\item{\textbf{Gray-Box (Limited-Knowledge) Attacks:}} The attacker has partial knowledge of the target system (e.g., the learning algorithm, architecture, etc.) but no knowledge of training data or the model's parameters. 
\item{\textbf{Black-Box (Zero-Knowledge) Attacks:}}
The attacker knows only the task the model is designed to perform and which kind of features are used by the system in general (e.g., if a malware detector has been trained to perform static or dynamic analysis). The attacker may also be able to analyse the model's responses in a black-box manner to get feedback on certain inputs.
\end{itemize}
In a black or gray box scenario, the attacker can build a surrogate ML model and try to devise the attacks against it as the attacks often transfer between different models. ~\cite{biggio13-ecml,demontis19-usenix}.

An attacker does not need to be an expert at machine learning to implement these attacks. Many can be acquired from open-source libraries online~\cite{melis2019secml, art2018-arxiv,papernot2018cleverhans,croce2020-icml}.



\section{Offensive AI vs Organizations}\label{sec:attack_model}
In this section, we provide an overview of offensive AI in the context of organizations. First we review a popular attack model for enterprise. Then we will identify how an AI-capable adversary impacts this model by discussing the adversary's new motivations, goals, capabilities, and requirements. Later in section \ref{sec:ai_vs_ckc}, we will detail the adversary's techniques based on our literature review.  

\subsection{The Attack Model}\label{subsec:basic_model}
There are a variety of threat agents which target organizations. These agents are cyber terrorists, cyber criminals, employees, hacktivists, nation states, online social hackers, script kiddies, and other organizations like competitors. There are also some non-target specific agents, such as certain botnets and worms, which threaten the security of an organization. A threat agent may be motivated for various reasons. For example, to (1) make money through theft or ransom, (2) gain information through espionage, (3) cause physical or psychological damage for sabotage, terrorism, fame, or revenge, (4) reach another organization, and (5) obtain foothold on the organization as an asset for later use \cite{TargetHa48:online}.
These agents not only pose a threat to the organization, but also its employees, customers, and the general public as well (e.g., attacks on critical infrastructure). 

In an attack, there may be number of attack steps which the threat agent must accomplish. These steps depend on the adversary's goal and strategy. For example, in an advanced persistent threat (APT)~\cite{messaoud2016advanced,chen2018special,alshamrani2019survey}, the adversary may need to reach an asset deep within the defender's network. This would require multiple steps involving reconnaissance, intrusion, lateral movement through the network, and so on. However, some attacks can involve just a single step. For example, a spear phishing attack in which the victim unwittingly provides confidential information or even transfers money. In this paper, we describe the adversary's attack steps using the MITRE ATT\&CK Matrix for Enterprise\footnote{\url{https://attack.mitre.org/}} which captures common adversarial tactics based on real-world observations.

Attacks which involve multiple steps can be thwarted if the defender identifies or blocks the attack early on. The more progress which an adversary makes, the harder it is for the defender to mitigate it. For example, it is better to stop a campaign during the initial intrusion phase than during the lateral movement phase where an unknown number of devices in the network have been compromised. This concept is referred to as the \textit{cyber kill chain}. From an offensive perspective, the adversary will want shorten and obscure the kill chain by being as to be as efficient and covert as possible. In particular, operation within a defender's network usually requires the attacker to operate through a remote connection or send commands to compromised devices (bots) from a command and control (C2). This generates presence in the defenders network which can be detected over time.

\subsection{The Impact of Offensive AI}
Conventional adversaries use manual effort, common tools, and expert knowledge to reach their goals. In contrast, an AI-capable adversary can use AI to automate its tasks, enhance its tools, and evade detection. 
These new abilities affect the cyber kill chain. 

First, let's discuss why an adversary would consider using AI in its offensive on an organization.

\subsubsection{The Three Motivators of Offensive AI}\label{subsubsec:motivators}
In our survey, we found that there are three core motivations for an adversary to use AI in an offensive against an organization: coverage, speed, and success.

\begin{description}[leftmargin=.5cm]
    \item[Coverage.] By using AI, an adversary can scale up its operations through automation to decrease human labor and increase the chances of success. For example, AI can be used to automatically craft and launch spear phishing attacks, distil and reason upon data collected from OSINT, maintain attacks on multiple organizations in parallel, and reach more assets within a network to gain a stronger foothold. In other words, AI enables adversaries to target more organizations with higher precision attacks with a smaller workforce.
    
    \item[Speed.] With AI, an adversary can reach its goals faster. For example, machine learning can be used to help extract credentials, intelligently select the next best target during lateral movement, spy on users to obtain information (e.g., perform speech to text on eavesdropped audio), or find zero-days in software. By reaching a goal faster, the adversary not only saves time for other ventures but can also minimize its presence (duration) within the defender's network.
    
    \item[Success.] By enhancing its operations with AI, an adversary increases its likelihood of success. Namely, ML can be used to (1) make the operation more covert by minimizing or camouflaging network traffic (such as C2 traffic) and by exploiting weaknesses in the defender's AI models such as an ML-based intrusion detection system (IDS), (2) identify opportunities such as good targets for social engineering attacks and novel vulnerabilities, (3) enable better attack vectors such as using deepfakes in spear phishing attacks, (4) plan optimal attack strategies, and (5) strengthen persistence in the network through automated bot coordination and malware obfuscation. 
\end{description}

We note that these motivations are not mutually exclusive. For example, the use of AI to automate a phishing campaign increases coverage, speed, and success. 

\subsubsection{AI-Capable Threat Agents}
It is clear that some AI-capable threat agents will be able to perform more sophisticated AI attacks than others. For example, state actors can potentially launch intelligent automated botnets where hacktivists will likely struggle in accomplishing the same. However, we have observed over the years that AI has become increasingly accessible, even to novice users. For example, there are a wide variety of open source deepfakes technologies online which are plug and play\footnote{\url{https://github.com/datamllab/awesome-deepfakes-materials}}. Therefore, the sophistication gap between certain threat agents may close over time as the availability to AI technology increases.

\begin{figure*}[t]
    \centering
    \includegraphics[width=\textwidth]{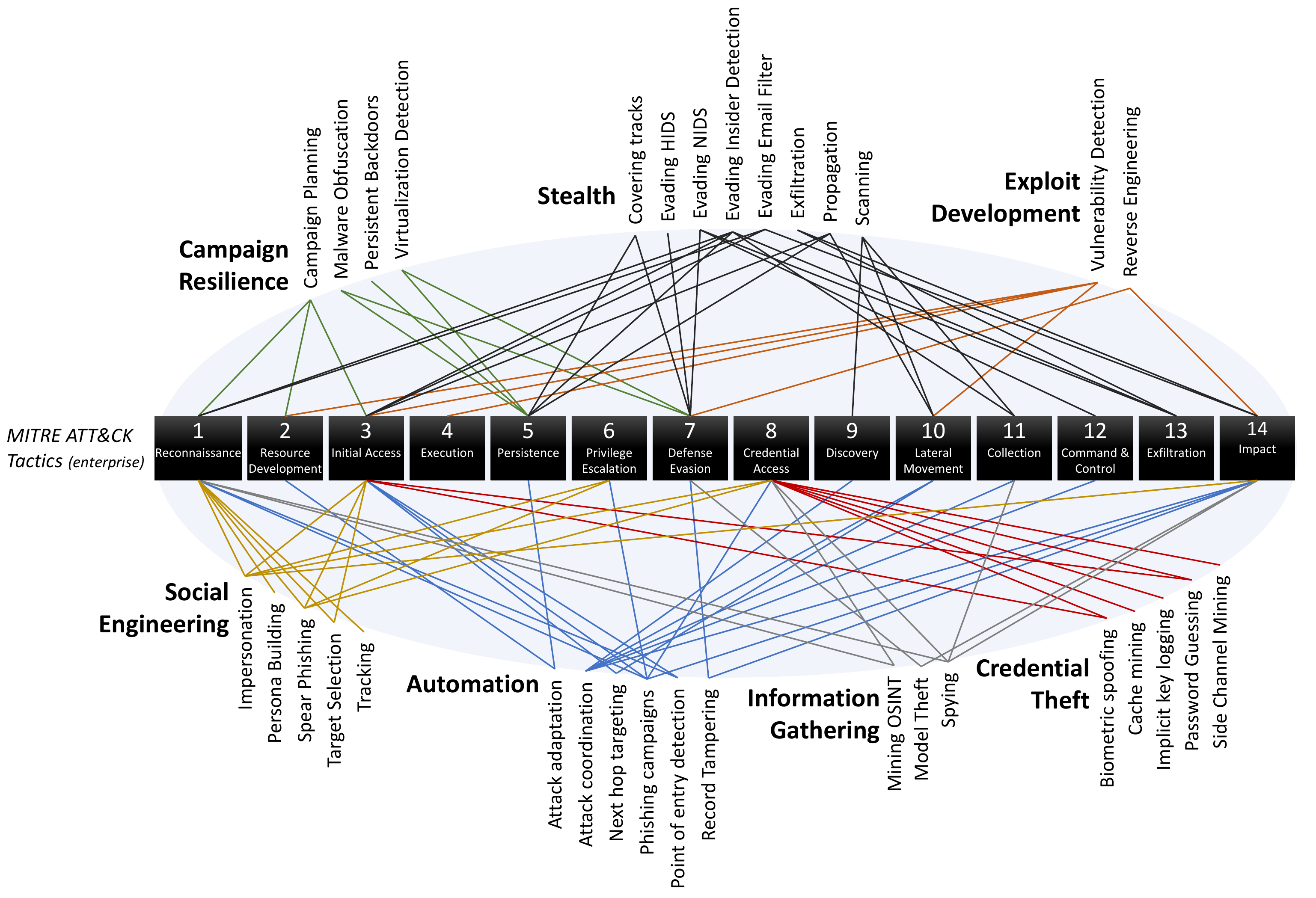}
    \caption{The 33 offensive AI capabilities (OAC) identified in our survey, mapped to the MITRE enterprise ATT\&CK model. An edge indicates that the OAC directly helps the attacker achieve the indicated attack step.}
    \label{fig:oac}
\end{figure*}

\subsubsection{New Attack Goals}
In addition to the conventional attack goals, AI-capable adversaries have new attack goals as well:

\begin{description}[leftmargin=.5cm]
\item[Sabotage.] The adversary may want to use its knowledge of AI to cause damage to the organization. For example, it may want to alter ML models in the organization's products and solutions by poisoning their dataset to alter performance or by planting a trojan in the model for later exploitation. Moreover, the adversary may want to perform an adversarial machine learning attack on an AI system. For example, to evade detection in surveillance \cite{sharif2016accessorize} or to tip financial or energy forecasts models in the adversary's favor. Finally, the adversary may also use generative AI to add or modify evidence in a realistic manner. For example, to modify or plant evidence in surveillance footage \cite{DeepFake24:online}, medical scans \cite{236284}, or financial records \cite{schreyer2019adversarial}.

\item[Espionage.] With AI, an adversary can improve its ability to spy on organizations and extract/infer meaningful information. For example, they can use speech to text algorithms and sentiment analysis to mine useful audio recordings \cite{
8636124} or steal credentials through acoustic or motion side channels \cite{liu2015snooping, shumailov2019hearing}. AI can also be used to extract latent information from encrypted web traffic \cite{236286}, and track users  through the organization's social media \cite{malhotra12-ASONAM}. Finally, the attacker may want to achieve an autonomous persistent foothold using swarm intelligence \cite{zelinka2018swarm}.

\item[Information Theft.] An AI-capable adversary may want to steal models trained by the organization to use in future white box adversarial machine learning attacks. Therefore, some data records and proprietary datasets may be targeted for the sake of training models. In particular, audio or video records of customers and employees may be stolen to create convincing deepfake impersonations. Finally, intellectual property may be targeted through AI powered reverse engineering tools \cite{hajipour20-arxiv}.
\end{description}

\subsubsection{New Attack Capabilities}\label{subsubsec:OAC}
Through our survey, we have identified 33 offensive AI capabilities (OAC) which directly improve the adversary's ability to achieve attack steps. These OACs can be grouped into seven OAC categories: (1) automation, (2) campaign resilience, (3) credential theft, (4) exploit development, (5) information gathering, (6) social engineering, and (7) stealth. Each of these capabilities can be tied to the three motivators introduced in section \ref{subsubsec:motivators}. 

In Fig. \ref{fig:oac}, we present the OACs and map their influence on the cyber kill chain (the MITRE enterprise ATT\&CK model). An edge in the figure means that the indicated OAC improves attacker's ability to achieve the given attack step. From the figure, we can see that offensive AI impacts every aspect of the attack model. Later in section \ref{sec:ai_vs_ckc} we will discuss each of these 33 OACs in greater detail. 

These capabilities are  materialized in one of two ways: 
\begin{description}[leftmargin=.5cm]
\item[AI-based tools] are programs which performs a specific task in adversary's arsenal. For example, a tool for intelligently predicting passwords  \cite{hitaj2019passgan, garg2019password}, obfuscating malware code \cite{datta2020deepobfuscode}, traffic shaping for evasion \cite{li19-cns,novo20-aisec,han2020practical}, puppeting a persona \cite{mirsky2021creation}, and so on. These tools are typically in the form of a machine learning model.

\item[AI-driven bots] are autonomous bots which can perform one or more attack steps without human intervention, or coordinate with other bots to efficiently reach their goal. These bots may use a combination of swarm intelligence \cite{castiglione14-jnca} and machine learning to operate.
\end{description}

\section{Survey of Offensive AI Capabilities}\label{sec:ai_vs_ckc}


In section \ref{subsubsec:OAC} we presented the 33 offensive AI capabilities. We will now describe each of the OACs in order of their 7 categories: automation, campaign resilience, credential theft, exploit development, information gathering, social engineering, and stealth. 

\subsection{Automation}
The process of automation gives adversaries a hands-off approach to accomplishing attack steps. This not only reduces effort, but also increases the adversary's flexibility and enables larger campaigns which are less dependent on C2 signals.

\subsubsection{Attack Adaptation}
Adversaries can use AI to help adapt their malware and attack efforts to unknown environments and find their intended targets. For example, identifying a system \cite{crowdstrike_2020} before attempting an exploit to increase the chances of success and avoid detection. In Black Hat'18, IBM researchers showed how a malware can trigger itself using DL by identifying a target's machine by analysing the victim's face, voice, and other attributes. With models such as decision trees, malware can locate and identify assets via complex rules like~\cite{lunghi2017untangling,leong2019messagetap}. Instead of transferring screenshots \cite{brumaghin2018old,arsene2020oil,zhang2018analysis,mueller2018indictment} DL can be used onsite to extract critical information. 

\subsubsection{Attack Coordination}
Cooperative bots can use AI to find the best times and targets to attack. For example, swarm intelligence \cite{beni2020swarm} is the study of autonomous coordination among bots in a decentralized manner. Researchers have proposed that botnets can use swarm intelligence as well. In \cite{zelinka2018swarm} the authors discuss a hypothetical swarm malware and in \cite{truong2019neural} the authors propose another which uses DL to trigger attacks. AI bots can also communicate information on asset locations to fulfill attacks (e.g., send a stolen credential or relevant exploit to a compromised machine).

\subsubsection{Next hop targeting}\label{subsubsec:next_hop}
During lateral movement, the adversary must select the next asset to scan or attack. Choosing poorly may prolong the attack and risk detection by the defenders. For example, consider a browser like Firefox which has 4325 key-value pairs denoting the individual configurations. 
Only some inter-plays of these configurations are vulnerable~\cite{otsuka2015learning,chen2014detecting}. Reinforcement learning can be used to train a detection model which can identify the best browser to target. 
As for planning multiple steps, a strategy can be formed  by using reinforcement learning on Petri nets \cite{bland20-cs} where attackers and defenders are modeled as competing players. Another approach is to use DL~\cite{Yousefi18-ICTSP, wu21} to explore  ``attack graphs" \cite{ou05-usenix-mulval} that contain the target's network structure and the vulnerabilities. Notably, the Q-learning algorithms have enabled the approach to work on large-scale enterprise networks~\cite{matta19-el}.


\subsubsection{Phishing Campaigns}
Phishing campaigns involve sending the same emails or robo-phone calls in mass. When someone falls prey and responds, the adversary takes over the conversation. These campaigns can be fully automated through AI like Google's assistant which can make phone calls on your behalf \cite{leviathan2018google,singh2020survey,rebryk2020convoice}. Furthermore, adversaries can increase their success through mass spear phishing campaigns powered with deepfakes, where (1) a bot calls a colleague of the victim (found via social media), (2) clones his/her voice with 5 seconds of audio \cite{jia18-NIPS}, and then (3) calls the victim in the colleague's voice to exploit their trust.

\subsubsection{Point of Entry Detection}
The adversary can use AI to identify and select the best attack vector for an initial infection. For example, in  \cite{cyberintrusionstats2019} statistical models on an organization's attributes were used to predict the number of intrusions it receives. The adversary can train a model on similar information to select the weakest organizations (low hanging fruits) and the strongest attack vectors. 

\subsubsection{Record Tampering}
An adversary may use AI to tamper records as part of their end-goal. For example, ML can be used to impact business decisions with synthetic data \cite{kumar2018ecommercegan}, to obstruct justice by tampering evidence \cite{DeepFake24:online}, to perform fraud \cite{schreyer2019adversarial} or to modify medical or satellite imagery \cite{236284}. As shown in \cite{236284}, DL-tampered records can fool human observers and can be accomplished autonomously onsite.

\subsection{Campaign Resilience}
In a campaign, adversaries try to ensure that their infrastructure and tools have a long life. Doing so helps maintain a foothold in the organization and enables reuse of tools and exploits for future and parallel campaigns. AI can be used to improve campaign resilience through planning, persistence, and obfuscation.

\subsubsection{Campaign Planning}
Some attacks require careful planning long before the attack campaign to ensure that all of the attacker's tools and resources are obtainable. ML-based cost benefit analysis tools, such as in \cite{manning2018towards}, may be used to identify which tools should be developed and how the attack infrastructure should be laid out (e.g., C2 servers, staging areas, etc). It could also be used to help identify other organizations that can be used as beach heads \cite{TargetHa48:online}. Moreover, ML can be used to plan a digital twin \cite{fuller2020digital,bitton2018deriving} of the victim's network (based on information from reconnaissance) to be created offsite for tuning AI models and developing malware.

\subsubsection{Malware Obfuscation}
ML models such as GANs can be used to obscure a malware's intent from an analyst. Doing so can enable reuse of the malware, hide the attacker's intents and infrastructure, and prolong an attack campaign. The concept is to take an existing piece of software and emit another piece that is functionally equivalent (similar to translation in NLP). For example, DeepObfusCode \cite{datta2020deepobfuscode} uses recurrent neural networks (RNN) to generate ciphered code. Alternatively, backdoors can be planted in open source projects and hidden using similar manners \cite{pasandi2019approximate}.

\subsubsection{Persistent Access}
An adversary can have bots establish multiple back doors per host and coordinate reinfection efforts among a swarm \cite{zelinka2018swarm}. Doing so achieves a foothold on an organization by slowing down the effort to purge the campaign. To avoid detection in payloads deployed during boot, the adversary can use a two-step payload which uses ML to identify when to deploy the malware and avoid detection \cite{Anderson2017EvadingML, Fang2020MalwareEvasion}. Moreover, a USB sized neural compute stick\footnote{https://software.intel.com/content/www/us/en/develop/articles/intel-movidius-neural-compute-stick.html} can be planted by an insider to enable covert and autonomous onsite DL operations.

\subsubsection{Virtualization Detection}
To avoid dynamic analysis and detection in sandboxes, an adversary may try to have the malware detect the sandbox before triggering. The malware could use ML to detect a virtual environment by measuring system timing (e.g., like in \cite{perianin2020end}) and other system properties.

\subsection{Credential Theft}
Although a system may be secure in terms of access control, side channels can be exploited with ML to obtain a user's credentials and vulnerabilities in AI systems can be used to avoid biometric security.  

\subsubsection{Biometric spoofing}
Biometric security is used for access to terminals (such as smartphones) and for performing automated surveillance  \cite{chinaSurv,wang2017face,ding2018editorial}. Recent works have shown how AI can generate ``Master Prints" which are deepfakes of fingerprints that can open nearly any partial print scanner (such as on a smartphone) \cite{bontrager2018deepmasterprints}. Face recognition systems can be fooled or evaded with the use of adversarial samples. For example, in \cite{sharif2016accessorize} where the authors generated colorful glasses that alters the perceived identity. Moreover, `sponge' samples \cite{shumailov2020sponge} can be used to slow down a surveillance camera until it is unresponsive or out of batteries (when remote). Voice authentication can also be evaded through adversarial samples, spoofed voice \cite{wang2019asvspoof}, and by cloning the target's voice with deep learning \cite{wang2019asvspoof}.

\subsubsection{Cache mining}
Information on credentials can be found in a system's cache and log dumps, but the large amount of data makes finding it a difficult task. However, the authors of \cite{wang2019unveiling} showed how ML can be used to identify credentials in cache dumps from graphic libraries. Another example is the work of \cite{calzavara15-tw} where an ML system was used to identify cookies containing session information. 

\subsubsection{Implicit key logging}
Over the last few years researchers have shown how AI can be used as an implicit key-logger by sensing side channel information from a physical environment. The side channels comes in one or a combination of the following aspects:
\begin{description}[leftmargin=.5cm]
    \item[Motion.] When tapping on a phone screen or typing on a keyboard, the device and nearby surfaces move and vibrate. A malware can use the smartphone's motion sensors to decipher the touch strokes on the phone \cite{hussain2016rise,javed2020alphalogger} and keystrokes on nearby keyboards \cite{marquardt2011sp}. Wearable devices can be exploited in a similar way as well \cite{liu2015good,maiti2018side}.
    \item[Audio.] Researchers have shown that, when pressed, each key gives of it's own unique sound which can be used to infer what is being typed \cite{liu2015snooping,compagno2017don}. Timing between key strokes is also a revealing factor due to the structure of the language and keyboard layout. Similar approaches have also been shown for inferring touches on smartphones \cite{shumailov2019hearing,yu2019indirect,lu2019keylisterber}.
    \item[Video.] In some cases, a nearby smartphone or compromised surveillance camera can be used to observe keystrokes, even when the surface is obscured. For example, via eye movements \cite{chen2018eyetell,wang2018gazerevealer,wang2019your}, device motion \cite{sun2016visible}, and hand motion \cite{balagani2018silk,lim2020revisiting}.
\end{description}

\subsubsection{Password Guessing}
Humans tend to select passwords with low entropy or with personal information such as dates. GANs can be used to intelligently brute-force passwords by learning from leaked password databases \cite{hitaj2019passgan}. Researchers have improved on this approach by using RNNs in the generation process \cite{nam2020recurrent}. However, the authors of \cite{garg2019password} found that models like \cite{hitaj2019passgan} do not work well on Russian passwords. Instead, adversaries may pass the GAN personal information on the user to improve the performance \cite{seymour2018generative}.

\subsubsection{Side Channel Mining}
ML algorithms are adept at extracting latent patterns in noisy data. Adversaries can leverage ML to extract secrets from side channels emitted from cryptographic algorithms. This has been accomplished on a variety of side channels including power consumption~\cite{kocher1999differential,lerman2014power}, electromagnetic emanations~\cite{gandolfi2001electromagnetic}, processing time~\cite{brumley2005remote}, cache hits/misses\cite{perianin2020end}. In general, ML can be used to mine nearly any kind of side channel \cite{lerman2013time, 10.1007/978-3-030-35869-3_8, picek2019curse,cagli2017convolutional,heuser2016side,picek2018performance,maghrebi2016breaking,Perin_Chmielewski_Batina_Picek_2020}. For example, credentials can be extracted from the timing of network traffic \cite{song2001timing}. 

\subsection{Exploit Development}
Adversaries work hard to understand the content and inner-workings of compiled software to (1) steal intellectual property, (2) share trade secrets, (3) and identify vulnerabilities which they can exploit.


\subsubsection{Reverse Engineering}
While interpreting compiled code, an adversary can use ML to help identify functions and behaviors, and guide the reversal process. For example binary code similarity can be used to identify well-known or reused behaviors \cite{shin2015recognizing,Xu_2017,bao2014byteweight,liu2018alphadiff,ding2019asm2vec,duan2020deepbindiff,ye2020misim} and autoencoder networks can be used to segment and identify behaviors in code, similar to the work of \cite{272248}. Furthermore, DL can potentially be used to lift compiled code up to a higher-level representation using graph transformation networks \cite{yun2019graph}, similar to semantic analysis in language processing. Protocols and state machines can also be reversed using ML. For example, CAN bus data in a vehicles \cite{huybrechts2017automatic}, network protocols \cite{li2015protocol}, and commands \cite{bossert2014towards, wang2011inferring}.

\subsubsection{Vulnerability Detection} 
There are a wide variety of software vulnerability detection techniques which can be broken down into static and dynamic approaches:

\begin{description}[leftmargin=.5cm]
    \item[Static.] For open source applications and libraries, the attacker can use ML tools for detecting known types of vulnerabilities in source code  \cite{NLP_static,feng2016scalable,li2018vuldeepecker,li2019comparative,chakraborty2020deep}. If its a commercial product (compiled as a binary) then methods such as \cite{272248} can be used to identify vulnerabilities by comparing parts of the program's control flow graph to known vulnerabilities.
    \item[Dynamic.] ML can also be used to perform guided input `fuzzing' which can reach buggy code faster \cite{she2019neuzz,she2020mtfuzz,wang2020systematic,li2020v,lin2020software,cheng2019optimizing,atlidakis2020pythia}. Many works have also shown how AI can mitigate the issue of symbolic execution's massive state space \cite{janota2018towards,samulowitz2007learning,liang2018machine,kurin2019improving,jiang2019survey}.
\end{description}


\subsection{Information Gathering}
AI scales well and is very good at data mining and language processing. These capabilities can be used by an adversary to collect and distil actionable intel for a campaign. 

\subsubsection{Mining OSINT}
In general, there are three ways in which AI can improve an adversary's OSINT. 
\begin{description}[leftmargin=.5cm]
    \item[Stealth.] The adversary can use AI to camouflage its probe traffic to resemble benign services like Google's web crawler  \cite{cohen2020dante}. Unlike heavy tools like Metagoofil \cite{laramies81:online}, ML can be used to minimize interactions by prioritizing sites and data elements \cite{ghazi2018supervised,guo2019deep}.
    \item[Gathering.] Network structure and elements can be identified using cluster analysis or graph-based anomaly detection \cite{akoglu2015graph}. Credentials and asset information can be found using methods like reinforcement learning on other organizations \cite{schwartz2019autonomous}. Finally, personnel structure can be extracted from social media using NLP-based web scrappers like Oxylabs\cite{Innovati92:online}.
 
    \item[Extraction.] Techniques like NLP can be used to translate foreign documents \cite{dabre2020survey}, identify relevant documents \cite{nasar2019textual,evangelista2020systematic}, extract relevant information from online sources \cite{github_telegram_contest,twds_building_news_aggr}, and locate valid identifiers\cite{malhotra12-ASONAM}.
\end{description}

\subsubsection{Model Theft}
An adversary may want to steal an AI model to (1) obtain it as intellectual property, (2) extract information about members of its training set \cite{shokri17-sp,Narayanan08-sp,Model_Inversion}, or (3) use it to perform a white-box attack against an organization. As described in section \ref{subsec:advml}, if the model can be queried (e.g., model as a service -MAAS), then its parameters~\cite{Juuti19-sp,jia20} and hyperparameters ~\cite{wang18-sp} can be copied by observing the model's responses. This can also be done through side-channel~\cite{Batina19-usenix} or hardware-level analysis~\cite{breier2020sniff}.


\subsubsection{Spying}
DL is extremely good at processing audio and video, and therefore can be used in spyware. For example, a compromised smartphone can map an office by (1) modeling each room with ultrasonic echo responses \cite{zhou2017batmapper}, (2) using object recognition \cite{jiao2019survey} to obtain physical penetration info (control terminals, locks, guards, etc), and (3) automatically mine relevant information from overheard conversations \cite{ren2019almost,nasar2019textual}. ML can also be used to analyze encrypted traffic. For example it can extract transcripts from encrypted voice calls~\cite{white11-sp}, identify applications ~\cite{al2020man}, and reveal internet searches \cite{236286}.

\subsection{Social Engineering}\label{subsec:socialeng}
The weakest links in an organization's security are its humans. Adversaries have long targeted humans by exploiting their emotions and trust. AI provides adversaries will enhanced capabilities to exploit humans further.

\subsubsection{Impersonation (Identity Theft)}
An adversary may want to impersonate someone for a scam, blackmail attempt, a defamation attack, or to perform a spear phishing attack with their identity. This can be accomplished using deepfake technologies which enable the adversary to reenact (puppet) the voice and face of a victim, or alter existing media content of a victim \cite{mirsky2021creation}. Recently, the technology has advanced to the state where reenactment can be performed in real-time \cite{nirkin2019fsgan}, and training only requires a few images \cite{Siarohin_2019_NeurIPS} or seconds of audio \cite{jia18-NIPS} from the victim. For high quality deepfakes, large amounts of audio/video data is still needed. However, when put under pressure, a victim may trust a deepfake even if it has a few abnormalities (e.g., in a phone call) \cite{workman2008wisecrackers}. Moreover, the audio/video data may be an end-goal and inside the organization (e.g., customer data).

\subsubsection{Persona Building}
Adversaries build fake personas on online social networks (OSN) to connect with their targets. To evade fake profile detectors, a profile can be cloned and slightly altered using AI \cite{salminen2019future,salminen2020enriching,spiliotopoulos2020data} so that they will appear different yet reflect the same personality. The adversary can then use a number of AI techniques to alter or mask the photos from detection \cite{shan2020fawkes,sun2018hybrid,li2019hiding,shaoanlu87:online}.
To build connections, a link prediction model can be used to maximize the acceptance rate \cite{wang2019deeptrust,kong2020dynamic} and a DL chatbot can be used to maintain the conversations~\cite{roller20-arxiv}.

\subsubsection{Spear Phishing}
Call-based spear phishing attacks can be enhanced using real-time deepfakes of someone the victim trusts. For example, this occured in 2019 when a CEO was scammed out \$240k \cite{fraudsters_mimic:online}. For text-based phishing, tweets \cite{zerofoxo16:online} and emails \cite{seymour2016weaponizing,seymour2018generative,das2019automated} can be generated to attract a specific victim, or style transfer techniques can be used to mimic a colleague \cite{fu2017style,yang2018unsupervised}.

\subsubsection{Target Selection}
An adversary can use AI to identify victims in the organization who are the most susceptible to social engineering attacks ~\cite{Abid18-EDBT-ICDT}. A regression model based on the target's social attributes (conversations, attended events, etc) can be used as well. Moreover, sentiment analysis can be used to find disgruntled employees to be recruited as insiders \cite{facewall_2019_Panagiotou,8636124,dhaoui2017social,ghiassi2018domain,8530517}.

\subsubsection{Tracking}
To study members of an organization, adversaries may track the member's activities. With ML, an adversary can trace personnel across different social media sites by content~\cite{malhotra12-ASONAM} and through facial recognition \cite{blackhat18-socialmapper}. ML models can also be used on OSN content to track a member's location \cite{pellet2019localising}. Finally, ML can also be used to discover hidden business relationships~\cite{zhang12-ICIS, ma09-DSC} from the news and from OSNs as well \cite{kumar2016improving,zhang2018link}.

\subsection{Stealth}
In multi step attacks, covert operations are necessary to ensure success. An adversary can either use or abuse AI to evade detection.

\subsubsection{Covering tracks}
To hide traces of the adversary's presence, anomaly detection can be performed on the logs to remove abnormal entries~\cite{cao17-iccc,debnath18-icdcs}. CryptoNets \cite{gilad2016cryptonets} can also be used to hide malware logs and onsite training data for later use. To avoid detection onsite, trojans can be planted in DL intrusion detection systems (IDS) in a supply chain attack at both the hardware \cite{breier2018practical,breier2018deeplaser} and software \cite{liu2017trojaning,li2020deep} levels. DL hardware trojans can use adversarial machine learning to avoid being detected \cite{hasegawa2020trojan}. 


\subsubsection{Evading HIDS (Malware Detectors)}
The struggle between security analysts and malware developers is a never-ending battle, with the malware quickly evolving and defeating detectors. In general, state-of-the-art detectors are vulnerable to evasion~\cite{kolosnjaji18-eusipco,demontis19-tdsc,maiorca20-cs}. 
For example, adversary can evade an ML-based HIDS that performs dynamic analysis by splitting the malware's code into small components executed by different processes~\cite{ispoglou16-usenix}. They can also evade ML-based detectors that perform static analysis by adding bytes to the executable~\cite{suciu19-spw} or code that does not affect the malware behavior~\cite{demetrio20-arxiv-blackbox,pierazzi20-sp, anderson18-arxiv, Zhiyang19-ieeeaccess, Fang2020MalwareEvasion}. Modifying the malware without breaking its malicious functionality is not easy. Attackers may use AI explanation tools like LIME~\cite{ribeiro16-kdd} to understand which parts of malware are being recognized by the detector and change them manually. Tools for evading ML-based detection can be found freely online ~\footnote{ \url{https://github.com/zangobot/secml\_malware}}.

\subsubsection{Evading NIDS (Network Intrusion Detection Systems)}
There are several ways an adversary can use AI to avoid detection while entering, traversing, and communicating over an organization's network. 
Regarding URL-based NIDSs, attackers can avoid phishing detectors by generating URLS that do not match known examples \cite{bahnsen2018deepphish}. Bots trying to contact their C2 server can generate URLs that appear legitimate to humans \cite{peck2019charbot}, or that can evade malicious-URL detectors\cite{sidi2020maskdga}. To evade traffic-based NIDSs, adversaries can shape their traffic \cite{novo20-aisec,han2020practical} or change their timing to hide it\cite{sharon2021tantra}.

\subsubsection{Evading Insider Detectors}
To avoid insider detection mechanisms, adversaries can mask their operations using ML. For example,  given some user's credentials, they can use information on the user's role and the organization's structure to ensure that operation performed looks legitimate~\cite{sutro20}.

\subsubsection{Evading Email Filter}
Many email services use machine learning to detect malicious emails. However, adversaries can use adversarial machine learning to evade detection~\cite{dalvi04,lowd05, lowd05-ceas,gao18-spw}. Similarly, malicious documents attached to emails, containing malware, can evade detection as well (e.g.,~\cite{li2020feature}). Finally, an adversary may send emails to be intentionally detected so that they will be added to the defender's training set, as part of a poisoning attack \cite{biggio11-mcs}.

\subsubsection{Exfiltration}
Similar to evading NIDSs, adversaries must evade detection when trying to exfiltrate data outside of the network. This can be accomplished by shaping traffic to match the outbound traffic~\cite{li19-cns} or by encoding the traffic within a permissible channel like Facebook chat \cite{rigaki2018bringing}. To hide the transfer better, an adversary could use DL to compress~\cite{patel2019survey} and even encrypt~\cite{abadi16-arxiv} the data being exfiltrated. To minimize throughput, audio and video media can be summarized to textual descriptions onsite with ML before exfiltration. Finally, if the network is air gapped (isolated from the Internet) \cite{guri2018bridgeware} then DL techniques can be used to hide data within side channels such as noise in audio \cite{jiang2020smartsteganogaphy}.

\subsubsection{Propagation \& Scanning}
For stealthy lateral movement, an adversary can configure their Petri nets or attack graphs (see section \ref{subsubsec:next_hop}) to avoid assets and subnets with certain IDSs and favour networks with more noise to hide in. Moreover, AI can be used to scan hosts and networks covertly by modeling its search patterns and network traffic according to locally observed patterns \cite{li19-cns}.

\section{User Study \& Threat Ranking}\label{sec:user_study}
In our literature review (section \ref{sec:ai_vs_ckc}) we identified the potential offensive AI capabilities (OAC) which an adversary can use to attack an organization. However, some OACs may be impractical, where others may pose much larger threats. Therefore, we performed a user study to rank these threats and understand their impact on the cyber kill chain.

\subsection{Survey Setup}
We surveyed 22 experts in both subjects of AI and cybersecurity. Our participants were CISOs, researchers, ethics experts, company founders, research managers, and other relevant professions. Exactly half of the participants were from academia and the other half were from industry (companies and government agencies). For example, some of our participants were from MITRE, IBM Research, Microsoft,  Airbus,  Bosch (RBEI), Fujitsu Ltd., Hitachi Ltd., Huawei Technologies, Nord Security, Institute for Infocomm Research (I2R), Purdue University, Georgia Institute of Technology, Munich Research Center, University of Cagliari, and the Nanyang Technological University (NTU). The responses of the participants have been anonymized and reflect their own personal views and not the views of their employers. 

The survey consisted of 204 questions which asked the participants to (1) rate different aspects of each OAC, (2) give their opinion on the utility of AI to the adversary in the cyber kill chain, and (3) give their opinion on the balance between the attacker and defender when both have AI. We used these responses to produce threat rankings and to gain insights on the threat of offensive AI to organizations.

Only 22 individuals participated in the survey because AI-cybersecurity experts are very busy and hard to reach. However, assuming there are 100k eligible respondents in the population, with a confidence level of 95\% we calculate that we have a margin of error of about 20\%. Moreover, since we have sampled a variety of major universities and companies, and since deviation in the responses is relatively small, we believe that the results capture a fair and meaningful view of the subject matter.

\begin{figure*}[t]
    \centering
    \includegraphics[width=\textwidth]{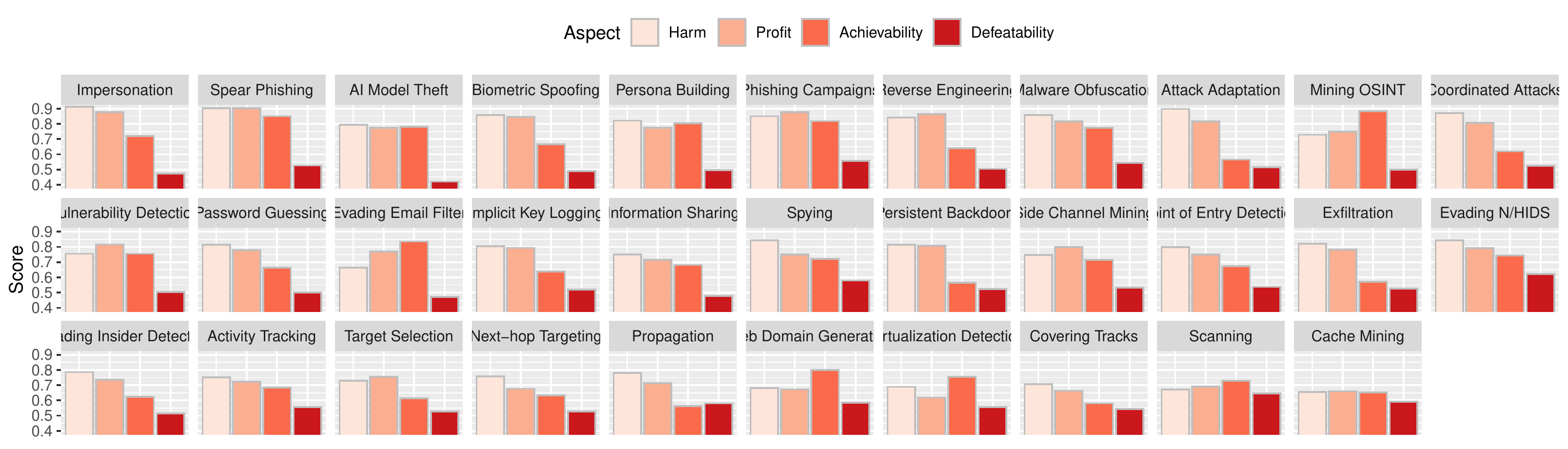}
    \caption{Survey results: the averaged and normalized opinion scores for each offensive AI capability (OAC) when used against an organization. The OACs are ordered according to their threat score, left to right starting from the first row.}
    \label{fig:threatrank_HPAD}
    \vspace{-1em}
\end{figure*}

\subsection{Threat Ranking}
In this section we measure and rank the various threats of an adversary which can utilize or exploit AI technologies to enhance their attacks. For each OAC the participants were asked to rate four aspects on the range of 1-7 (low to high):
\begin{description}[leftmargin=.5cm]
    \item[Profit ($P$):] The amount of benefit which a threat agent gains by using AI compared to using non-AI methods. For example, attack success, flexibility, coverage, automation, and persistence. Here profit assumes that the AI tool has already been implemented. 
    \item[Achievability ($A$):] How easy is it for the attacker to use AI for this task considering that the adversary must implement, train, test and deploy the AI.
    \item[Defeatability ($D$):] How easy is it for the defender to detect or prevent the AI-based attack. Here, a higher score is bad for the adversary (1=hard to defeat, 7=easy to defeat).
    \item[Harm ($H$):] The amount of harm which an AI-capable adversary can inflict in terms of physical, physiological, or monetary damage (including effort put into mitigating the attack).
\end{description}

We say that an adversary is motivated to perform an attack if there is high profit $P$ and high achievability $A$. Moreover, if there is high $P$ but low $A$ or vice versa, some actors may be tempted to try anyways. Therefore, we model the motivation of using an OAC as $M=\frac{1}{2}(P + A)$. However, just because there is motivation, it does not mean that there is a risk. If the AI attack can be easily detected or prevented, then no amount of motivation will make the OAC a risk. Therefore, we model risk as $R=\frac{M}{D}$ where a low defeatability (hard to prevent) increases $R$ and a high defeatability (easy to prevent) lowers $R$. Risk can also be viewed as the likelihood of the attack occurring, or the likelihood of an attack success. Finally, to model threat, we must consider the amount of harm done to the organization. An OAC with high $R$ but no consequences is less of a threat. Therefore, we model our threat score as
\begin{equation}
    T = H\frac{\frac{1}{2}(P + A)}{D} = H\frac{M}{D} = HR  
\end{equation}
Before computing $T$, we normalize $P$, $A$, $D$, and $H$ from the range 1-7 to 0-1. This way, a threat score greater than 1 indicates a significant threat because for these scores (1) the adversary will attempt the attack ($M>D$), and (2) the level of harm will be greater than the ability to prevent the attack ($\frac{D}{M}< H \le 1$). We can also see from our model that as an adversary's motivation increases over defeatability, the amount of harm deemed threatening decreases. This is intuitive because if an attack is easy to achieve and highly profitable, then it will be performed more often. Therefore, even if it is less harmful, attacks will occur frequently so the damage will be higher in the long run.

\begin{figure}[t]
    \centering
\begin{minipage}{0.48\columnwidth}
    \includegraphics[width=\columnwidth]{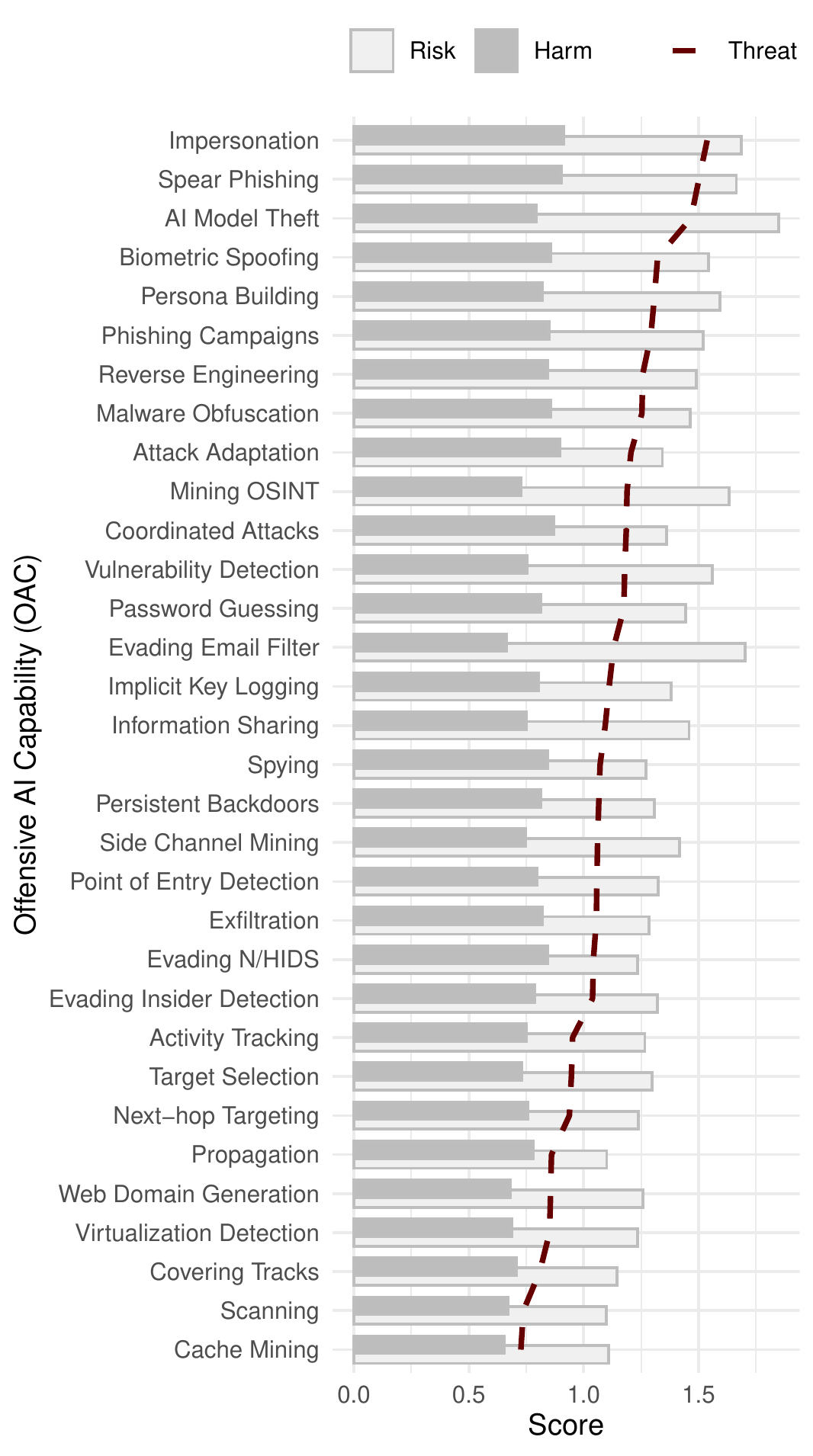}
    \caption{Survey results: the offensive AI capabilities ranked according to their threat scores.}
    \label{fig:threatrank_all}

\end{minipage}
\begin{minipage}{0.48\columnwidth}

\begin{minipage}{\columnwidth}
    \centering
    \includegraphics[width=\columnwidth]{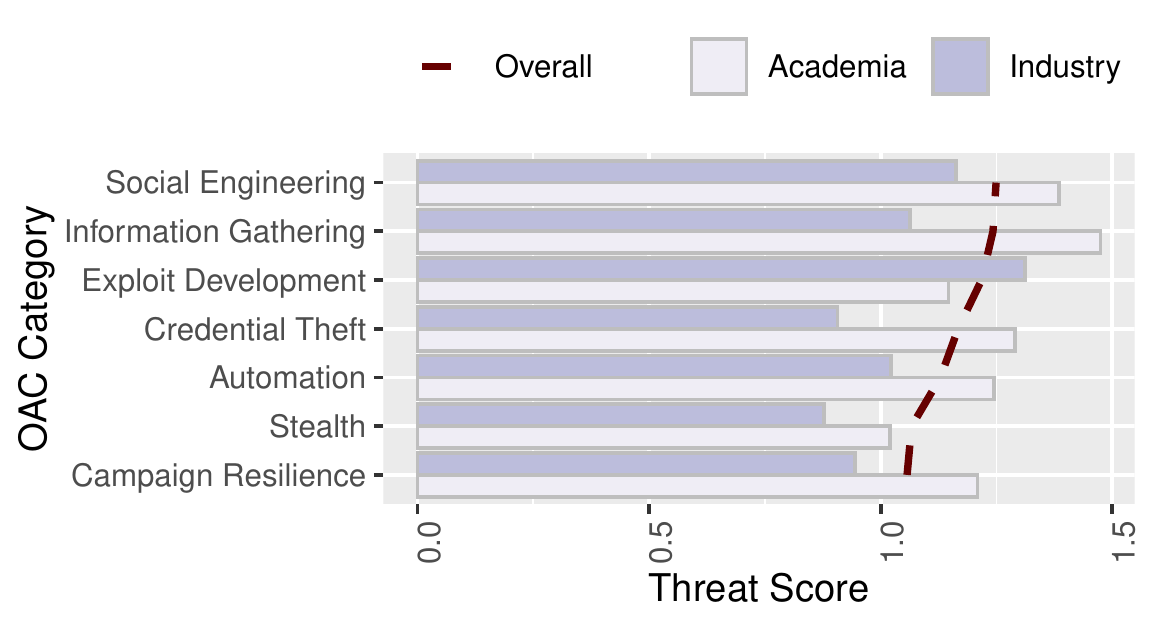}
    \caption{Survey results: the offensive AI capability categories ranked according to their average threat scores. The scores from industry and academia participants are also presented separately.}
    \label{fig:threatrank_categories}
  
\end{minipage}
      \vspace{1em}
\begin{minipage}{\columnwidth}

\end{minipage}

    \centering
    \includegraphics[width=\columnwidth]{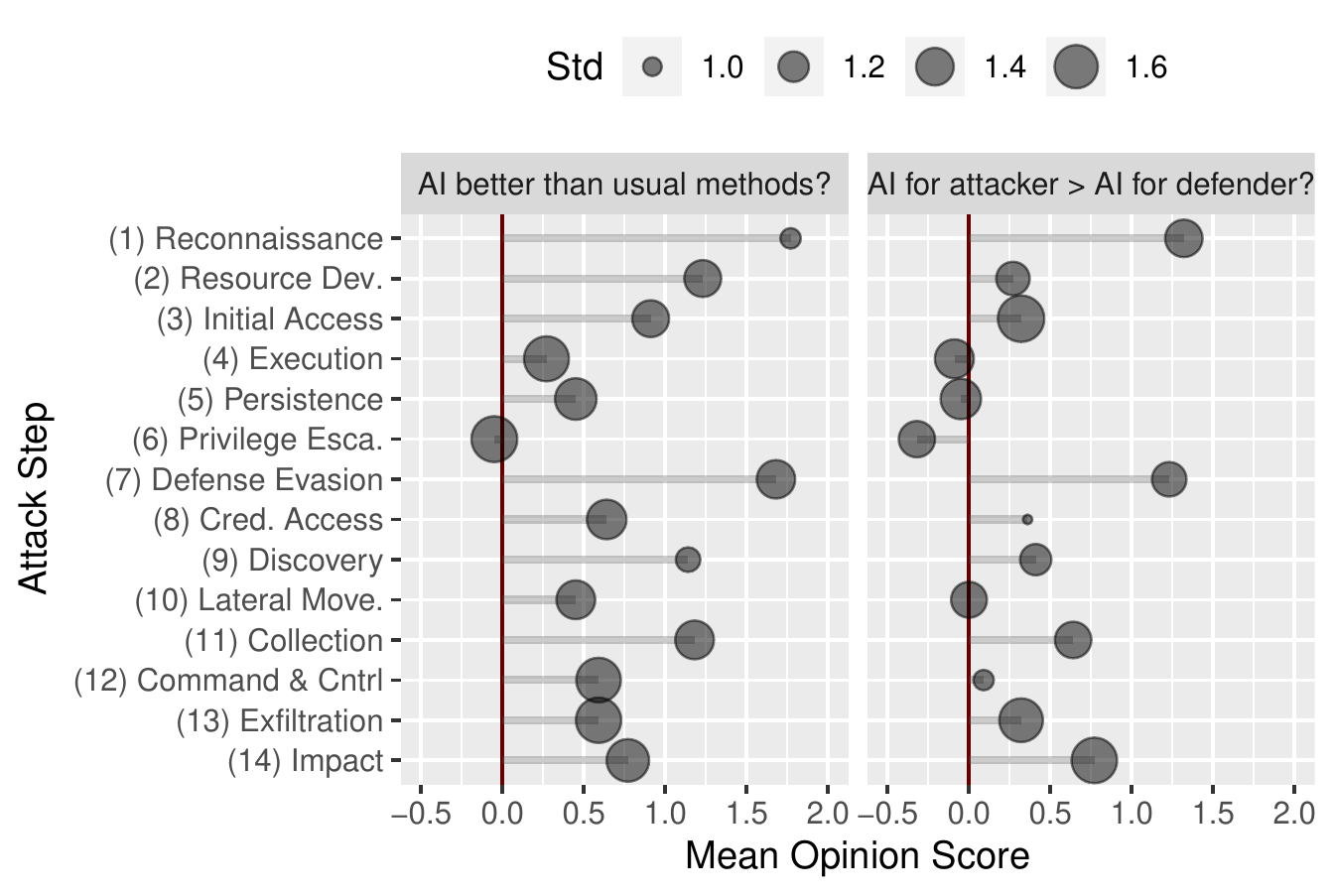}
    \caption{Survey results: Mean opinion scores on whether (1) it is more beneficial for the adversary to use AI over conventional methods, and (2) AI benefits attackers more than AI benefits defenders. The scores range from -3 to +3.}
    \label{fig:ckc_MOS}

\end{minipage}
\end{figure}

\subsubsection{OAC Threat Ranking}
In Fig. \ref{fig:threatrank_HPAD} we present the average $P$, $A$, $D$, and $H$ scores for each OAC. In Fig. \ref{fig:threatrank_all} we present the OACs ranked according to their threat score $T$, and contrast their risk scores $R$ to their harm scores $H$.

The results show that 23 of the OACs (72\%) are considered to be significant threats (have a $T>1$). In general we observe that the top threats mostly relate to social engineering and malware development. The top three OACs are impersonation, spear phishing, and model theft. These OACs have significantly larger threat scores than the others because they are (1) easy to achieve, (2) have high payoffs, (3) are hard to prevent, and (4) cause the most harm (top left of Fig. \ref{fig:threatrank_HPAD}). 
Interestingly, the use of AI to run phishing campaigns is considered a large threat even though it has a relatively high $D$ score. We believe this is because, with AI, an adversary can both increase the number and quality of the phishing attacks. Therefore, even if 99\% of the attempts fail, some will get through and cause the organization damage.
The least significant threats were scanning and cache mining which are perceived to have have little benefit for the adversary because they pose a high risk of detection. Other low ranked threats include some on-site automation for propagation, target selection, lateral movement, and covering tracks.

\subsubsection{Industry vs Academia}
In Fig. \ref{fig:threatrank_categories} we look at the average threat scores for each OAC \textit{category}, and contrast the opinions of members from academia to those from industry. 

In general, academia views AI as a more significant threat to organizations than industry. One can argue that the discrepancy is because industry tends to be more practical and grounded in the present, where academia considers potential threats thus considering the future. For example, when looking at the threat scores from academia, all of the categories are considered significant threats ($T>1$). However, when looking at the industry's responses, the categories of stealth, credential theft, and campaign resilience are not. This may be because these concepts have presented (proven) themselves less in the wild than the others.

Regardless, both industry and academia agree on the top three most threatening OAC categories: (1) exploit development, (2) social engineering, and (3) information gathering. This is because, for these categories, the attacker benefits greatly from using AI ($P$), can easy implement the relevant AI tools ($A$), the attack causes considerable damage ($H$), and there is little the defender can do to prevent them ($D$) (indicated in Fig. \ref{fig:threatrank_HPAD}). For example, deepfakes are easy to implement yet hard to detect in practice (e.g., in a phone call), and extracting private information from side channels and online resources can be accomplished with little intervention.

Surprisingly, both academia and industry consider the use of AI for stealth as the least threatening OAC category in general. Even though there has been a great deal of work showing how IDS models are vulnerable \cite{suciu19-spw,novo20-aisec}, IDS evasion approaches were considered the second most defeatable OAC after intelligent scanning. This may have to do with the fact that the adversary cannot evaluate its AI-based evasion techniques inside the actual network, and thus risks detection.

Overall, there were some disagreements between industry and academia regarding the most threatening OACs. The top-10 most threatening OACs for organizations (out of 33) were ranked as follows:

\noindent\makebox[\textwidth][c]{
\resizebox{0.8\columnwidth}{!}{

\begin{minipage}{.5\columnwidth}

    \vspace{.5em}
  \textbf{Industry's Perspective}
  \begin{enumerate}
        \item Reverse Engineering
        \item Impersonation
        \item AI Model Theft
        \item Spear Phishing
        \item Persona Building
        \item Phishing Campaigns
        \item Information Sharing
        \item Malware Obfuscation
        \item Vulnerability Detection
        \item Password Guessing
    \vspace{.5em}
  \end{enumerate}
\end{minipage}
\begin{minipage}{.5\columnwidth}
\vspace{.5em}
  \textbf{Academia's Perspective}
  \begin{enumerate}
        \item Biometric Spoofing
        \item Impersonation
        \item Spear Phishing
        \item AI Model Theft
        \item Mining OSINT
        \item Spying
        \item Target Selection
        \item Side Channel Mining
        \item Coordinated Attacks
        \item Attack Adaptation
    \end{enumerate}
    
\vspace{.5em}
\end{minipage}}}

We note that academia views biometric spoofing as the top threat, where industry doesn't consider it in their top 10. We think this is because the latest research on this topic involves ML which can be evaded (e.g., \cite{sharif2016accessorize,bontrager2018deepmasterprints}). In contrast to academia, industry views this OAC as less harmful to the organization and less profitable to the adversary, perhaps because biometric security is not a common defense used in organization. Regardless, biometric spoofing is still considered the 4-th highest threat overall (Fig. \ref{fig:threatrank_all}).  
Another insight is that academia is more concerned about the use of ML for spyware, side-channels, target selection, and attack adaptation than industry. This may be because these are topics which have long been discussed in academia, but have yet to cause major disruptions in the real-world. For industry, they are more concerned with the use of AI for exploit development and social engineering, likely because these are threats which are out of their control.  

Additional figures which compare the responses of industry to academia can be found online\footnote{\url{https://tinyurl.com/t735m6st}\label{foot:extra_figs}}.

\subsection{Impact on the Cyber Kill Chain}\label{subsec:survey_ckc}
For each of the 14 MITRE ATT\&CK steps, we asked the participants whether they agree or disagree\footnote{Measured using a 7-step likert scale ranging from strongly disagree (-3) to neutral (0) to strongly agree (+3).} to the following statements: (1) It more beneficial for the attacker to use AI than conventional methods in this attack step, and (2) AI benefits the attacker more than AI benefits the defender. The objective of these questions were to identify how AI impacts the kill chain and whether AI forms any asymmetry between the attacker and defender.

In Fig. \ref{fig:ckc_MOS} we present the mean opinion scores along with their standard deviations (additional histograms can be found online\footref{foot:extra_figs}). Overall, our participants felt that AI enhances the adversary's ability to traverse the kill chain. In particular, we observe that adversary benefits considerably from AI during the first three steps. One explanation is that these attacks are maintained offsite and thus are easier to develop and have less risk.
Moreover, we understand from the results that there is a general feeling that defenders do not have a good way to preventing adversarial machine learning attacks. Therefore, AI not only improves defense evasion but also gives the attacker a considerable advantage over the defender in this regard.

Our participants also felt that an adversary with AI has a somewhat greater advantage over a defender with AI for most attack steps. In particular, the defender cannot effectively utilize AI to prevent reconnaissance except for mitigating a few kinds of social engineering attacks. Moreover, the adversary has many new uses for AI during the impact step, such as the tampering of records, where the defender does not. However, the participants felt that the defender has an advantage when using AI to detect execution, persistence, and privilege escalation. This is understandable since the defender can train and evaluate models onsite whereas the attacker cannot.




\section{Discussion}\label{sec:discussion}
In this section, we share our insights on our findings and discuss the road ahead.

\subsection{Insights, Observations, \& Limitations}

\noindent\textbf{Top Threats.}
It is understandable why the highest ranked threats to organizations relate to social engineering attacks and software analysis (vulnerability detection and reverse engineering). It is because these attacks are out of the defender's control. For example, humans are the weakest link, even with security awareness training. However, with deepfakes, even less can be done to mitigate these social engineering attacks.
The same holds for software analysis where ML has proven itself to work well with languages and even compiled binaries \cite{ye2020misim}. 
As mentioned earlier, we believe the reason academia is the most concerned with biometrics is because it almost exclusively uses ML, and academia is well aware of ML's flaws. On the other hand, industry members know that organizations do not often employ biometric security. Therefore, they perceive AI attacks on their software and personnel as the greatest threats.

\noindent\textbf{The Near Future.}
Over the next few years, we believe that there will be an increase of offensive AI incidents, but only at the front and back of the attack model (recon., resource development, and impact --such as record tampering). This is because currently AI cannot effectively learn on its own. Therefore, we aren't likely to see botnets that can autonomously and dynamically interact with a diverse set of complex systems (like an organization's network) in the near future. Therefore, since modern adversaries have limited information on the organizations' network, they
are restricted to attacks where the data collection, model development, training, and evaluation occur offsite. In particular, we note that DL models are large and require a considerable amount of resources to run. This makes them easy to detect when transferred into the network or executed onsite. However, the model's footprint will become less anomalous over time as DL proliferates. In the near future, we also expect that phishing campaigns will become more rampant and dangerous as humans and bots are given the ability to make convincing deepfake phishing calls.


\noindent\textbf{AI is a Double Edged Sword.}
We observed that AI technologies for security can also be used in an offensive manner. Some technologies are dual purpose. For example, the ML research into disassembly, vulnerability detection, and penetration testing. Some technologies can be repurposed. For example, instead of using explainable AI to validate malware detection, it can be used to hide artifacts. And some technologies can be inverted. For example, an insider detection model can be used to help cover tracks and avoid detection. To help raise awareness, we recommend that researchers note the implications of their work, even for defensive technologies.
One caveat is that the `sword' is not symmetric depending on the wielder. For example, generative AI (deepfakes) is better for the attacker, but anomaly detection is better for the defender.

\subsection{The Industry's Perspective}
Using logic to automate attacks is not new to industry – for instance, in 2015, security researchers from FireEye \cite{intelligence2015hammertoss} found that advanced Russian cyber threat groups built a malware called HAMMERTOSS that used rules based automation to blend its traffic into normal traffic by checking for regular office hours in the time zone and then operating only in that time range. However, the scale and speed that offensive AI capabilities can endow attackers can be damaging. 

According to 2019 Verizon Data Breach report analysis of 140 security breaches \cite{dbir2019verizon}, the meantime to compromising an organization and exfiltrating the data ranges is already in the order of minutes. Organizations are already finding it difficult to combat automated offensive tactics and anticipate attacks to get stealthier in the future. For instance, according to the final report released by the US National Security Commission on AI in 2021 \cite{report2021nscai},  the warning is clear ``The U.S. government is not prepared to defend the United States in the coming artificial intelligence (AI) era.'' The final report reasons that this is ``Because of AI, adversaries will be able to act with micro-precision, but at macro-scale and with greater speed. They will use AI to enhance cyber attacks and digital disinformation campaigns and to target individuals in new ways.'' 

Most organizations see offensive AI as an imminent threat – 49\% of 102 cybersecurity organizations surveyed by Forrester market research in 2020\cite{report2020forrester}, anticipate offensive AI techniques to manifest in the next 12 months. As a result, more organizations are turning to ways to defend against these attacks. A 2021 survey \cite{report2021MIT} of 309 organizations’ business leaders, C-Suite executives found that 96\% of the organizations surveyed are already making investments to guard against AI-powered attacks as they anticipate more automation than what their defenses can handle.

\subsection{What's on the Horizon}
With AI's rapid pace of development and open accessibility, we expect to see a noticeable shift in attack strategies on organizations. First, we foresee that the number of deepfake phishing incidents will increase. This is because the technology (1) is mature, (2) is harder to mitigate than regular phishing, (3) is more effective at exploiting trust, (4) can expedite attacks, and (5) is new as phishing tactic so people are not expecting it. Second, we expect that AI will enable adversaries to target more organizations in parallel and more frequently. As a result, instead of being covert, adversaries may chose to overwhelm the defender's response teams with thousands of attempts for the chance of one success. Finally, as adversaries begin to use AI-enabled bots, defenders will be forced to automate their defences with bots as well. Keeping humans in the loop to control and determine high level strategies is a practical and ethical requirement. However, further discussion and research is necessary to form safe and agreeable policies.


\subsection{What can be done?}
\noindent\textbf{Attacks Using AI.} Industry and academia should focus on developing solutions for mitigating the top threats. Personnel can be shown what to expect from AI-powered social engineering and further research can be done on detecting deepfakes, but in a manner which is robust to a dynamic adversary \cite{mirsky2021creation}. Moreover, we recommend research into post-processing tools that can protect software from analysis after development (i.e., anti-vulnerability detection).

\noindent\textbf{Attacks Against AI.} The advantages and vulnerabilities of AI have profoundly questioned their widespread adoption, especially in mission-critical and cybersecurity-related tasks. In the meantime, organizations are working on automating the development and operations of ML models (MLOps), without focusing too much on ML security-related issues. To bridge this gap, we argue that extending the current MLOps paradigm to also encompass ML security (MLSecOps) may be a relevant way towards improving the security posture of such organizations. To this end, we envision the incorporation of security testing, protection and monitoring of AI/ML models into MLOps. Doing so will enable organizations to seamlessly deploy and maintain more secure and reliable AI/ML models.



\section{Conclusion}\label{sec:conclusion}
In this survey we first explored, categorized, and identified the threats of offensive AI against organizations (sections \ref{sec:background} and \ref{sec:attack_model}). We then detailed the threats and ranked them through a user study with experts from the domain (sections \ref{sec:ai_vs_ckc} and \ref{sec:user_study}). Finally, we provided insights into our results and gave directions for future work (section \ref{sec:discussion}). We hope this survey will be meaningful and helpful to the community in addressing the imminent threat of offensive AI.

\section{Acknowledgments}\label{sec:ak}
The authors would like to thank Laurynas Adomaitis, Sin G. Teo, Manojkumar Parmar, Charles Hart, Matilda Rhode, Dr. Daniele Sgandurra, Dr. Pin-Yu Chen, Evan Downing, and Didier Contis for taking the time to participate in our survey. We note that the views reflect the participant's personal experiences and does not reflect the view of the participant's employer. This material is based upon work supported by the Zuckerman STEM Leadership Program.

{\begin{spacing}{.915}
\bibliographystyle{ACM-Reference-Format}
    \bibliography{bib.bib}\end{spacing}}


\begin{thebibliography}{255}


\ifx \showCODEN    \undefined \def \showCODEN     #1{\unskip}     \fi
\ifx \showDOI      \undefined \def \showDOI       #1{#1}\fi
\ifx \showISBNx    \undefined \def \showISBNx     #1{\unskip}     \fi
\ifx \showISBNxiii \undefined \def \showISBNxiii  #1{\unskip}     \fi
\ifx \showISSN     \undefined \def \showISSN      #1{\unskip}     \fi
\ifx \showLCCN     \undefined \def \showLCCN      #1{\unskip}     \fi
\ifx \shownote     \undefined \def \shownote      #1{#1}          \fi
\ifx \showarticletitle \undefined \def \showarticletitle #1{#1}   \fi
\ifx \showURL      \undefined \def \showURL       {\relax}        \fi
\providecommand\bibfield[2]{#2}
\providecommand\bibinfo[2]{#2}
\providecommand\natexlab[1]{#1}
\providecommand\showeprint[2][]{arXiv:#2}

\bibitem[\protect\citeauthoryear{??}{bla}{[n.d.]}]%
        {blackhat18-socialmapper}
 \bibinfo{year}{[n.d.]}\natexlab{}.
\newblock \bibinfo{title}{Black {Hat} {USA} 2018}.
\newblock
\newblock
\urldef\tempurl%
\url{https://www.blackhat.com/us-18/arsenal.html#social-mapper-social-media-correlation-through-facial-recognition}
\showURL{%
\tempurl}


\bibitem[\protect\citeauthoryear{??}{git}{[n.d.]}]%
        {github_telegram_contest}
 \bibinfo{year}{[n.d.]}\natexlab{}.
\newblock \bibinfo{title}{Telegram Contest}.
\newblock \bibinfo{howpublished}{{https://github.com/IlyaGusev/tgcontest}}.
\newblock
\newblock
\shownote{(Accessed on 10/14/2020).}


\bibitem[\protect\citeauthoryear{??}{dbi}{2019}]%
        {dbir2019verizon}
 \bibinfo{year}{2019}\natexlab{}.
\newblock \showarticletitle{2019 Data Breach Investigations Report}.
\newblock \bibinfo{journal}{\emph{Verizon, Inc}} (\bibinfo{year}{2019}).
\newblock


\bibitem[\protect\citeauthoryear{??}{rep}{2020}]%
        {report2020forrester}
 \bibinfo{year}{2020}\natexlab{}.
\newblock \showarticletitle{The Emergence of Offensive AI}.
\newblock \bibinfo{journal}{\emph{Forrester}} (\bibinfo{year}{2020}).
\newblock


\bibitem[\protect\citeauthoryear{??}{cro}{2020}]%
        {crowdstrike_2020}
 \bibinfo{year}{2020}\natexlab{}.
\newblock \bibinfo{title}{Our Work with the DNC: Setting the record straight}.
\newblock
\newblock
\urldef\tempurl%
\url{https://www.crowdstrike.com/blog/bears-midst-intrusion-democratic-national-committee/}
\showURL{%
\tempurl}


\bibitem[\protect\citeauthoryear{??}{272}{2021}]%
        {272248}
 \bibinfo{year}{2021}\natexlab{}.
\newblock \showarticletitle{DeepReflect: Discovering Malicious Functionality
  through Binary Reconstruction}. In \bibinfo{booktitle}{\emph{30th {USENIX}
  Security Symposium ({USENIX} Security 21)}}. \bibinfo{publisher}{{USENIX}
  Association}.
\newblock
\urldef\tempurl%
\url{https://www.usenix.org/conference/usenixsecurity21/presentation/downing}
\showURL{%
\tempurl}


\bibitem[\protect\citeauthoryear{??}{rep}{2021a}]%
        {report2021nscai}
 \bibinfo{year}{2021}\natexlab{a}.
\newblock \showarticletitle{Final Report - National Security Commission on
  Artificial Intelligence}.
\newblock \bibinfo{journal}{\emph{National Security Commission on Artificial
  Intelligence}} (\bibinfo{year}{2021}).
\newblock


\bibitem[\protect\citeauthoryear{??}{rep}{2021b}]%
        {report2021MIT}
 \bibinfo{year}{2021}\natexlab{b}.
\newblock \showarticletitle{Preparing for AI-enabled cyberattacks}.
\newblock \bibinfo{journal}{\emph{MIT Technology Review Insights}}
  (\bibinfo{year}{2021}).
\newblock


\bibitem[\protect\citeauthoryear{Abadi and Andersen}{Abadi and
  Andersen}{2016}]%
        {abadi16-arxiv}
\bibfield{author}{\bibinfo{person}{Martín Abadi} {and}
  \bibinfo{person}{David~G. Andersen}.} \bibinfo{year}{2016}\natexlab{}.
\newblock \showarticletitle{Learning to {Protect} {Communications} with
  {Adversarial} {Neural} {Cryptography}}.
\newblock \bibinfo{journal}{\emph{arXiv}} (\bibinfo{year}{2016}).
\newblock
\urldef\tempurl%
\url{https://arxiv.org/abs/1610.06918}
\showURL{%
\tempurl}


\bibitem[\protect\citeauthoryear{{Abd El-Jawad}, {Hodhod}, and {Omar}}{{Abd
  El-Jawad} et~al\mbox{.}}{2018}]%
        {8636124}
\bibfield{author}{\bibinfo{person}{M.~H. {Abd El-Jawad}}, \bibinfo{person}{R.
  {Hodhod}}, {and} \bibinfo{person}{Y.~M.~K. {Omar}}.}
  \bibinfo{year}{2018}\natexlab{}.
\newblock \showarticletitle{Sentiment Analysis of Social Media Networks Using
  Machine Learning}. In \bibinfo{booktitle}{\emph{2018 14th International
  Computer Engineering Conference (ICENCO)}}. \bibinfo{pages}{174--176}.
\newblock
\urldef\tempurl%
\url{https://doi.org/10.1109/ICENCO.2018.8636124}
\showDOI{\tempurl}


\bibitem[\protect\citeauthoryear{Abid, Imine, and Rusinowitch}{Abid
  et~al\mbox{.}}{2018}]%
        {Abid18-EDBT-ICDT}
\bibfield{author}{\bibinfo{person}{Y. Abid}, \bibinfo{person}{Abdessamad
  Imine}, {and} \bibinfo{person}{Micha{\"e}l Rusinowitch}.}
  \bibinfo{year}{2018}\natexlab{}.
\newblock \showarticletitle{Sensitive Attribute Prediction for Social Networks
  Users}. In \bibinfo{booktitle}{\emph{EDBT/ICDT Workshops}}.
\newblock


\bibitem[\protect\citeauthoryear{Aghakhani, Meng, Wang, Kruegel, and
  Vigna}{Aghakhani et~al\mbox{.}}{2020}]%
        {aghakhani20-bullseye}
\bibfield{author}{\bibinfo{person}{Hojjat Aghakhani}, \bibinfo{person}{Dongyu
  Meng}, \bibinfo{person}{Yu-Xiang Wang}, \bibinfo{person}{Christopher
  Kruegel}, {and} \bibinfo{person}{Giovanni Vigna}.}
  \bibinfo{year}{2020}\natexlab{}.
\newblock \showarticletitle{Bullseye {Polytope}: {A} {Scalable} {Clean}-{Label}
  {Poisoning} {Attack} with {Improved} {Transferability}}.
\newblock \bibinfo{journal}{\emph{arXiv preprint arXiv:2005.00191}}
  (\bibinfo{year}{2020}).
\newblock


\bibitem[\protect\citeauthoryear{Akoglu, Tong, and Koutra}{Akoglu
  et~al\mbox{.}}{2015}]%
        {akoglu2015graph}
\bibfield{author}{\bibinfo{person}{Leman Akoglu}, \bibinfo{person}{Hanghang
  Tong}, {and} \bibinfo{person}{Danai Koutra}.}
  \bibinfo{year}{2015}\natexlab{}.
\newblock \showarticletitle{Graph based anomaly detection and description: a
  survey}.
\newblock \bibinfo{journal}{\emph{Data mining and knowledge discovery}}
  \bibinfo{volume}{29}, \bibinfo{number}{3} (\bibinfo{year}{2015}),
  \bibinfo{pages}{626--688}.
\newblock


\bibitem[\protect\citeauthoryear{Al-Hababi and Tokgoz}{Al-Hababi and
  Tokgoz}{2020}]%
        {al2020man}
\bibfield{author}{\bibinfo{person}{Abdulrahman Al-Hababi} {and}
  \bibinfo{person}{Sezer~C Tokgoz}.} \bibinfo{year}{2020}\natexlab{}.
\newblock \showarticletitle{Man-in-the-Middle Attacks to Detect and Identify
  Services in Encrypted Network Flows using Machine Learning}. In
  \bibinfo{booktitle}{\emph{2020 3rd International Conference on Advanced
  Communication Technologies and Networking (CommNet)}}. IEEE,
  \bibinfo{pages}{1--5}.
\newblock


\bibitem[\protect\citeauthoryear{Alshamrani, Myneni, Chowdhary, and
  Huang}{Alshamrani et~al\mbox{.}}{2019}]%
        {alshamrani2019survey}
\bibfield{author}{\bibinfo{person}{Adel Alshamrani}, \bibinfo{person}{Sowmya
  Myneni}, \bibinfo{person}{Ankur Chowdhary}, {and} \bibinfo{person}{Dijiang
  Huang}.} \bibinfo{year}{2019}\natexlab{}.
\newblock \showarticletitle{A survey on advanced persistent threats:
  Techniques, solutions, challenges, and research opportunities}.
\newblock \bibinfo{journal}{\emph{IEEE Communications Surveys \& Tutorials}}
  \bibinfo{volume}{21}, \bibinfo{number}{2} (\bibinfo{year}{2019}),
  \bibinfo{pages}{1851--1877}.
\newblock


\bibitem[\protect\citeauthoryear{Anderson}{Anderson}{2017}]%
        {Anderson2017EvadingML}
\bibfield{author}{\bibinfo{person}{H. Anderson}.}
  \bibinfo{year}{2017}\natexlab{}.
\newblock \showarticletitle{Evading Machine Learning Malware Detection}.
\newblock


\bibitem[\protect\citeauthoryear{Anderson, Kharkar, Filar, Evans, and
  Roth}{Anderson et~al\mbox{.}}{2018}]%
        {anderson18-arxiv}
\bibfield{author}{\bibinfo{person}{Hyrum~S. Anderson}, \bibinfo{person}{Anant
  Kharkar}, \bibinfo{person}{Bobby Filar}, \bibinfo{person}{David Evans}, {and}
  \bibinfo{person}{Phil Roth}.} \bibinfo{year}{2018}\natexlab{}.
\newblock \bibinfo{title}{Learning to Evade Static PE Machine Learning Malware
  Models via Reinforcement Learning}.
\newblock
\newblock
\showeprint[arxiv]{1801.08917}~[cs.CR]


\bibitem[\protect\citeauthoryear{Arsene}{Arsene}{2020}]%
        {arsene2020oil}
\bibfield{author}{\bibinfo{person}{L. Arsene}.}
  \bibinfo{year}{2020}\natexlab{}.
\newblock \bibinfo{title}{Oil \& Gas Spearphishing Campaigns Drop Agent Tesla
  Spyware in Advance of Historic OPEC+ Deal}.
\newblock
\newblock
\urldef\tempurl%
\url{https://labs.bitdefender.com/2020/04/oil-gas-spearphishing-campaigns-drop-agent-tesla-spyware-in-advance-of-historic-opec-deal/}
\showURL{%
\tempurl}


\bibitem[\protect\citeauthoryear{Atlidakis, Geambasu, Godefroid, Polishchuk,
  and Ray}{Atlidakis et~al\mbox{.}}{2020}]%
        {atlidakis2020pythia}
\bibfield{author}{\bibinfo{person}{Vaggelis Atlidakis}, \bibinfo{person}{Roxana
  Geambasu}, \bibinfo{person}{Patrice Godefroid}, \bibinfo{person}{Marina
  Polishchuk}, {and} \bibinfo{person}{Baishakhi Ray}.}
  \bibinfo{year}{2020}\natexlab{}.
\newblock \showarticletitle{Pythia: Grammar-Based Fuzzing of REST APIs with
  Coverage-guided Feedback and Learning-based Mutations}.
\newblock \bibinfo{journal}{\emph{arXiv preprint arXiv:2005.11498}}
  (\bibinfo{year}{2020}).
\newblock


\bibitem[\protect\citeauthoryear{Bagdasaryan and Shmatikov}{Bagdasaryan and
  Shmatikov}{2020}]%
        {Bagdasaryan20}
\bibfield{author}{\bibinfo{person}{Eugene Bagdasaryan} {and}
  \bibinfo{person}{Vitaly Shmatikov}.} \bibinfo{year}{2020}\natexlab{}.
\newblock \bibinfo{title}{Blind Backdoors in Deep Learning Models}.
\newblock
\newblock


\bibitem[\protect\citeauthoryear{Bahnsen, Torroledo, Camacho, and
  Villegas}{Bahnsen et~al\mbox{.}}{2018}]%
        {bahnsen2018deepphish}
\bibfield{author}{\bibinfo{person}{Alejandro~Correa Bahnsen},
  \bibinfo{person}{Ivan Torroledo}, \bibinfo{person}{Luis~David Camacho}, {and}
  \bibinfo{person}{Sergio Villegas}.} \bibinfo{year}{2018}\natexlab{}.
\newblock \showarticletitle{DeepPhish: Simulating Malicious AI}. In
  \bibinfo{booktitle}{\emph{2018 APWG Symposium on Electronic Crime Research
  (eCrime)}}. \bibinfo{pages}{1--8}.
\newblock


\bibitem[\protect\citeauthoryear{Balagani, Conti, Gasti, Georgiev, Gurtler,
  Lain, Miller, Molas, Samarin, Saraci, et~al\mbox{.}}{Balagani
  et~al\mbox{.}}{2018}]%
        {balagani2018silk}
\bibfield{author}{\bibinfo{person}{Kiran~S Balagani}, \bibinfo{person}{Mauro
  Conti}, \bibinfo{person}{Paolo Gasti}, \bibinfo{person}{Martin Georgiev},
  \bibinfo{person}{Tristan Gurtler}, \bibinfo{person}{Daniele Lain},
  \bibinfo{person}{Charissa Miller}, \bibinfo{person}{Kendall Molas},
  \bibinfo{person}{Nikita Samarin}, \bibinfo{person}{Eugen Saraci},
  {et~al\mbox{.}}} \bibinfo{year}{2018}\natexlab{}.
\newblock \showarticletitle{Silk-tv: Secret information leakage from keystroke
  timing videos}. In \bibinfo{booktitle}{\emph{European Symposium on Research
  in Computer Security}}. Springer, \bibinfo{pages}{263--280}.
\newblock


\bibitem[\protect\citeauthoryear{Bao, Burket, Woo, Turner, and Brumley}{Bao
  et~al\mbox{.}}{2014}]%
        {bao2014byteweight}
\bibfield{author}{\bibinfo{person}{Tiffany Bao}, \bibinfo{person}{Jonathan
  Burket}, \bibinfo{person}{Maverick Woo}, \bibinfo{person}{Rafael Turner},
  {and} \bibinfo{person}{David Brumley}.} \bibinfo{year}{2014}\natexlab{}.
\newblock \showarticletitle{$\{$BYTEWEIGHT$\}$: Learning to recognize functions
  in binary code}. In \bibinfo{booktitle}{\emph{23rd $\{$USENIX$\}$ Security
  Symposium ($\{$USENIX$\}$ Security 14)}}. \bibinfo{pages}{845--860}.
\newblock


\bibitem[\protect\citeauthoryear{Barreno, Nelson, Joseph, and Tygar}{Barreno
  et~al\mbox{.}}{2010}]%
        {barreno10}
\bibfield{author}{\bibinfo{person}{Marco Barreno}, \bibinfo{person}{Blaine
  Nelson}, \bibinfo{person}{Anthony Joseph}, {and} \bibinfo{person}{J. Tygar}.}
  \bibinfo{year}{2010}\natexlab{}.
\newblock \showarticletitle{The security of machine learning}.
\newblock \bibinfo{journal}{\emph{Machine Learning}}  \bibinfo{volume}{81}
  (\bibinfo{year}{2010}), \bibinfo{pages}{121--148}.
\newblock


\bibitem[\protect\citeauthoryear{Batina, Bhasin, Jap, and Picek}{Batina
  et~al\mbox{.}}{2019}]%
        {Batina19-usenix}
\bibfield{author}{\bibinfo{person}{Lejla Batina}, \bibinfo{person}{Shivam
  Bhasin}, \bibinfo{person}{Dirmanto Jap}, {and} \bibinfo{person}{Stjepan
  Picek}.} \bibinfo{year}{2019}\natexlab{}.
\newblock \showarticletitle{{CSI} {NN}: Reverse Engineering of Neural Network
  Architectures Through Electromagnetic Side Channel}. In
  \bibinfo{booktitle}{\emph{28th {USENIX} Security Symposium ({USENIX} Security
  19)}}. \bibinfo{publisher}{{USENIX} Association}, \bibinfo{address}{Santa
  Clara, CA}, \bibinfo{pages}{515--532}.
\newblock
\showISBNx{978-1-939133-06-9}
\urldef\tempurl%
\url{https://www.usenix.org/conference/usenixsecurity19/presentation/batina}
\showURL{%
\tempurl}


\bibitem[\protect\citeauthoryear{Beni}{Beni}{2020}]%
        {beni2020swarm}
\bibfield{author}{\bibinfo{person}{Gerardo Beni}.}
  \bibinfo{year}{2020}\natexlab{}.
\newblock \showarticletitle{Swarm intelligence}.
\newblock \bibinfo{journal}{\emph{Complex Social and Behavioral Systems: Game
  Theory and Agent-Based Models}} (\bibinfo{year}{2020}),
  \bibinfo{pages}{791--818}.
\newblock


\bibitem[\protect\citeauthoryear{Biggio, Corona, Fumera, Giacinto, and
  Roli}{Biggio et~al\mbox{.}}{2011}]%
        {biggio11-mcs}
\bibfield{author}{\bibinfo{person}{Battista Biggio}, \bibinfo{person}{Igino
  Corona}, \bibinfo{person}{Giorgio Fumera}, \bibinfo{person}{Giorgio
  Giacinto}, {and} \bibinfo{person}{Fabio Roli}.}
  \bibinfo{year}{2011}\natexlab{}.
\newblock \showarticletitle{Bagging Classifiers for Fighting Poisoning Attacks
  in Adversarial Classification Tasks}. In \bibinfo{booktitle}{\emph{10th
  International Workshop on Multiple Classifier Systems (MCS)}}
  \emph{(\bibinfo{series}{Lecture Notes in Computer Science},
  Vol.~\bibinfo{volume}{6713})}, \bibfield{editor}{\bibinfo{person}{Carlo
  Sansone}, \bibinfo{person}{Josef Kittler}, {and} \bibinfo{person}{Fabio
  Roli}} (Eds.). \bibinfo{publisher}{Springer-Verlag},
  \bibinfo{pages}{350--359}.
\newblock


\bibitem[\protect\citeauthoryear{Biggio, Corona, Maiorca, Nelson,
  \v{S}rndi\'{c}, Laskov, Giacinto, and Roli}{Biggio et~al\mbox{.}}{2013}]%
        {biggio13-ecml}
\bibfield{author}{\bibinfo{person}{B. Biggio}, \bibinfo{person}{I. Corona},
  \bibinfo{person}{D. Maiorca}, \bibinfo{person}{B. Nelson},
  \bibinfo{person}{N. \v{S}rndi\'{c}}, \bibinfo{person}{P. Laskov},
  \bibinfo{person}{G. Giacinto}, {and} \bibinfo{person}{F. Roli}.}
  \bibinfo{year}{2013}\natexlab{}.
\newblock \showarticletitle{Evasion attacks against machine learning at test
  time}. In \bibinfo{booktitle}{\emph{Machine Learning and Knowledge Discovery
  in Databases (ECML PKDD), Part III}} \emph{(\bibinfo{series}{LNCS},
  Vol.~\bibinfo{volume}{8190})}, \bibfield{editor}{\bibinfo{person}{Hendrik
  Blockeel}, \bibinfo{person}{Kristian Kersting}, \bibinfo{person}{Siegfried
  Nijssen}, {and} \bibinfo{person}{Filip \v{Z}elezn\'{y}}} (Eds.).
  \bibinfo{publisher}{Springer Berlin Heidelberg}, \bibinfo{pages}{387--402}.
\newblock


\bibitem[\protect\citeauthoryear{Biggio, Nelson, and Laskov}{Biggio
  et~al\mbox{.}}{2012}]%
        {biggio12-icml}
\bibfield{author}{\bibinfo{person}{Battista Biggio}, \bibinfo{person}{Blaine
  Nelson}, {and} \bibinfo{person}{Pavel Laskov}.}
  \bibinfo{year}{2012}\natexlab{}.
\newblock \showarticletitle{Poisoning attacks against support vector machines},
  In \bibinfo{booktitle}{29th Int'l Conf. on Machine Learning},
  \bibfield{editor}{\bibinfo{person}{John Langford} {and}
  \bibinfo{person}{Joelle Pineau}} (Eds.).
\newblock \bibinfo{journal}{\emph{Int'l Conf. on Machine Learning (ICML)}},
  \bibinfo{pages}{1807--1814}.
\newblock


\bibitem[\protect\citeauthoryear{Biggio and Roli}{Biggio and Roli}{2018}]%
        {biggio18-pr}
\bibfield{author}{\bibinfo{person}{B. Biggio} {and} \bibinfo{person}{F. Roli}.}
  \bibinfo{year}{2018}\natexlab{}.
\newblock \showarticletitle{Wild {Patterns}: {Ten} {Years} {After} the {Rise}
  of {Adversarial} {Machine} {Learning}}.
\newblock \bibinfo{journal}{\emph{Pattern Recognition}}  \bibinfo{volume}{84}
  (\bibinfo{year}{2018}), \bibinfo{pages}{317--331}.
\newblock


\bibitem[\protect\citeauthoryear{Bitton, Gluck, Stan, Inokuchi, Ohta, Yamada,
  Yagyu, Elovici, and Shabtai}{Bitton et~al\mbox{.}}{2018}]%
        {bitton2018deriving}
\bibfield{author}{\bibinfo{person}{Ron Bitton}, \bibinfo{person}{Tomer Gluck},
  \bibinfo{person}{Orly Stan}, \bibinfo{person}{Masaki Inokuchi},
  \bibinfo{person}{Yoshinobu Ohta}, \bibinfo{person}{Yoshiyuki Yamada},
  \bibinfo{person}{Tomohiko Yagyu}, \bibinfo{person}{Yuval Elovici}, {and}
  \bibinfo{person}{Asaf Shabtai}.} \bibinfo{year}{2018}\natexlab{}.
\newblock \showarticletitle{Deriving a cost-effective digital twin of an ICS to
  facilitate security evaluation}. In \bibinfo{booktitle}{\emph{European
  Symposium on Research in Computer Security}}. Springer,
  \bibinfo{pages}{533--554}.
\newblock


\bibitem[\protect\citeauthoryear{Bland, Petty, Whitaker, Maxwell, and
  Cantrell}{Bland et~al\mbox{.}}{2020}]%
        {bland20-cs}
\bibfield{author}{\bibinfo{person}{John~A. Bland}, \bibinfo{person}{Mikel~D.
  Petty}, \bibinfo{person}{Tymaine~S. Whitaker}, \bibinfo{person}{Katia~P.
  Maxwell}, {and} \bibinfo{person}{Walter~Alan Cantrell}.}
  \bibinfo{year}{2020}\natexlab{}.
\newblock \showarticletitle{Machine Learning Cyberattack and Defense
  Strategies}.
\newblock \bibinfo{journal}{\emph{Computers \& Security}}  \bibinfo{volume}{92}
  (\bibinfo{year}{2020}), \bibinfo{pages}{101738}.
\newblock
\showISSN{0167-4048}
\urldef\tempurl%
\url{https://doi.org/10.1016/j.cose.2020.101738}
\showDOI{\tempurl}


\bibitem[\protect\citeauthoryear{Bontrager, Roy, Togelius, Memon, and
  Ross}{Bontrager et~al\mbox{.}}{2018}]%
        {bontrager2018deepmasterprints}
\bibfield{author}{\bibinfo{person}{Philip Bontrager}, \bibinfo{person}{Aditi
  Roy}, \bibinfo{person}{Julian Togelius}, \bibinfo{person}{Nasir Memon}, {and}
  \bibinfo{person}{Arun Ross}.} \bibinfo{year}{2018}\natexlab{}.
\newblock \showarticletitle{Deepmasterprints: Generating masterprints for
  dictionary attacks via latent variable evolution}. In
  \bibinfo{booktitle}{\emph{2018 IEEE 9th International Conference on
  Biometrics Theory, Applications and Systems (BTAS)}}. IEEE,
  \bibinfo{pages}{1--9}.
\newblock


\bibitem[\protect\citeauthoryear{Bossert, Guih{\'e}ry, and Hiet}{Bossert
  et~al\mbox{.}}{2014}]%
        {bossert2014towards}
\bibfield{author}{\bibinfo{person}{Georges Bossert},
  \bibinfo{person}{Fr{\'e}d{\'e}ric Guih{\'e}ry}, {and}
  \bibinfo{person}{Guillaume Hiet}.} \bibinfo{year}{2014}\natexlab{}.
\newblock \showarticletitle{Towards automated protocol reverse engineering
  using semantic information}. In \bibinfo{booktitle}{\emph{Proceedings of the
  9th ACM symposium on Information, computer and communications security}}.
  \bibinfo{pages}{51--62}.
\newblock


\bibitem[\protect\citeauthoryear{Breier, Hou, Jap, Ma, Bhasin, and Liu}{Breier
  et~al\mbox{.}}{2018a}]%
        {breier2018deeplaser}
\bibfield{author}{\bibinfo{person}{Jakub Breier}, \bibinfo{person}{Xiaolu Hou},
  \bibinfo{person}{Dirmanto Jap}, \bibinfo{person}{Lei Ma},
  \bibinfo{person}{Shivam Bhasin}, {and} \bibinfo{person}{Yang Liu}.}
  \bibinfo{year}{2018}\natexlab{a}.
\newblock \showarticletitle{Deeplaser: Practical fault attack on deep neural
  networks}.
\newblock \bibinfo{journal}{\emph{arXiv preprint arXiv:1806.05859}}
  (\bibinfo{year}{2018}).
\newblock


\bibitem[\protect\citeauthoryear{Breier, Hou, Jap, Ma, Bhasin, and Liu}{Breier
  et~al\mbox{.}}{2018b}]%
        {breier2018practical}
\bibfield{author}{\bibinfo{person}{Jakub Breier}, \bibinfo{person}{Xiaolu Hou},
  \bibinfo{person}{Dirmanto Jap}, \bibinfo{person}{Lei Ma},
  \bibinfo{person}{Shivam Bhasin}, {and} \bibinfo{person}{Yang Liu}.}
  \bibinfo{year}{2018}\natexlab{b}.
\newblock \showarticletitle{Practical fault attack on deep neural networks}. In
  \bibinfo{booktitle}{\emph{Proceedings of the 2018 ACM SIGSAC Conference on
  Computer and Communications Security}}. \bibinfo{pages}{2204--2206}.
\newblock


\bibitem[\protect\citeauthoryear{Breier, Jap, Hou, Bhasin, and Liu}{Breier
  et~al\mbox{.}}{2020}]%
        {breier2020sniff}
\bibfield{author}{\bibinfo{person}{Jakub Breier}, \bibinfo{person}{Dirmanto
  Jap}, \bibinfo{person}{Xiaolu Hou}, \bibinfo{person}{Shivam Bhasin}, {and}
  \bibinfo{person}{Yang Liu}.} \bibinfo{year}{2020}\natexlab{}.
\newblock \showarticletitle{SNIFF: Reverse Engineering of Neural Networks with
  Fault Attacks}.
\newblock \bibinfo{journal}{\emph{arXiv preprint arXiv:2002.11021}}
  (\bibinfo{year}{2020}).
\newblock


\bibitem[\protect\citeauthoryear{Brumaghin, Unterbrink, and Tacheau}{Brumaghin
  et~al\mbox{.}}{2018}]%
        {brumaghin2018old}
\bibfield{author}{\bibinfo{person}{Edmund Brumaghin}, \bibinfo{person}{Holger
  Unterbrink}, {and} \bibinfo{person}{Emmanuel Tacheau}.}
  \bibinfo{year}{2018}\natexlab{}.
\newblock \bibinfo{title}{Old dog, new tricks - Analysing new RTF-based
  campaign distributing Agent Tesla, Loki with PyREbox}.
\newblock
  \bibinfo{howpublished}{{https://blog.talosintelligence.com/2018/10/old-dog-new-tricks-analysing-new-rtf\_15.html}}.
\newblock


\bibitem[\protect\citeauthoryear{Brumley and Boneh}{Brumley and Boneh}{2005}]%
        {brumley2005remote}
\bibfield{author}{\bibinfo{person}{David Brumley} {and} \bibinfo{person}{Dan
  Boneh}.} \bibinfo{year}{2005}\natexlab{}.
\newblock \showarticletitle{Remote timing attacks are practical}.
\newblock \bibinfo{journal}{\emph{Computer Networks}} \bibinfo{volume}{48},
  \bibinfo{number}{5} (\bibinfo{year}{2005}), \bibinfo{pages}{701--716}.
\newblock


\bibitem[\protect\citeauthoryear{Brundage, Avin, Clark, Toner, Eckersley,
  Garfinkel, Dafoe, Scharre, Zeitzoff, Filar, et~al\mbox{.}}{Brundage
  et~al\mbox{.}}{2018}]%
        {brundage2018malicious}
\bibfield{author}{\bibinfo{person}{Miles Brundage}, \bibinfo{person}{Shahar
  Avin}, \bibinfo{person}{Jack Clark}, \bibinfo{person}{Helen Toner},
  \bibinfo{person}{Peter Eckersley}, \bibinfo{person}{Ben Garfinkel},
  \bibinfo{person}{Allan Dafoe}, \bibinfo{person}{Paul Scharre},
  \bibinfo{person}{Thomas Zeitzoff}, \bibinfo{person}{Bobby Filar},
  {et~al\mbox{.}}} \bibinfo{year}{2018}\natexlab{}.
\newblock \showarticletitle{The malicious use of artificial intelligence:
  Forecasting, prevention, and mitigation}.
\newblock \bibinfo{journal}{\emph{arXiv preprint arXiv:1802.07228}}
  (\bibinfo{year}{2018}).
\newblock


\bibitem[\protect\citeauthoryear{Cagli, Dumas, and Prouff}{Cagli
  et~al\mbox{.}}{2017}]%
        {cagli2017convolutional}
\bibfield{author}{\bibinfo{person}{Eleonora Cagli}, \bibinfo{person}{C{\'e}cile
  Dumas}, {and} \bibinfo{person}{Emmanuel Prouff}.}
  \bibinfo{year}{2017}\natexlab{}.
\newblock \showarticletitle{Convolutional neural networks with data
  augmentation against jitter-based countermeasures}. In
  \bibinfo{booktitle}{\emph{International Conference on Cryptographic Hardware
  and Embedded Systems}}. Springer, \bibinfo{pages}{45--68}.
\newblock


\bibitem[\protect\citeauthoryear{Caldwell, Andrews, Tanay, and
  Griffin}{Caldwell et~al\mbox{.}}{2020}]%
        {caldwell2020ai}
\bibfield{author}{\bibinfo{person}{M Caldwell}, \bibinfo{person}{JTA Andrews},
  \bibinfo{person}{T Tanay}, {and} \bibinfo{person}{LD Griffin}.}
  \bibinfo{year}{2020}\natexlab{}.
\newblock \showarticletitle{AI-enabled future crime}.
\newblock \bibinfo{journal}{\emph{Crime Science}} \bibinfo{volume}{9},
  \bibinfo{number}{1} (\bibinfo{year}{2020}), \bibinfo{pages}{1--13}.
\newblock


\bibitem[\protect\citeauthoryear{Calzavara, Tolomei, Casini, Bugliesi, and
  Orlando}{Calzavara et~al\mbox{.}}{2015}]%
        {calzavara15-tw}
\bibfield{author}{\bibinfo{person}{Stefano Calzavara},
  \bibinfo{person}{Gabriele Tolomei}, \bibinfo{person}{Andrea Casini},
  \bibinfo{person}{Michele Bugliesi}, {and} \bibinfo{person}{Salvatore
  Orlando}.} \bibinfo{year}{2015}\natexlab{}.
\newblock \showarticletitle{A Supervised Learning Approach to Protect Client
  Authentication on the Web}.
\newblock \bibinfo{journal}{\emph{ACM Trans. Web}} \bibinfo{volume}{9},
  \bibinfo{number}{3}, Article \bibinfo{articleno}{15} (\bibinfo{date}{June}
  \bibinfo{year}{2015}), \bibinfo{numpages}{30}~pages.
\newblock
\showISSN{1559-1131}
\urldef\tempurl%
\url{https://doi.org/10.1145/2754933}
\showDOI{\tempurl}


\bibitem[\protect\citeauthoryear{{Cao}, {Qiao}, and {Lyu}}{{Cao}
  et~al\mbox{.}}{2017}]%
        {cao17-iccc}
\bibfield{author}{\bibinfo{person}{Q. {Cao}}, \bibinfo{person}{Y. {Qiao}},
  {and} \bibinfo{person}{Z. {Lyu}}.} \bibinfo{year}{2017}\natexlab{}.
\newblock \showarticletitle{Machine learning to detect anomalies in web log
  analysis}. In \bibinfo{booktitle}{\emph{2017 3rd IEEE International
  Conference on Computer and Communications (ICCC)}}.
  \bibinfo{pages}{519--523}.
\newblock


\bibitem[\protect\citeauthoryear{Castiglione, Prisco, Santis, Fiore, and
  Palmieri}{Castiglione et~al\mbox{.}}{2014}]%
        {castiglione14-jnca}
\bibfield{author}{\bibinfo{person}{Aniello Castiglione},
  \bibinfo{person}{Roberto~De Prisco}, \bibinfo{person}{Alfredo~De Santis},
  \bibinfo{person}{Ugo Fiore}, {and} \bibinfo{person}{Francesco Palmieri}.}
  \bibinfo{year}{2014}\natexlab{}.
\newblock \showarticletitle{A botnet-based command and control approach relying
  on swarm intelligence}.
\newblock \bibinfo{journal}{\emph{Journal of Network and Computer
  Applications}}  \bibinfo{volume}{38} (\bibinfo{year}{2014}),
  \bibinfo{pages}{22--33}.
\newblock
\showISSN{1084-8045}
\urldef\tempurl%
\url{https://doi.org/10.1016/j.jnca.2013.05.002}
\showDOI{\tempurl}


\bibitem[\protect\citeauthoryear{Chakraborty, Alam, Dey, Chattopadhyay, and
  Mukhopadhyay}{Chakraborty et~al\mbox{.}}{2018}]%
        {chakraborty18-cs}
\bibfield{author}{\bibinfo{person}{Anirban Chakraborty},
  \bibinfo{person}{Manaar Alam}, \bibinfo{person}{Vishal Dey},
  \bibinfo{person}{Anupam Chattopadhyay}, {and} \bibinfo{person}{Debdeep
  Mukhopadhyay}.} \bibinfo{year}{2018}\natexlab{}.
\newblock \showarticletitle{Adversarial {Attacks} and {Defences}: {A}
  {Survey}}.
\newblock \bibinfo{journal}{\emph{arXiv:1810.00069 [cs, stat]}}
  \bibinfo{volume}{ACM Computing Survey} (\bibinfo{date}{Sept.}
  \bibinfo{year}{2018}).
\newblock
\urldef\tempurl%
\url{http://arxiv.org/abs/1810.00069}
\showURL{%
\tempurl}
\newblock
\shownote{arXiv: 1810.00069.}


\bibitem[\protect\citeauthoryear{Chakraborty, Krishna, Ding, and
  Ray}{Chakraborty et~al\mbox{.}}{2020}]%
        {chakraborty2020deep}
\bibfield{author}{\bibinfo{person}{Saikat Chakraborty}, \bibinfo{person}{Rahul
  Krishna}, \bibinfo{person}{Yangruibo Ding}, {and} \bibinfo{person}{Baishakhi
  Ray}.} \bibinfo{year}{2020}\natexlab{}.
\newblock \showarticletitle{Deep Learning based Vulnerability Detection: Are We
  There Yet?}
\newblock \bibinfo{journal}{\emph{arXiv preprint arXiv:2009.07235}}
  (\bibinfo{year}{2020}).
\newblock


\bibitem[\protect\citeauthoryear{Chen, Su, Yeh, and Yung}{Chen
  et~al\mbox{.}}{2018b}]%
        {chen2018special}
\bibfield{author}{\bibinfo{person}{Jiageng Chen}, \bibinfo{person}{Chunhua Su},
  \bibinfo{person}{Kuo-Hui Yeh}, {and} \bibinfo{person}{Moti Yung}.}
  \bibinfo{year}{2018}\natexlab{b}.
\newblock \bibinfo{title}{Special issue on advanced persistent threat}.
\newblock
\newblock


\bibitem[\protect\citeauthoryear{Chen, Qiao, Wei, Zhong, and Huang}{Chen
  et~al\mbox{.}}{2014}]%
        {chen2014detecting}
\bibfield{author}{\bibinfo{person}{Wei Chen}, \bibinfo{person}{Xiaoqiang Qiao},
  \bibinfo{person}{Jun Wei}, \bibinfo{person}{Hua Zhong}, {and}
  \bibinfo{person}{Xiang Huang}.} \bibinfo{year}{2014}\natexlab{}.
\newblock \showarticletitle{Detecting inter-component configuration errors in
  proactive: a relation-aware method}. In \bibinfo{booktitle}{\emph{2014 14th
  International Conference on Quality Software}}. IEEE,
  \bibinfo{pages}{184--189}.
\newblock


\bibitem[\protect\citeauthoryear{Chen, Liu, Li, Lu, and Song}{Chen
  et~al\mbox{.}}{2017}]%
        {chen17-arxiv}
\bibfield{author}{\bibinfo{person}{X. Chen}, \bibinfo{person}{C. Liu},
  \bibinfo{person}{B. Li}, \bibinfo{person}{K. Lu}, {and} \bibinfo{person}{D.
  Song}.} \bibinfo{year}{2017}\natexlab{}.
\newblock \showarticletitle{Targeted {Backdoor} {Attacks} on {Deep} {Learning}
  {Systems} {Using} {Data} {Poisoning}}.
\newblock \bibinfo{journal}{\emph{ArXiv e-prints}}
  \bibinfo{volume}{abs/1712.05526} (\bibinfo{year}{2017}).
\newblock


\bibitem[\protect\citeauthoryear{Chen, Li, Zhang, Zhang, and Hedgpeth}{Chen
  et~al\mbox{.}}{2018a}]%
        {chen2018eyetell}
\bibfield{author}{\bibinfo{person}{Yimin Chen}, \bibinfo{person}{Tao Li},
  \bibinfo{person}{Rui Zhang}, \bibinfo{person}{Yanchao Zhang}, {and}
  \bibinfo{person}{Terri Hedgpeth}.} \bibinfo{year}{2018}\natexlab{a}.
\newblock \showarticletitle{Eyetell: Video-assisted touchscreen keystroke
  inference from eye movements}. In \bibinfo{booktitle}{\emph{2018 IEEE
  Symposium on Security and Privacy (SP)}}. IEEE, \bibinfo{pages}{144--160}.
\newblock


\bibitem[\protect\citeauthoryear{Cheng, Zhang, Zhang, Wu, Li, Fu, and Li}{Cheng
  et~al\mbox{.}}{2019}]%
        {cheng2019optimizing}
\bibfield{author}{\bibinfo{person}{Liang Cheng}, \bibinfo{person}{Yang Zhang},
  \bibinfo{person}{Yi Zhang}, \bibinfo{person}{Chen Wu},
  \bibinfo{person}{Zhangtan Li}, \bibinfo{person}{Yu Fu}, {and}
  \bibinfo{person}{Haisheng Li}.} \bibinfo{year}{2019}\natexlab{}.
\newblock \showarticletitle{Optimizing seed inputs in fuzzing with machine
  learning}. In \bibinfo{booktitle}{\emph{2019 IEEE/ACM 41st International
  Conference on Software Engineering: Companion Proceedings (ICSE-Companion)}}.
  IEEE, \bibinfo{pages}{244--245}.
\newblock


\bibitem[\protect\citeauthoryear{Cohen, Mirsky, Kamp, Martin, Elovici, Puzis,
  and Shabtai}{Cohen et~al\mbox{.}}{2020}]%
        {cohen2020dante}
\bibfield{author}{\bibinfo{person}{Dvir Cohen}, \bibinfo{person}{Yisroel
  Mirsky}, \bibinfo{person}{Manuel Kamp}, \bibinfo{person}{Tobias Martin},
  \bibinfo{person}{Yuval Elovici}, \bibinfo{person}{Rami Puzis}, {and}
  \bibinfo{person}{Asaf Shabtai}.} \bibinfo{year}{2020}\natexlab{}.
\newblock \showarticletitle{DANTE: A framework for mining and monitoring
  darknet traffic}. In \bibinfo{booktitle}{\emph{European Symposium on Research
  in Computer Security}}. Springer, \bibinfo{pages}{88--109}.
\newblock


\bibitem[\protect\citeauthoryear{Compagno, Conti, Lain, and Tsudik}{Compagno
  et~al\mbox{.}}{2017}]%
        {compagno2017don}
\bibfield{author}{\bibinfo{person}{Alberto Compagno}, \bibinfo{person}{Mauro
  Conti}, \bibinfo{person}{Daniele Lain}, {and} \bibinfo{person}{Gene Tsudik}.}
  \bibinfo{year}{2017}\natexlab{}.
\newblock \showarticletitle{Don't Skype \& Type! Acoustic Eavesdropping in
  Voice-Over-IP}. In \bibinfo{booktitle}{\emph{Proceedings of the 2017 ACM on
  Asia Conference on Computer and Communications Security}}.
  \bibinfo{pages}{703--715}.
\newblock


\bibitem[\protect\citeauthoryear{Croce and Hein}{Croce and Hein}{2020}]%
        {croce2020-icml}
\bibfield{author}{\bibinfo{person}{Francesco Croce} {and}
  \bibinfo{person}{Matthias Hein}.} \bibinfo{year}{2020}\natexlab{}.
\newblock \showarticletitle{Reliable evaluation of adversarial robustness with
  an ensemble of diverse parameter-free attacks}. In
  \bibinfo{booktitle}{\emph{ICML}}.
\newblock


\bibitem[\protect\citeauthoryear{Dabre, Chu, and Kunchukuttan}{Dabre
  et~al\mbox{.}}{2020}]%
        {dabre2020survey}
\bibfield{author}{\bibinfo{person}{Raj Dabre}, \bibinfo{person}{Chenhui Chu},
  {and} \bibinfo{person}{Anoop Kunchukuttan}.} \bibinfo{year}{2020}\natexlab{}.
\newblock \showarticletitle{A survey of multilingual neural machine
  translation}.
\newblock \bibinfo{journal}{\emph{ACM Computing Surveys (CSUR)}}
  \bibinfo{volume}{53}, \bibinfo{number}{5} (\bibinfo{year}{2020}),
  \bibinfo{pages}{1--38}.
\newblock


\bibitem[\protect\citeauthoryear{Dalvi, Domingos, Mausam, Sanghai, and
  Verma}{Dalvi et~al\mbox{.}}{2004}]%
        {dalvi04}
\bibfield{author}{\bibinfo{person}{Nilesh Dalvi}, \bibinfo{person}{Pedro
  Domingos}, \bibinfo{person}{Mausam}, \bibinfo{person}{Sumit Sanghai}, {and}
  \bibinfo{person}{Deepak Verma}.} \bibinfo{year}{2004}\natexlab{}.
\newblock \showarticletitle{Adversarial classification}. In
  \bibinfo{booktitle}{\emph{Tenth ACM SIGKDD International Conference on
  Knowledge Discovery and Data Mining (KDD)}}. \bibinfo{address}{Seattle},
  \bibinfo{pages}{99--108}.
\newblock


\bibitem[\protect\citeauthoryear{Das and Verma}{Das and Verma}{2019}]%
        {das2019automated}
\bibfield{author}{\bibinfo{person}{Avisha Das} {and} \bibinfo{person}{Rakesh
  Verma}.} \bibinfo{year}{2019}\natexlab{}.
\newblock \bibinfo{title}{Automated email Generation for Targeted Attacks using
  Natural Language}.
\newblock
\newblock
\showeprint[arxiv]{1908.06893}~[cs.CL]


\bibitem[\protect\citeauthoryear{Datta}{Datta}{2020}]%
        {datta2020deepobfuscode}
\bibfield{author}{\bibinfo{person}{Siddhartha Datta}.}
  \bibinfo{year}{2020}\natexlab{}.
\newblock \bibinfo{title}{DeepObfusCode: Source Code Obfuscation Through
  Sequence-to-Sequence Networks}.
\newblock
\newblock
\showeprint[arxiv]{1909.01837}~[cs.CR]


\bibitem[\protect\citeauthoryear{{Debnath}, {Solaimani}, {Gulzar}, {Arora},
  {Lumezanu}, {Xu}, {Zong}, {Zhang}, {Jiang}, and {Khan}}{{Debnath}
  et~al\mbox{.}}{2018}]%
        {debnath18-icdcs}
\bibfield{author}{\bibinfo{person}{B. {Debnath}}, \bibinfo{person}{M.
  {Solaimani}}, \bibinfo{person}{M.~A.~G. {Gulzar}}, \bibinfo{person}{N.
  {Arora}}, \bibinfo{person}{C. {Lumezanu}}, \bibinfo{person}{J. {Xu}},
  \bibinfo{person}{B. {Zong}}, \bibinfo{person}{H. {Zhang}},
  \bibinfo{person}{G. {Jiang}}, {and} \bibinfo{person}{L. {Khan}}.}
  \bibinfo{year}{2018}\natexlab{}.
\newblock \showarticletitle{LogLens: A Real-Time Log Analysis System}. In
  \bibinfo{booktitle}{\emph{2018 IEEE 38th International Conference on
  Distributed Computing Systems (ICDCS)}}. \bibinfo{pages}{1052--1062}.
\newblock


\bibitem[\protect\citeauthoryear{Demetrio, Biggio, Lagorio, Roli, and
  Armando}{Demetrio et~al\mbox{.}}{2020}]%
        {demetrio20-arxiv-blackbox}
\bibfield{author}{\bibinfo{person}{Luca Demetrio}, \bibinfo{person}{Battista
  Biggio}, \bibinfo{person}{Giovanni Lagorio}, \bibinfo{person}{Fabio Roli},
  {and} \bibinfo{person}{Alessandro Armando}.} \bibinfo{year}{2020}\natexlab{}.
\newblock \showarticletitle{Functionality-preserving {Black}-box {Optimization}
  of {Adversarial} {Windows} {Malware}}.
\newblock \bibinfo{journal}{\emph{arXiv:2003.13526 [cs]}}
  (\bibinfo{date}{Sept.} \bibinfo{year}{2020}).
\newblock
\urldef\tempurl%
\url{http://arxiv.org/abs/2003.13526}
\showURL{%
\tempurl}
\newblock
\shownote{arXiv: 2003.13526.}


\bibitem[\protect\citeauthoryear{Demontis, Melis, Pintor, Jagielski, Biggio,
  Oprea, Nita-Rotaru, and Roli}{Demontis et~al\mbox{.}}{2019a}]%
        {demontis19-usenix}
\bibfield{author}{\bibinfo{person}{Ambra Demontis}, \bibinfo{person}{Marco
  Melis}, \bibinfo{person}{Maura Pintor}, \bibinfo{person}{Matthew Jagielski},
  \bibinfo{person}{Battista Biggio}, \bibinfo{person}{Alina Oprea},
  \bibinfo{person}{Cristina Nita-Rotaru}, {and} \bibinfo{person}{Fabio Roli}.}
  \bibinfo{year}{2019}\natexlab{a}.
\newblock \showarticletitle{Why {Do} {Adversarial} {Attacks} {Transfer}?
  {Explaining} {Transferability} of {Evasion} and {Poisoning} {Attacks}}. In
  \bibinfo{booktitle}{\emph{28th {USENIX} {Security} {Symposium} ({USENIX}
  {Security} 19)}}. \bibinfo{publisher}{USENIX Association}.
\newblock


\bibitem[\protect\citeauthoryear{Demontis, Melis, Pintor, Jagielski, Biggio,
  Oprea, Nita-Rotaru, and Roli}{Demontis et~al\mbox{.}}{2019b}]%
        {demontis19-tdsc}
\bibfield{author}{\bibinfo{person}{Ambra Demontis}, \bibinfo{person}{Marco
  Melis}, \bibinfo{person}{Maura Pintor}, \bibinfo{person}{Matthew Jagielski},
  \bibinfo{person}{Battista Biggio}, \bibinfo{person}{Alina Oprea},
  \bibinfo{person}{Cristina Nita-Rotaru}, {and} \bibinfo{person}{Fabio Roli}.}
  \bibinfo{year}{2019}\natexlab{b}.
\newblock \showarticletitle{Why {Do} {Adversarial} {Attacks} {Transfer}?
  {Explaining} {Transferability} of {Evasion} and {Poisoning} {Attacks}}. In
  \bibinfo{booktitle}{\emph{28th {USENIX} {Security} {Symposium} ({USENIX}
  {Security} 19)}}. \bibinfo{publisher}{USENIX Association}.
\newblock


\bibitem[\protect\citeauthoryear{Dhaoui, Webster, and Tan}{Dhaoui
  et~al\mbox{.}}{2017}]%
        {dhaoui2017social}
\bibfield{author}{\bibinfo{person}{Chedia Dhaoui}, \bibinfo{person}{Cynthia~M
  Webster}, {and} \bibinfo{person}{Lay~Peng Tan}.}
  \bibinfo{year}{2017}\natexlab{}.
\newblock \showarticletitle{Social media sentiment analysis: lexicon versus
  machine learning}.
\newblock \bibinfo{journal}{\emph{Journal of Consumer Marketing}}
  (\bibinfo{year}{2017}).
\newblock


\bibitem[\protect\citeauthoryear{Ding, Huang, Patel, and Lovell}{Ding
  et~al\mbox{.}}{2018}]%
        {ding2018editorial}
\bibfield{author}{\bibinfo{person}{Changxing Ding}, \bibinfo{person}{Kaiqi
  Huang}, \bibinfo{person}{Vishal~M Patel}, {and} \bibinfo{person}{Brian~C
  Lovell}.} \bibinfo{year}{2018}\natexlab{}.
\newblock \showarticletitle{Special issue on Video Surveillance-oriented
  Biometrics}.
\newblock \bibinfo{journal}{\emph{Pattern Recognition Letters}}
  \bibinfo{volume}{107} (\bibinfo{year}{2018}), \bibinfo{pages}{1--2}.
\newblock


\bibitem[\protect\citeauthoryear{Ding, Fung, and Charland}{Ding
  et~al\mbox{.}}{2019}]%
        {ding2019asm2vec}
\bibfield{author}{\bibinfo{person}{Steven~HH Ding},
  \bibinfo{person}{Benjamin~CM Fung}, {and} \bibinfo{person}{Philippe
  Charland}.} \bibinfo{year}{2019}\natexlab{}.
\newblock \showarticletitle{Asm2vec: Boosting static representation robustness
  for binary clone search against code obfuscation and compiler optimization}.
  In \bibinfo{booktitle}{\emph{2019 IEEE Symposium on Security and Privacy
  (SP)}}. IEEE, \bibinfo{pages}{472--489}.
\newblock


\bibitem[\protect\citeauthoryear{Duan, Li, Wang, and Yin}{Duan
  et~al\mbox{.}}{2020}]%
        {duan2020deepbindiff}
\bibfield{author}{\bibinfo{person}{Yue Duan}, \bibinfo{person}{Xuezixiang Li},
  \bibinfo{person}{Jinghan Wang}, {and} \bibinfo{person}{Heng Yin}.}
  \bibinfo{year}{2020}\natexlab{}.
\newblock \showarticletitle{DEEPBINDIFF: Learning Program-Wide Code
  Representations for Binary Diffing}. In \bibinfo{booktitle}{\emph{Proceedings
  of the 27th Annual Network and Distributed System Security Symposium
  (NDSS’20)}}.
\newblock


\bibitem[\protect\citeauthoryear{Evangelista, Sassi, Romero, and
  Napolitano}{Evangelista et~al\mbox{.}}{2020}]%
        {evangelista2020systematic}
\bibfield{author}{\bibinfo{person}{Jo{\~a}o Rafael~Gon{\c{c}}alves
  Evangelista}, \bibinfo{person}{Renato~Jos{\'e} Sassi},
  \bibinfo{person}{M{\'a}rcio Romero}, {and} \bibinfo{person}{Domingos
  Napolitano}.} \bibinfo{year}{2020}\natexlab{}.
\newblock \showarticletitle{Systematic Literature Review to Investigate the
  Application of Open Source Intelligence (OSINT) with Artificial
  Intelligence}.
\newblock \bibinfo{journal}{\emph{Journal of Applied Security Research}}
  (\bibinfo{year}{2020}), \bibinfo{pages}{1--25}.
\newblock


\bibitem[\protect\citeauthoryear{{Fang}, {Wang}, {Li}, {Wu}, {Zhou}, and
  {Huang}}{{Fang} et~al\mbox{.}}{2019}]%
        {Fang2020MalwareEvasion}
\bibfield{author}{\bibinfo{person}{Z. {Fang}}, \bibinfo{person}{J. {Wang}},
  \bibinfo{person}{B. {Li}}, \bibinfo{person}{S. {Wu}}, \bibinfo{person}{Y.
  {Zhou}}, {and} \bibinfo{person}{H. {Huang}}.}
  \bibinfo{year}{2019}\natexlab{}.
\newblock \showarticletitle{Evading Anti-Malware Engines With Deep
  Reinforcement Learning}.
\newblock \bibinfo{journal}{\emph{IEEE Access}}  \bibinfo{volume}{7}
  (\bibinfo{year}{2019}), \bibinfo{pages}{48867--48879}.
\newblock
\urldef\tempurl%
\url{https://doi.org/10.1109/ACCESS.2019.2908033}
\showDOI{\tempurl}


\bibitem[\protect\citeauthoryear{Feng, Zhou, Xu, Cheng, Testa, and Yin}{Feng
  et~al\mbox{.}}{2016}]%
        {feng2016scalable}
\bibfield{author}{\bibinfo{person}{Qian Feng}, \bibinfo{person}{Rundong Zhou},
  \bibinfo{person}{Chengcheng Xu}, \bibinfo{person}{Yao Cheng},
  \bibinfo{person}{Brian Testa}, {and} \bibinfo{person}{Heng Yin}.}
  \bibinfo{year}{2016}\natexlab{}.
\newblock \showarticletitle{Scalable graph-based bug search for firmware
  images}. In \bibinfo{booktitle}{\emph{Proceedings of the 2016 ACM SIGSAC
  Conference on Computer and Communications Security}}.
  \bibinfo{pages}{480--491}.
\newblock


\bibitem[\protect\citeauthoryear{Fu, Tan, Peng, Zhao, and Yan}{Fu
  et~al\mbox{.}}{2017}]%
        {fu2017style}
\bibfield{author}{\bibinfo{person}{Zhenxin Fu}, \bibinfo{person}{Xiaoye Tan},
  \bibinfo{person}{Nanyun Peng}, \bibinfo{person}{Dongyan Zhao}, {and}
  \bibinfo{person}{Rui Yan}.} \bibinfo{year}{2017}\natexlab{}.
\newblock \showarticletitle{Style transfer in text: Exploration and
  evaluation}.
\newblock \bibinfo{journal}{\emph{arXiv preprint arXiv:1711.06861}}
  (\bibinfo{year}{2017}).
\newblock


\bibitem[\protect\citeauthoryear{Fuller, Fan, Day, and Barlow}{Fuller
  et~al\mbox{.}}{2020}]%
        {fuller2020digital}
\bibfield{author}{\bibinfo{person}{Aidan Fuller}, \bibinfo{person}{Zhong Fan},
  \bibinfo{person}{Charles Day}, {and} \bibinfo{person}{Chris Barlow}.}
  \bibinfo{year}{2020}\natexlab{}.
\newblock \showarticletitle{Digital twin: Enabling technologies, challenges and
  open research}.
\newblock \bibinfo{journal}{\emph{IEEE Access}}  \bibinfo{volume}{8}
  (\bibinfo{year}{2020}), \bibinfo{pages}{108952--108971}.
\newblock


\bibitem[\protect\citeauthoryear{Gandolfi, Mourtel, and Olivier}{Gandolfi
  et~al\mbox{.}}{2001}]%
        {gandolfi2001electromagnetic}
\bibfield{author}{\bibinfo{person}{Karine Gandolfi},
  \bibinfo{person}{Christophe Mourtel}, {and} \bibinfo{person}{Francis
  Olivier}.} \bibinfo{year}{2001}\natexlab{}.
\newblock \showarticletitle{Electromagnetic analysis: Concrete results}. In
  \bibinfo{booktitle}{\emph{International workshop on cryptographic hardware
  and embedded systems}}. Springer, \bibinfo{pages}{251--261}.
\newblock


\bibitem[\protect\citeauthoryear{{Gao}, {Lanchantin}, {Soffa}, and {Qi}}{{Gao}
  et~al\mbox{.}}{2018}]%
        {gao18-spw}
\bibfield{author}{\bibinfo{person}{J. {Gao}}, \bibinfo{person}{J.
  {Lanchantin}}, \bibinfo{person}{M.~L. {Soffa}}, {and} \bibinfo{person}{Y.
  {Qi}}.} \bibinfo{year}{2018}\natexlab{}.
\newblock \showarticletitle{Black-Box Generation of Adversarial Text Sequences
  to Evade Deep Learning Classifiers}. In \bibinfo{booktitle}{\emph{2018 IEEE
  Security and Privacy Workshops (SPW)}}. \bibinfo{pages}{50--56}.
\newblock
\urldef\tempurl%
\url{https://doi.org/10.1109/SPW.2018.00016}
\showDOI{\tempurl}


\bibitem[\protect\citeauthoryear{Garg and Ahuja}{Garg and Ahuja}{2019}]%
        {garg2019password}
\bibfield{author}{\bibinfo{person}{Vernit Garg} {and} \bibinfo{person}{Laxmi
  Ahuja}.} \bibinfo{year}{2019}\natexlab{}.
\newblock \showarticletitle{Password Guessing Using Deep Learning}. In
  \bibinfo{booktitle}{\emph{2019 2nd International Conference on Power Energy,
  Environment and Intelligent Control (PEEIC)}}. IEEE, \bibinfo{pages}{38--40}.
\newblock


\bibitem[\protect\citeauthoryear{Ghazi, Anwar, Mumtaz, Saleem, and Tahir}{Ghazi
  et~al\mbox{.}}{2018}]%
        {ghazi2018supervised}
\bibfield{author}{\bibinfo{person}{Yumna Ghazi}, \bibinfo{person}{Zahid Anwar},
  \bibinfo{person}{Rafia Mumtaz}, \bibinfo{person}{Shahzad Saleem}, {and}
  \bibinfo{person}{Ali Tahir}.} \bibinfo{year}{2018}\natexlab{}.
\newblock \showarticletitle{A supervised machine learning based approach for
  automatically extracting high-level threat intelligence from unstructured
  sources}. In \bibinfo{booktitle}{\emph{2018 International Conference on
  Frontiers of Information Technology (FIT)}}. IEEE, \bibinfo{pages}{129--134}.
\newblock


\bibitem[\protect\citeauthoryear{Ghiassi and Lee}{Ghiassi and Lee}{2018}]%
        {ghiassi2018domain}
\bibfield{author}{\bibinfo{person}{Manoochehr Ghiassi} {and} \bibinfo{person}{S
  Lee}.} \bibinfo{year}{2018}\natexlab{}.
\newblock \showarticletitle{A domain transferable lexicon set for Twitter
  sentiment analysis using a supervised machine learning approach}.
\newblock \bibinfo{journal}{\emph{Expert Systems with Applications}}
  \bibinfo{volume}{106} (\bibinfo{year}{2018}), \bibinfo{pages}{197--216}.
\newblock


\bibitem[\protect\citeauthoryear{Gilad-Bachrach, Dowlin, Laine, Lauter,
  Naehrig, and Wernsing}{Gilad-Bachrach et~al\mbox{.}}{2016}]%
        {gilad2016cryptonets}
\bibfield{author}{\bibinfo{person}{Ran Gilad-Bachrach}, \bibinfo{person}{Nathan
  Dowlin}, \bibinfo{person}{Kim Laine}, \bibinfo{person}{Kristin Lauter},
  \bibinfo{person}{Michael Naehrig}, {and} \bibinfo{person}{John Wernsing}.}
  \bibinfo{year}{2016}\natexlab{}.
\newblock \showarticletitle{Cryptonets: Applying neural networks to encrypted
  data with high throughput and accuracy}. In
  \bibinfo{booktitle}{\emph{International Conference on Machine Learning}}.
  \bibinfo{pages}{201--210}.
\newblock


\bibitem[\protect\citeauthoryear{Goodfellow, Pouget-Abadie, Mirza, Xu,
  Warde-Farley, Ozair, Courville, and Bengio}{Goodfellow et~al\mbox{.}}{2014}]%
        {goodfellow14-nips}
\bibfield{author}{\bibinfo{person}{Ian~J. Goodfellow}, \bibinfo{person}{Jean
  Pouget-Abadie}, \bibinfo{person}{Mehdi Mirza}, \bibinfo{person}{Bing Xu},
  \bibinfo{person}{David Warde-Farley}, \bibinfo{person}{Sherjil Ozair},
  \bibinfo{person}{Aaron Courville}, {and} \bibinfo{person}{Yoshua Bengio}.}
  \bibinfo{year}{2014}\natexlab{}.
\newblock \showarticletitle{Generative Adversarial Nets}. In
  \bibinfo{booktitle}{\emph{Proceedings of the 27th International Conference on
  Neural Information Processing Systems - Volume 2}} (Montreal, Canada)
  \emph{(\bibinfo{series}{NIPS'14})}. \bibinfo{publisher}{MIT Press},
  \bibinfo{address}{Cambridge, MA, USA}, \bibinfo{pages}{2672–2680}.
\newblock


\bibitem[\protect\citeauthoryear{Goodfellow, Shlens, and Szegedy}{Goodfellow
  et~al\mbox{.}}{2015}]%
        {goodfellow15-iclr}
\bibfield{author}{\bibinfo{person}{Ian~J. Goodfellow},
  \bibinfo{person}{Jonathon Shlens}, {and} \bibinfo{person}{Christian
  Szegedy}.} \bibinfo{year}{2015}\natexlab{}.
\newblock \showarticletitle{Explaining and Harnessing Adversarial Examples}. In
  \bibinfo{booktitle}{\emph{International Conference on Learning
  Representations}}.
\newblock


\bibitem[\protect\citeauthoryear{Gu, Dolan-Gavitt, and Garg}{Gu
  et~al\mbox{.}}{2017}]%
        {gu17-nips}
\bibfield{author}{\bibinfo{person}{Tianyu Gu}, \bibinfo{person}{Brendan
  Dolan-Gavitt}, {and} \bibinfo{person}{Siddharth Garg}.}
  \bibinfo{year}{2017}\natexlab{}.
\newblock \showarticletitle{{BadNets}: {Identifying} {Vulnerabilities} in the
  {Machine} {Learning} {Model} {Supply} {Chain}}. In
  \bibinfo{booktitle}{\emph{{NIPS} {Workshop} on {Machine} {Learning} and
  {Computer} {Security}}}, Vol.~\bibinfo{volume}{abs/1708.06733}.
\newblock


\bibitem[\protect\citeauthoryear{Guo, Fan, Pang, Yang, Ai, Zamani, Wu, Croft,
  and Cheng}{Guo et~al\mbox{.}}{2019}]%
        {guo2019deep}
\bibfield{author}{\bibinfo{person}{Jiafeng Guo}, \bibinfo{person}{Yixing Fan},
  \bibinfo{person}{Liang Pang}, \bibinfo{person}{Liu Yang},
  \bibinfo{person}{Qingyao Ai}, \bibinfo{person}{Hamed Zamani},
  \bibinfo{person}{Chen Wu}, \bibinfo{person}{W~Bruce Croft}, {and}
  \bibinfo{person}{Xueqi Cheng}.} \bibinfo{year}{2019}\natexlab{}.
\newblock \showarticletitle{A deep look into neural ranking models for
  information retrieval}.
\newblock \bibinfo{journal}{\emph{Information Processing \& Management}}
  (\bibinfo{year}{2019}), \bibinfo{pages}{102067}.
\newblock


\bibitem[\protect\citeauthoryear{Guri and Elovici}{Guri and Elovici}{2018}]%
        {guri2018bridgeware}
\bibfield{author}{\bibinfo{person}{Mordechai Guri} {and} \bibinfo{person}{Yuval
  Elovici}.} \bibinfo{year}{2018}\natexlab{}.
\newblock \showarticletitle{Bridgeware: The air-gap malware}.
\newblock \bibinfo{journal}{\emph{Commun. ACM}} \bibinfo{volume}{61},
  \bibinfo{number}{4} (\bibinfo{year}{2018}), \bibinfo{pages}{74--82}.
\newblock


\bibitem[\protect\citeauthoryear{Hajipour, Malinowski, and Fritz}{Hajipour
  et~al\mbox{.}}{2020}]%
        {hajipour20-arxiv}
\bibfield{author}{\bibinfo{person}{Hossein Hajipour}, \bibinfo{person}{Mateusz
  Malinowski}, {and} \bibinfo{person}{Mario Fritz}.}
  \bibinfo{year}{2020}\natexlab{}.
\newblock \bibinfo{title}{IReEn: Iterative Reverse-Engineering of Black-Box
  Functions via Neural Program Synthesis}.
\newblock
\newblock
\showeprint[arxiv]{2006.10720}~[cs.LG]


\bibitem[\protect\citeauthoryear{Han, Wang, Zhong, Chen, Yang, Lu, Shi, and
  Yin}{Han et~al\mbox{.}}{2020}]%
        {han2020practical}
\bibfield{author}{\bibinfo{person}{Dongqi Han}, \bibinfo{person}{Zhiliang
  Wang}, \bibinfo{person}{Ying Zhong}, \bibinfo{person}{Wenqi Chen},
  \bibinfo{person}{Jiahai Yang}, \bibinfo{person}{Shuqiang Lu},
  \bibinfo{person}{Xingang Shi}, {and} \bibinfo{person}{Xia Yin}.}
  \bibinfo{year}{2020}\natexlab{}.
\newblock \showarticletitle{Practical traffic-space adversarial attacks on
  learning-based nidss}.
\newblock \bibinfo{journal}{\emph{arXiv preprint arXiv:2005.07519}}
  (\bibinfo{year}{2020}).
\newblock


\bibitem[\protect\citeauthoryear{Hasegawa, Yanagisawa, and Togawa}{Hasegawa
  et~al\mbox{.}}{2020}]%
        {hasegawa2020trojan}
\bibfield{author}{\bibinfo{person}{Kento Hasegawa}, \bibinfo{person}{Masao
  Yanagisawa}, {and} \bibinfo{person}{Nozomu Togawa}.}
  \bibinfo{year}{2020}\natexlab{}.
\newblock \showarticletitle{Trojan-Net Classification for Gate-Level Hardware
  Design Utilizing Boundary Net Structures}.
\newblock \bibinfo{journal}{\emph{IEICE TRANSACTIONS on Information and
  Systems}} \bibinfo{volume}{103}, \bibinfo{number}{7} (\bibinfo{year}{2020}),
  \bibinfo{pages}{1618--1622}.
\newblock


\bibitem[\protect\citeauthoryear{Heuser, Picek, Guilley, and Mentens}{Heuser
  et~al\mbox{.}}{2016}]%
        {heuser2016side}
\bibfield{author}{\bibinfo{person}{Annelie Heuser}, \bibinfo{person}{Stjepan
  Picek}, \bibinfo{person}{Sylvain Guilley}, {and} \bibinfo{person}{Nele
  Mentens}.} \bibinfo{year}{2016}\natexlab{}.
\newblock \showarticletitle{Side-channel analysis of lightweight ciphers: Does
  lightweight equal easy?}. In \bibinfo{booktitle}{\emph{International Workshop
  on Radio Frequency Identification: Security and Privacy Issues}}. Springer,
  \bibinfo{pages}{91--104}.
\newblock


\bibitem[\protect\citeauthoryear{{Hidano}, {Murakami}, {Katsumata}, {Kiyomoto},
  and {Hanaoka}}{{Hidano} et~al\mbox{.}}{2017}]%
        {Model_Inversion}
\bibfield{author}{\bibinfo{person}{S. {Hidano}}, \bibinfo{person}{T.
  {Murakami}}, \bibinfo{person}{S. {Katsumata}}, \bibinfo{person}{S.
  {Kiyomoto}}, {and} \bibinfo{person}{G. {Hanaoka}}.}
  \bibinfo{year}{2017}\natexlab{}.
\newblock \showarticletitle{Model Inversion Attacks for Prediction Systems:
  Without Knowledge of Non-Sensitive Attributes}. In
  \bibinfo{booktitle}{\emph{2017 15th Annual Conference on Privacy, Security
  and Trust (PST)}}. \bibinfo{pages}{115--11509}.
\newblock
\urldef\tempurl%
\url{https://doi.org/10.1109/PST.2017.00023}
\showDOI{\tempurl}


\bibitem[\protect\citeauthoryear{Hitaj, Gasti, Ateniese, and Perez-Cruz}{Hitaj
  et~al\mbox{.}}{2019}]%
        {hitaj2019passgan}
\bibfield{author}{\bibinfo{person}{Briland Hitaj}, \bibinfo{person}{Paolo
  Gasti}, \bibinfo{person}{Giuseppe Ateniese}, {and} \bibinfo{person}{Fernando
  Perez-Cruz}.} \bibinfo{year}{2019}\natexlab{}.
\newblock \showarticletitle{Passgan: A deep learning approach for password
  guessing}. In \bibinfo{booktitle}{\emph{International Conference on Applied
  Cryptography and Network Security}}. Springer, \bibinfo{pages}{217--237}.
\newblock


\bibitem[\protect\citeauthoryear{Huang, Joseph, Nelson, Rubinstein, and
  Tygar}{Huang et~al\mbox{.}}{2011}]%
        {huang11}
\bibfield{author}{\bibinfo{person}{L. Huang}, \bibinfo{person}{A.~D. Joseph},
  \bibinfo{person}{B. Nelson}, \bibinfo{person}{B. Rubinstein}, {and}
  \bibinfo{person}{J.~D. Tygar}.} \bibinfo{year}{2011}\natexlab{}.
\newblock \showarticletitle{Adversarial Machine Learning}. In
  \bibinfo{booktitle}{\emph{4th ACM Workshop on Artificial Intelligence and
  Security ({AISec 2011})}}. \bibinfo{address}{Chicago, IL, USA},
  \bibinfo{pages}{43--57}.
\newblock


\bibitem[\protect\citeauthoryear{Hussain, Al-Haiqi, Zaidan, Zaidan, Kiah,
  Anuar, and Abdulnabi}{Hussain et~al\mbox{.}}{2016}]%
        {hussain2016rise}
\bibfield{author}{\bibinfo{person}{Muzammil Hussain}, \bibinfo{person}{Ahmed
  Al-Haiqi}, \bibinfo{person}{AA Zaidan}, \bibinfo{person}{BB Zaidan},
  \bibinfo{person}{ML~Mat Kiah}, \bibinfo{person}{Nor~Badrul Anuar}, {and}
  \bibinfo{person}{Mohamed Abdulnabi}.} \bibinfo{year}{2016}\natexlab{}.
\newblock \showarticletitle{The rise of keyloggers on smartphones: A survey and
  insight into motion-based tap inference attacks}.
\newblock \bibinfo{journal}{\emph{Pervasive and Mobile Computing}}
  \bibinfo{volume}{25} (\bibinfo{year}{2016}), \bibinfo{pages}{1--25}.
\newblock


\bibitem[\protect\citeauthoryear{Huybrechts, Vanommeslaeghe, Blontrock,
  Van~Barel, and Hellinckx}{Huybrechts et~al\mbox{.}}{2017}]%
        {huybrechts2017automatic}
\bibfield{author}{\bibinfo{person}{Thomas Huybrechts}, \bibinfo{person}{Yon
  Vanommeslaeghe}, \bibinfo{person}{Dries Blontrock}, \bibinfo{person}{Gregory
  Van~Barel}, {and} \bibinfo{person}{Peter Hellinckx}.}
  \bibinfo{year}{2017}\natexlab{}.
\newblock \showarticletitle{Automatic reverse engineering of CAN bus data using
  machine learning techniques}. In \bibinfo{booktitle}{\emph{International
  Conference on P2P, Parallel, Grid, Cloud and Internet Computing}}. Springer,
  \bibinfo{pages}{751--761}.
\newblock


\bibitem[\protect\citeauthoryear{Ilin}{Ilin}{[n.d.]}]%
        {twds_building_news_aggr}
\bibfield{author}{\bibinfo{person}{Ivan Ilin}.}
  \bibinfo{year}{[n.d.]}\natexlab{}.
\newblock \bibinfo{title}{Building a news aggregator from scratch: news
  filtering, classification, grouping in threads and ranking}.
\newblock
  \bibinfo{howpublished}{https://towardsdatascience.com/building-a-news-aggregator-from-scratch-news-filtering-classification-grouping-in-threads-and-7b0bbf619b68}.
\newblock
\newblock
\shownote{(Accessed on 10/14/2020).}


\bibitem[\protect\citeauthoryear{Intelligence}{Intelligence}{2015}]%
        {intelligence2015hammertoss}
\bibfield{author}{\bibinfo{person}{Fire Eye~Threat Intelligence}.}
  \bibinfo{year}{2015}\natexlab{}.
\newblock \showarticletitle{HAMMERTOSS: stealthy tactics define a Russian cyber
  threat group}.
\newblock \bibinfo{journal}{\emph{Milpitas, CA: FireEye, Inc}}
  (\bibinfo{year}{2015}).
\newblock


\bibitem[\protect\citeauthoryear{Ispoglou and Payer}{Ispoglou and
  Payer}{2016}]%
        {ispoglou16-usenix}
\bibfield{author}{\bibinfo{person}{Kyriakos~K. Ispoglou} {and}
  \bibinfo{person}{Mathias Payer}.} \bibinfo{year}{2016}\natexlab{}.
\newblock \showarticletitle{malWASH: Washing Malware to Evade Dynamic
  Analysis}. In \bibinfo{booktitle}{\emph{10th {USENIX} Workshop on Offensive
  Technologies ({WOOT} 16)}}. \bibinfo{publisher}{{USENIX} Association},
  \bibinfo{address}{Austin, TX}.
\newblock
\urldef\tempurl%
\url{https://www.usenix.org/conference/woot16/workshop-program/presentation/ispoglou}
\showURL{%
\tempurl}


\bibitem[\protect\citeauthoryear{Janota}{Janota}{2018}]%
        {janota2018towards}
\bibfield{author}{\bibinfo{person}{Mikol{\'a}s Janota}.}
  \bibinfo{year}{2018}\natexlab{}.
\newblock \showarticletitle{Towards Generalization in QBF Solving via Machine
  Learning.}. In \bibinfo{booktitle}{\emph{AAAI}}. \bibinfo{pages}{6607--6614}.
\newblock


\bibitem[\protect\citeauthoryear{Javed, Beg, Asim, Baker, and Al-Bayatti}{Javed
  et~al\mbox{.}}{2020}]%
        {javed2020alphalogger}
\bibfield{author}{\bibinfo{person}{Abdul~Rehman Javed},
  \bibinfo{person}{Mirza~Omer Beg}, \bibinfo{person}{Muhammad Asim},
  \bibinfo{person}{Thar Baker}, {and} \bibinfo{person}{Ali~Hilal Al-Bayatti}.}
  \bibinfo{year}{2020}\natexlab{}.
\newblock \showarticletitle{AlphaLogger: Detecting motion-based side-channel
  attack using smartphone keystrokes}.
\newblock \bibinfo{journal}{\emph{Journal of Ambient Intelligence and Humanized
  Computing}} (\bibinfo{year}{2020}), \bibinfo{pages}{1--14}.
\newblock


\bibitem[\protect\citeauthoryear{Jia, Choquette-Choo, Chandrasekaran, and
  Papernot}{Jia et~al\mbox{.}}{2021}]%
        {jia20}
\bibfield{author}{\bibinfo{person}{Hengrui Jia},
  \bibinfo{person}{Christopher~A. Choquette-Choo}, \bibinfo{person}{Varun
  Chandrasekaran}, {and} \bibinfo{person}{Nicolas Papernot}.}
  \bibinfo{year}{2021}\natexlab{}.
\newblock \bibinfo{title}{Entangled Watermarks as a Defense against Model
  Extraction}.
\newblock
\newblock
\showeprint[arxiv]{2002.12200}~[cs.CR]


\bibitem[\protect\citeauthoryear{Jia, Zhang, Weiss, Wang, Shen, Ren, Chen,
  Nguyen, Pang, Lopez~Moreno, and Wu}{Jia et~al\mbox{.}}{2018}]%
        {jia18-NIPS}
\bibfield{author}{\bibinfo{person}{Ye Jia}, \bibinfo{person}{Yu Zhang},
  \bibinfo{person}{Ron Weiss}, \bibinfo{person}{Quan Wang},
  \bibinfo{person}{Jonathan Shen}, \bibinfo{person}{Fei Ren},
  \bibinfo{person}{zhifeng Chen}, \bibinfo{person}{Patrick Nguyen},
  \bibinfo{person}{Ruoming Pang}, \bibinfo{person}{Ignacio Lopez~Moreno}, {and}
  \bibinfo{person}{Yonghui Wu}.} \bibinfo{year}{2018}\natexlab{}.
\newblock \showarticletitle{Transfer Learning from Speaker Verification to
  Multispeaker Text-To-Speech Synthesis}.
\newblock In \bibinfo{booktitle}{\emph{Advances in Neural Information
  Processing Systems 31}}, \bibfield{editor}{\bibinfo{person}{S.~Bengio},
  \bibinfo{person}{H.~Wallach}, \bibinfo{person}{H.~Larochelle},
  \bibinfo{person}{K.~Grauman}, \bibinfo{person}{N.~Cesa-Bianchi}, {and}
  \bibinfo{person}{R.~Garnett}} (Eds.). \bibinfo{publisher}{Curran Associates,
  Inc.}, \bibinfo{pages}{4480--4490}.
\newblock
\urldef\tempurl%
\url{http://papers.nips.cc/paper/7700-transfer-learning-from-speaker-verification-to-multispeaker-text-to-speech-synthesis.pdf}
\showURL{%
\tempurl}


\bibitem[\protect\citeauthoryear{Jiang, Yu, Sun, and Zeng}{Jiang
  et~al\mbox{.}}{2019}]%
        {jiang2019survey}
\bibfield{author}{\bibinfo{person}{Jian Jiang}, \bibinfo{person}{Xiangzhan Yu},
  \bibinfo{person}{Yan Sun}, {and} \bibinfo{person}{Haohua Zeng}.}
  \bibinfo{year}{2019}\natexlab{}.
\newblock \showarticletitle{A Survey of the Software Vulnerability Discovery
  Using Machine Learning Techniques}. In
  \bibinfo{booktitle}{\emph{International Conference on Artificial Intelligence
  and Security}}. Springer, \bibinfo{pages}{308--317}.
\newblock


\bibitem[\protect\citeauthoryear{Jiang, Ye, Huang, Shang, and Zheng}{Jiang
  et~al\mbox{.}}{2020}]%
        {jiang2020smartsteganogaphy}
\bibfield{author}{\bibinfo{person}{Shunzhi Jiang}, \bibinfo{person}{Dengpan
  Ye}, \bibinfo{person}{Jiaqing Huang}, \bibinfo{person}{Yueyun Shang}, {and}
  \bibinfo{person}{Zhuoyuan Zheng}.} \bibinfo{year}{2020}\natexlab{}.
\newblock \showarticletitle{SmartSteganogaphy: Light-weight generative audio
  steganography model for smart embedding application}.
\newblock \bibinfo{journal}{\emph{Journal of Network and Computer
  Applications}}  \bibinfo{volume}{165} (\bibinfo{year}{2020}),
  \bibinfo{pages}{102689}.
\newblock


\bibitem[\protect\citeauthoryear{Jiao, Zhang, Liu, Yang, Li, Feng, and Qu}{Jiao
  et~al\mbox{.}}{2019}]%
        {jiao2019survey}
\bibfield{author}{\bibinfo{person}{Licheng Jiao}, \bibinfo{person}{Fan Zhang},
  \bibinfo{person}{Fang Liu}, \bibinfo{person}{Shuyuan Yang},
  \bibinfo{person}{Lingling Li}, \bibinfo{person}{Zhixi Feng}, {and}
  \bibinfo{person}{Rong Qu}.} \bibinfo{year}{2019}\natexlab{}.
\newblock \showarticletitle{A survey of deep learning-based object detection}.
\newblock \bibinfo{journal}{\emph{IEEE Access}}  \bibinfo{volume}{7}
  (\bibinfo{year}{2019}), \bibinfo{pages}{128837--128868}.
\newblock


\bibitem[\protect\citeauthoryear{Joseph, Nelson, Rubinstein, and Tygar}{Joseph
  et~al\mbox{.}}{2018}]%
        {joseph18-advml-book}
\bibfield{author}{\bibinfo{person}{Anthony~D. Joseph}, \bibinfo{person}{Blaine
  Nelson}, \bibinfo{person}{Benjamin I.~P. Rubinstein}, {and}
  \bibinfo{person}{J.D. Tygar}.} \bibinfo{year}{2018}\natexlab{}.
\newblock \bibinfo{booktitle}{\emph{Adversarial Machine Learning}}.
\newblock \bibinfo{publisher}{Cambridge University Press}.
\newblock


\bibitem[\protect\citeauthoryear{{Juuti}, {Szyller}, {Marchal}, and
  {Asokan}}{{Juuti} et~al\mbox{.}}{2019}]%
        {Juuti19-sp}
\bibfield{author}{\bibinfo{person}{M. {Juuti}}, \bibinfo{person}{S. {Szyller}},
  \bibinfo{person}{S. {Marchal}}, {and} \bibinfo{person}{N. {Asokan}}.}
  \bibinfo{year}{2019}\natexlab{}.
\newblock \showarticletitle{PRADA: Protecting Against DNN Model Stealing
  Attacks}. In \bibinfo{booktitle}{\emph{2019 IEEE European Symposium on
  Security and Privacy (EuroS P)}}. \bibinfo{pages}{512--527}.
\newblock
\urldef\tempurl%
\url{https://doi.org/10.1109/EuroSP.2019.00044}
\showDOI{\tempurl}


\bibitem[\protect\citeauthoryear{Knake}{Knake}{2017}]%
        {Knake17}
\bibfield{author}{\bibinfo{person}{Robert~K. Knake}.}
  \bibinfo{year}{2017}\natexlab{}.
\newblock \bibinfo{booktitle}{\emph{A Cyberattack on the U.S. Power Grid}}.
\newblock \bibinfo{type}{{T}echnical {R}eport}. \bibinfo{institution}{Council
  on Foreign Relations}.
\newblock
\urldef\tempurl%
\url{http://www.jstor.org/stable/resrep05652}
\showURL{%
\tempurl}


\bibitem[\protect\citeauthoryear{Kocher, Jaffe, and Jun}{Kocher
  et~al\mbox{.}}{1999}]%
        {kocher1999differential}
\bibfield{author}{\bibinfo{person}{Paul Kocher}, \bibinfo{person}{Joshua
  Jaffe}, {and} \bibinfo{person}{Benjamin Jun}.}
  \bibinfo{year}{1999}\natexlab{}.
\newblock \showarticletitle{Differential power analysis}. In
  \bibinfo{booktitle}{\emph{Annual international cryptology conference}}.
  Springer, \bibinfo{pages}{388--397}.
\newblock


\bibitem[\protect\citeauthoryear{Koh and Liang}{Koh and Liang}{2017}]%
        {koh17-icml}
\bibfield{author}{\bibinfo{person}{P.~W. Koh} {and} \bibinfo{person}{P.
  Liang}.} \bibinfo{year}{2017}\natexlab{}.
\newblock \showarticletitle{Understanding {Black}-box {Predictions} via
  {Influence} {Functions}}. In \bibinfo{booktitle}{\emph{International
  {Conference} on {Machine} {Learning} ({ICML})}}.
\newblock


\bibitem[\protect\citeauthoryear{Kolosnjaji, Demontis, Biggio, Maiorca,
  Giacinto, Eckert, and Roli}{Kolosnjaji et~al\mbox{.}}{2018}]%
        {kolosnjaji18-eusipco}
\bibfield{author}{\bibinfo{person}{Bojan Kolosnjaji}, \bibinfo{person}{Ambra
  Demontis}, \bibinfo{person}{Battista Biggio}, \bibinfo{person}{Davide
  Maiorca}, \bibinfo{person}{Giorgio Giacinto}, \bibinfo{person}{Claudia
  Eckert}, {and} \bibinfo{person}{Fabio Roli}.}
  \bibinfo{year}{2018}\natexlab{}.
\newblock \showarticletitle{Adversarial {Malware} {Binaries}: {Evading} {Deep}
  {Learning} for {Malware} {Detection} in {Executables}}. In
  \bibinfo{booktitle}{\emph{26th {European} {Signal} {Processing} {Conf}.}}
  \emph{(\bibinfo{series}{{EUSIPCO}})}. \bibinfo{publisher}{IEEE},
  \bibinfo{address}{Rome}, \bibinfo{pages}{533--537}.
\newblock


\bibitem[\protect\citeauthoryear{Kong and Tong}{Kong and Tong}{2020}]%
        {kong2020dynamic}
\bibfield{author}{\bibinfo{person}{Ru Kong} {and} \bibinfo{person}{Xiangrong
  Tong}.} \bibinfo{year}{2020}\natexlab{}.
\newblock \showarticletitle{Dynamic Weighted Heuristic Trust Path Search
  Algorithm}.
\newblock \bibinfo{journal}{\emph{IEEE Access}}  \bibinfo{volume}{8}
  (\bibinfo{year}{2020}), \bibinfo{pages}{157382--157390}.
\newblock


\bibitem[\protect\citeauthoryear{Krebs}{Krebs}{2014}]%
        {TargetHa48:online}
\bibfield{author}{\bibinfo{person}{Brian Krebs}.}
  \bibinfo{year}{2014}\natexlab{}.
\newblock \bibinfo{title}{Target Hackers Broke in Via HVAC Company – Krebs on
  Security}.
\newblock
  \bibinfo{howpublished}{{https://krebsonsecurity.com/2014/02/target-hackers-broke-in-via-hvac-company/}}.
\newblock
\newblock
\shownote{(Accessed on 04/15/2021).}


\bibitem[\protect\citeauthoryear{Kumar, Biswas, and Sanyal}{Kumar
  et~al\mbox{.}}{2018}]%
        {kumar2018ecommercegan}
\bibfield{author}{\bibinfo{person}{Ashutosh Kumar}, \bibinfo{person}{Arijit
  Biswas}, {and} \bibinfo{person}{Subhajit Sanyal}.}
  \bibinfo{year}{2018}\natexlab{}.
\newblock \bibinfo{title}{eCommerceGAN : A Generative Adversarial Network for
  E-commerce}.
\newblock
\newblock
\showeprint[arxiv]{1801.03244}~[cs.LG]


\bibitem[\protect\citeauthoryear{Kumar and Rathore}{Kumar and Rathore}{2016}]%
        {kumar2016improving}
\bibfield{author}{\bibinfo{person}{Ashish Kumar} {and} \bibinfo{person}{NC
  Rathore}.} \bibinfo{year}{2016}\natexlab{}.
\newblock \showarticletitle{Improving attribute inference attack using link
  prediction in online social networks}.
\newblock In \bibinfo{booktitle}{\emph{Recent Advances in Mathematics,
  Statistics and Computer Science}}. \bibinfo{publisher}{World Scientific},
  \bibinfo{pages}{494--503}.
\newblock


\bibitem[\protect\citeauthoryear{Kurin, Godil, Whiteson, and Catanzaro}{Kurin
  et~al\mbox{.}}{2019}]%
        {kurin2019improving}
\bibfield{author}{\bibinfo{person}{Vitaly Kurin}, \bibinfo{person}{Saad Godil},
  \bibinfo{person}{Shimon Whiteson}, {and} \bibinfo{person}{Bryan Catanzaro}.}
  \bibinfo{year}{2019}\natexlab{}.
\newblock \showarticletitle{Improving SAT solver heuristics with graph networks
  and reinforcement learning}.
\newblock \bibinfo{journal}{\emph{arXiv preprint arXiv:1909.11830}}
  (\bibinfo{year}{2019}).
\newblock


\bibitem[\protect\citeauthoryear{Leetaru}{Leetaru}{2019}]%
        {DeepFake24:online}
\bibfield{author}{\bibinfo{person}{Kalev Leetaru}.}
  \bibinfo{year}{2019}\natexlab{}.
\newblock \bibinfo{title}{Deep Fakes' Greatest Threat Is Surveillance Video}.
\newblock
  \bibinfo{howpublished}{{https://www.forbes.com/sites/kalevleetaru/2019/08/26/deep-fakes-greatest-threat-is-surveillance-video/?sh=73c35a6c4550}}.
\newblock
\newblock
\shownote{(Accessed on 04/15/2021).}


\bibitem[\protect\citeauthoryear{Leong, Perez, and Dean}{Leong
  et~al\mbox{.}}{2019}]%
        {leong2019messagetap}
\bibfield{author}{\bibinfo{person}{R. Leong}, \bibinfo{person}{D. Perez}, {and}
  \bibinfo{person}{T. Dean}.} \bibinfo{year}{2019}\natexlab{}.
\newblock \bibinfo{title}{MESSAGETAP: Who’s Reading Your Text Messages?}
\newblock
\newblock
\urldef\tempurl%
\url{https://www.fireeye.com/blog/threat-research/2019/10/messagetap-who-is-reading-your-text-messages.html}
\showURL{%
\tempurl}


\bibitem[\protect\citeauthoryear{Lerman, Bontempi, and Markowitch}{Lerman
  et~al\mbox{.}}{2014}]%
        {lerman2014power}
\bibfield{author}{\bibinfo{person}{Liran Lerman}, \bibinfo{person}{Gianluca
  Bontempi}, {and} \bibinfo{person}{Olivier Markowitch}.}
  \bibinfo{year}{2014}\natexlab{}.
\newblock \showarticletitle{Power analysis attack: an approach based on machine
  learning}.
\newblock \bibinfo{journal}{\emph{International Journal of Applied
  Cryptography}} \bibinfo{volume}{3}, \bibinfo{number}{2}
  (\bibinfo{year}{2014}), \bibinfo{pages}{97--115}.
\newblock


\bibitem[\protect\citeauthoryear{Lerman, Bontempi, Taieb, and
  Markowitch}{Lerman et~al\mbox{.}}{2013}]%
        {lerman2013time}
\bibfield{author}{\bibinfo{person}{Liran Lerman}, \bibinfo{person}{Gianluca
  Bontempi}, \bibinfo{person}{Souhaib~Ben Taieb}, {and}
  \bibinfo{person}{Olivier Markowitch}.} \bibinfo{year}{2013}\natexlab{}.
\newblock \showarticletitle{A time series approach for profiling attack}. In
  \bibinfo{booktitle}{\emph{International Conference on Security, Privacy, and
  Applied Cryptography Engineering}}. Springer, \bibinfo{pages}{75--94}.
\newblock


\bibitem[\protect\citeauthoryear{Leslie, Harang, Knachel, and Kott}{Leslie
  et~al\mbox{.}}{2019}]%
        {cyberintrusionstats2019}
\bibfield{author}{\bibinfo{person}{Nandi~O. Leslie},
  \bibinfo{person}{Richard~E. Harang}, \bibinfo{person}{Lawrence~P. Knachel},
  {and} \bibinfo{person}{Alexander Kott}.} \bibinfo{year}{2019}\natexlab{}.
\newblock \showarticletitle{Statistical Models for the Number of Successful
  Cyber Intrusions}.
\newblock \bibinfo{journal}{\emph{CoRR}}  \bibinfo{volume}{abs/1901.04531}
  (\bibinfo{year}{2019}).
\newblock
\showeprint[arxiv]{1901.04531}
\urldef\tempurl%
\url{http://arxiv.org/abs/1901.04531}
\showURL{%
\tempurl}


\bibitem[\protect\citeauthoryear{Leviathan and Matias}{Leviathan and
  Matias}{2018}]%
        {leviathan2018google}
\bibfield{author}{\bibinfo{person}{Yaniv Leviathan} {and}
  \bibinfo{person}{Yossi Matias}.} \bibinfo{year}{2018}\natexlab{}.
\newblock \showarticletitle{Google Duplex: an AI system for accomplishing
  real-world tasks over the phone}.
\newblock  (\bibinfo{year}{2018}).
\newblock


\bibitem[\protect\citeauthoryear{Li, Shuai, Wang, and Tang}{Li
  et~al\mbox{.}}{2015}]%
        {li2015protocol}
\bibfield{author}{\bibinfo{person}{Haifeng Li}, \bibinfo{person}{Bo Shuai},
  \bibinfo{person}{Jian Wang}, {and} \bibinfo{person}{Chaojing Tang}.}
  \bibinfo{year}{2015}\natexlab{}.
\newblock \showarticletitle{Protocol reverse engineering using LDA and
  association analysis}. In \bibinfo{booktitle}{\emph{2015 11th International
  Conference on Computational Intelligence and Security (CIS)}}. IEEE,
  \bibinfo{pages}{312--316}.
\newblock


\bibitem[\protect\citeauthoryear{Li, Zhou, Li, Yan, and Zhu}{Li
  et~al\mbox{.}}{2019b}]%
        {li19-cns}
\bibfield{author}{\bibinfo{person}{J. Li}, \bibinfo{person}{L. Zhou},
  \bibinfo{person}{H. Li}, \bibinfo{person}{L. Yan}, {and} \bibinfo{person}{H.
  Zhu}.} \bibinfo{year}{2019}\natexlab{b}.
\newblock \showarticletitle{Dynamic {Traffic} {Feature} {Camouflaging} via
  {Generative} {Adversarial} {Networks}}. In \bibinfo{booktitle}{\emph{2019
  {IEEE} {Conference} on {Communications} and {Network} {Security} ({CNS})}}.
  \bibinfo{pages}{268--276}.
\newblock
\urldef\tempurl%
\url{https://doi.org/10.1109/CNS.2019.8802772}
\showDOI{\tempurl}


\bibitem[\protect\citeauthoryear{Li, Ma, Xue, and Zhao}{Li
  et~al\mbox{.}}{2020b}]%
        {li2020deep}
\bibfield{author}{\bibinfo{person}{Shaofeng Li}, \bibinfo{person}{Shiqing Ma},
  \bibinfo{person}{Minhui Xue}, {and} \bibinfo{person}{Benjamin Zi~Hao Zhao}.}
  \bibinfo{year}{2020}\natexlab{b}.
\newblock \showarticletitle{Deep Learning Backdoors}.
\newblock \bibinfo{journal}{\emph{arXiv preprint arXiv:2007.08273}}
  (\bibinfo{year}{2020}).
\newblock


\bibitem[\protect\citeauthoryear{Li, Ji, Lyu, Chen, Chen, Gu, Wu, and Beyah}{Li
  et~al\mbox{.}}{2020a}]%
        {li2020v}
\bibfield{author}{\bibinfo{person}{Yuwei Li}, \bibinfo{person}{Shouling Ji},
  \bibinfo{person}{Chenyang Lyu}, \bibinfo{person}{Yuan Chen},
  \bibinfo{person}{Jianhai Chen}, \bibinfo{person}{Qinchen Gu},
  \bibinfo{person}{Chunming Wu}, {and} \bibinfo{person}{Raheem Beyah}.}
  \bibinfo{year}{2020}\natexlab{a}.
\newblock \showarticletitle{V-Fuzz: Vulnerability Prediction-Assisted
  Evolutionary Fuzzing for Binary Programs}.
\newblock \bibinfo{journal}{\emph{IEEE Transactions on Cybernetics}}
  (\bibinfo{year}{2020}).
\newblock


\bibitem[\protect\citeauthoryear{Li, Wang, Wang, Ke, and Tan}{Li
  et~al\mbox{.}}{2020c}]%
        {li2020feature}
\bibfield{author}{\bibinfo{person}{Yuanzhang Li}, \bibinfo{person}{Yaxiao
  Wang}, \bibinfo{person}{Ye Wang}, \bibinfo{person}{Lishan Ke}, {and}
  \bibinfo{person}{Yu-an Tan}.} \bibinfo{year}{2020}\natexlab{c}.
\newblock \showarticletitle{A feature-vector generative adversarial network for
  evading PDF malware classifiers}.
\newblock \bibinfo{journal}{\emph{Information Sciences}}  \bibinfo{volume}{523}
  (\bibinfo{year}{2020}), \bibinfo{pages}{38--48}.
\newblock


\bibitem[\protect\citeauthoryear{Li, Yang, Wu, and Lyu}{Li
  et~al\mbox{.}}{2019a}]%
        {li2019hiding}
\bibfield{author}{\bibinfo{person}{Yuezun Li}, \bibinfo{person}{Xin Yang},
  \bibinfo{person}{Baoyuan Wu}, {and} \bibinfo{person}{Siwei Lyu}.}
  \bibinfo{year}{2019}\natexlab{a}.
\newblock \showarticletitle{Hiding faces in plain sight: Disrupting ai face
  synthesis with adversarial perturbations}.
\newblock \bibinfo{journal}{\emph{arXiv preprint arXiv:1906.09288}}
  (\bibinfo{year}{2019}).
\newblock


\bibitem[\protect\citeauthoryear{Li, Zou, Tang, Zhang, Sun, and Jin}{Li
  et~al\mbox{.}}{2019c}]%
        {li2019comparative}
\bibfield{author}{\bibinfo{person}{Zhen Li}, \bibinfo{person}{Deqing Zou},
  \bibinfo{person}{Jing Tang}, \bibinfo{person}{Zhihao Zhang},
  \bibinfo{person}{Mingqian Sun}, {and} \bibinfo{person}{Hai Jin}.}
  \bibinfo{year}{2019}\natexlab{c}.
\newblock \showarticletitle{A comparative study of deep learning-based
  vulnerability detection system}.
\newblock \bibinfo{journal}{\emph{IEEE Access}}  \bibinfo{volume}{7}
  (\bibinfo{year}{2019}), \bibinfo{pages}{103184--103197}.
\newblock


\bibitem[\protect\citeauthoryear{Li, Zou, Xu, Ou, Jin, Wang, Deng, and
  Zhong}{Li et~al\mbox{.}}{2018}]%
        {li2018vuldeepecker}
\bibfield{author}{\bibinfo{person}{Zhen Li}, \bibinfo{person}{Deqing Zou},
  \bibinfo{person}{Shouhuai Xu}, \bibinfo{person}{Xinyu Ou},
  \bibinfo{person}{Hai Jin}, \bibinfo{person}{Sujuan Wang},
  \bibinfo{person}{Zhijun Deng}, {and} \bibinfo{person}{Yuyi Zhong}.}
  \bibinfo{year}{2018}\natexlab{}.
\newblock \showarticletitle{Vuldeepecker: A deep learning-based system for
  vulnerability detection}.
\newblock \bibinfo{journal}{\emph{arXiv preprint arXiv:1801.01681}}
  (\bibinfo{year}{2018}).
\newblock


\bibitem[\protect\citeauthoryear{Liang, Oh, Mathew, Thomas, Li, and
  Ganesh}{Liang et~al\mbox{.}}{2018}]%
        {liang2018machine}
\bibfield{author}{\bibinfo{person}{Jia~Hui Liang}, \bibinfo{person}{Chanseok
  Oh}, \bibinfo{person}{Minu Mathew}, \bibinfo{person}{Ciza Thomas},
  \bibinfo{person}{Chunxiao Li}, {and} \bibinfo{person}{Vijay Ganesh}.}
  \bibinfo{year}{2018}\natexlab{}.
\newblock \showarticletitle{Machine learning-based restart policy for CDCL SAT
  solvers}. In \bibinfo{booktitle}{\emph{International Conference on Theory and
  Applications of Satisfiability Testing}}. Springer, \bibinfo{pages}{94--110}.
\newblock


\bibitem[\protect\citeauthoryear{Lim, Price, Monrose, and Frahm}{Lim
  et~al\mbox{.}}{2020}]%
        {lim2020revisiting}
\bibfield{author}{\bibinfo{person}{John Lim}, \bibinfo{person}{True Price},
  \bibinfo{person}{Fabian Monrose}, {and} \bibinfo{person}{Jan-Michael Frahm}.}
  \bibinfo{year}{2020}\natexlab{}.
\newblock \showarticletitle{Revisiting the Threat Space for Vision-based
  Keystroke Inference Attacks}.
\newblock \bibinfo{journal}{\emph{arXiv preprint arXiv:2009.05796}}
  (\bibinfo{year}{2020}).
\newblock


\bibitem[\protect\citeauthoryear{Lin, Wen, Han, Zhang, and Xiang}{Lin
  et~al\mbox{.}}{2020}]%
        {lin2020software}
\bibfield{author}{\bibinfo{person}{Guanjun Lin}, \bibinfo{person}{Sheng Wen},
  \bibinfo{person}{Qing-Long Han}, \bibinfo{person}{Jun Zhang}, {and}
  \bibinfo{person}{Yang Xiang}.} \bibinfo{year}{2020}\natexlab{}.
\newblock \showarticletitle{Software Vulnerability Detection Using Deep Neural
  Networks: A Survey}.
\newblock \bibinfo{journal}{\emph{Proc. IEEE}} \bibinfo{volume}{108},
  \bibinfo{number}{10} (\bibinfo{year}{2020}), \bibinfo{pages}{1825--1848}.
\newblock


\bibitem[\protect\citeauthoryear{Liu, Huo, Zhang, Li, Li, Piao, and Zou}{Liu
  et~al\mbox{.}}{2018}]%
        {liu2018alphadiff}
\bibfield{author}{\bibinfo{person}{Bingchang Liu}, \bibinfo{person}{Wei Huo},
  \bibinfo{person}{Chao Zhang}, \bibinfo{person}{Wenchao Li},
  \bibinfo{person}{Feng Li}, \bibinfo{person}{Aihua Piao}, {and}
  \bibinfo{person}{Wei Zou}.} \bibinfo{year}{2018}\natexlab{}.
\newblock \showarticletitle{$\alpha$diff: cross-version binary code similarity
  detection with dnn}. In \bibinfo{booktitle}{\emph{Proceedings of the 33rd
  ACM/IEEE International Conference on Automated Software Engineering}}.
  \bibinfo{pages}{667--678}.
\newblock


\bibitem[\protect\citeauthoryear{Liu and Lang}{Liu and Lang}{2019}]%
        {liu2019machine}
\bibfield{author}{\bibinfo{person}{Hongyu Liu} {and} \bibinfo{person}{Bo
  Lang}.} \bibinfo{year}{2019}\natexlab{}.
\newblock \showarticletitle{Machine learning and deep learning methods for
  intrusion detection systems: A survey}.
\newblock \bibinfo{journal}{\emph{applied sciences}} \bibinfo{volume}{9},
  \bibinfo{number}{20} (\bibinfo{year}{2019}), \bibinfo{pages}{4396}.
\newblock


\bibitem[\protect\citeauthoryear{Liu, Wang, Kar, Chen, Yang, and Gruteser}{Liu
  et~al\mbox{.}}{2015a}]%
        {liu2015snooping}
\bibfield{author}{\bibinfo{person}{Jian Liu}, \bibinfo{person}{Yan Wang},
  \bibinfo{person}{Gorkem Kar}, \bibinfo{person}{Yingying Chen},
  \bibinfo{person}{Jie Yang}, {and} \bibinfo{person}{Marco Gruteser}.}
  \bibinfo{year}{2015}\natexlab{a}.
\newblock \showarticletitle{Snooping keystrokes with mm-level audio ranging on
  a single phone}. In \bibinfo{booktitle}{\emph{Proceedings of the 21st Annual
  International Conference on Mobile Computing and Networking}}.
  \bibinfo{pages}{142--154}.
\newblock


\bibitem[\protect\citeauthoryear{Liu, Zhou, Diao, Li, and Zhang}{Liu
  et~al\mbox{.}}{2015b}]%
        {liu2015good}
\bibfield{author}{\bibinfo{person}{Xiangyu Liu}, \bibinfo{person}{Zhe Zhou},
  \bibinfo{person}{Wenrui Diao}, \bibinfo{person}{Zhou Li}, {and}
  \bibinfo{person}{Kehuan Zhang}.} \bibinfo{year}{2015}\natexlab{b}.
\newblock \showarticletitle{When good becomes evil: Keystroke inference with
  smartwatch}. In \bibinfo{booktitle}{\emph{Proceedings of the 22nd ACM SIGSAC
  Conference on Computer and Communications Security}}.
  \bibinfo{pages}{1273--1285}.
\newblock


\bibitem[\protect\citeauthoryear{Liu, Ma, Aafer, Lee, Zhai, Wang, and
  Zhang}{Liu et~al\mbox{.}}{2017}]%
        {liu2017trojaning}
\bibfield{author}{\bibinfo{person}{Yingqi Liu}, \bibinfo{person}{Shiqing Ma},
  \bibinfo{person}{Yousra Aafer}, \bibinfo{person}{Wen-Chuan Lee},
  \bibinfo{person}{Juan Zhai}, \bibinfo{person}{Weihang Wang}, {and}
  \bibinfo{person}{Xiangyu Zhang}.} \bibinfo{year}{2017}\natexlab{}.
\newblock \showarticletitle{Trojaning attack on neural networks}.
\newblock  (\bibinfo{year}{2017}).
\newblock


\bibitem[\protect\citeauthoryear{Lowd and Meek}{Lowd and Meek}{2005a}]%
        {lowd05}
\bibfield{author}{\bibinfo{person}{Daniel Lowd} {and}
  \bibinfo{person}{Christopher Meek}.} \bibinfo{year}{2005}\natexlab{a}.
\newblock \showarticletitle{Adversarial Learning}. In
  \bibinfo{booktitle}{\emph{Proc. 11th ACM SIGKDD International Conference on
  Knowledge Discovery and Data Mining (KDD)}}. \bibinfo{publisher}{ACM Press},
  \bibinfo{address}{Chicago, IL, USA}, \bibinfo{pages}{641--647}.
\newblock


\bibitem[\protect\citeauthoryear{Lowd and Meek}{Lowd and Meek}{2005b}]%
        {lowd05-ceas}
\bibfield{author}{\bibinfo{person}{Daniel Lowd} {and}
  \bibinfo{person}{Christopher Meek}.} \bibinfo{year}{2005}\natexlab{b}.
\newblock \showarticletitle{Good word attacks on statistical spam filters}. In
  \bibinfo{booktitle}{\emph{Second Conference on Email and Anti-Spam (CEAS)}}.
  \bibinfo{address}{Mountain View, CA, USA}.
\newblock


\bibitem[\protect\citeauthoryear{Lu, Yu, Chen, Zhu, Xu, Xue, and Li}{Lu
  et~al\mbox{.}}{2019}]%
        {lu2019keylisterber}
\bibfield{author}{\bibinfo{person}{Li Lu}, \bibinfo{person}{Jiadi Yu},
  \bibinfo{person}{Yingying Chen}, \bibinfo{person}{Yanmin Zhu},
  \bibinfo{person}{Xiangyu Xu}, \bibinfo{person}{Guangtao Xue}, {and}
  \bibinfo{person}{Minglu Li}.} \bibinfo{year}{2019}\natexlab{}.
\newblock \showarticletitle{KeyLiSterber: Inferring keystrokes on QWERTY
  keyboard of touch screen through acoustic signals}. In
  \bibinfo{booktitle}{\emph{IEEE INFOCOM 2019-IEEE Conference on Computer
  Communications}}. IEEE, \bibinfo{pages}{775--783}.
\newblock


\bibitem[\protect\citeauthoryear{Lunghi, Horejsi, and Pernet}{Lunghi
  et~al\mbox{.}}{2017}]%
        {lunghi2017untangling}
\bibfield{author}{\bibinfo{person}{Daniel Lunghi}, \bibinfo{person}{Jaromir
  Horejsi}, {and} \bibinfo{person}{Cedric Pernet}.}
  \bibinfo{year}{2017}\natexlab{}.
\newblock \bibinfo{title}{Untangling the Patchwork Cyberespionage Group}.
\newblock
\newblock
\urldef\tempurl%
\url{https://documents.trendmicro.com/assets/tech-brief-untangling-the-patchwork-cyberespionage-group.pdf}
\showURL{%
\tempurl}


\bibitem[\protect\citeauthoryear{Ma, Sheng, and Pant}{Ma et~al\mbox{.}}{2009}]%
        {ma09-DSC}
\bibfield{author}{\bibinfo{person}{Zhongming Ma}, \bibinfo{person}{Olivia
  Sheng}, {and} \bibinfo{person}{Gautam Pant}.}
  \bibinfo{year}{2009}\natexlab{}.
\newblock \showarticletitle{Discovering company revenue relations from news: A
  network approach}.
\newblock \bibinfo{journal}{\emph{Decision Support Systems}}
  \bibinfo{volume}{47} (\bibinfo{date}{11} \bibinfo{year}{2009}),
  \bibinfo{pages}{408--414}.
\newblock
\urldef\tempurl%
\url{https://doi.org/10.1016/j.dss.2009.04.007}
\showDOI{\tempurl}


\bibitem[\protect\citeauthoryear{Maghrebi, Portigliatti, and Prouff}{Maghrebi
  et~al\mbox{.}}{2016}]%
        {maghrebi2016breaking}
\bibfield{author}{\bibinfo{person}{Houssem Maghrebi}, \bibinfo{person}{Thibault
  Portigliatti}, {and} \bibinfo{person}{Emmanuel Prouff}.}
  \bibinfo{year}{2016}\natexlab{}.
\newblock \showarticletitle{Breaking cryptographic implementations using deep
  learning techniques}. In \bibinfo{booktitle}{\emph{International Conference
  on Security, Privacy, and Applied Cryptography Engineering}}. Springer,
  \bibinfo{pages}{3--26}.
\newblock


\bibitem[\protect\citeauthoryear{Mahadi, Mohamed, Mohamad, Makhtar, Kadir, and
  Mamat}{Mahadi et~al\mbox{.}}{2018}]%
        {mahadi2018survey}
\bibfield{author}{\bibinfo{person}{Nurul~Afnan Mahadi},
  \bibinfo{person}{Mohamad~Afendee Mohamed}, \bibinfo{person}{Amirul~Ihsan
  Mohamad}, \bibinfo{person}{Mokhairi Makhtar}, \bibinfo{person}{Mohd
  Fadzil~Abdul Kadir}, {and} \bibinfo{person}{Mustafa Mamat}.}
  \bibinfo{year}{2018}\natexlab{}.
\newblock \showarticletitle{A survey of machine learning techniques for
  behavioral-based biometric user authentication}.
\newblock \bibinfo{journal}{\emph{Recent Advances in Cryptography and Network
  Security}} (\bibinfo{year}{2018}), \bibinfo{pages}{43--54}.
\newblock


\bibitem[\protect\citeauthoryear{Maiorca, Demontis, Biggio, Roli, and
  Giacinto}{Maiorca et~al\mbox{.}}{2020}]%
        {maiorca20-cs}
\bibfield{author}{\bibinfo{person}{Davide Maiorca}, \bibinfo{person}{Ambra
  Demontis}, \bibinfo{person}{Battista Biggio}, \bibinfo{person}{Fabio Roli},
  {and} \bibinfo{person}{Giorgio Giacinto}.} \bibinfo{year}{2020}\natexlab{}.
\newblock \showarticletitle{Adversarial {Detection} of {Flash} {Malware}:
  {Limitations} and {Open} {Issues}}.
\newblock \bibinfo{journal}{\emph{Computers \& Security}}  \bibinfo{volume}{96}
  (\bibinfo{year}{2020}), \bibinfo{pages}{101901}.
\newblock
\showISSN{0167-4048}
\urldef\tempurl%
\url{https://doi.org/10.1016/j.cose.2020.101901}
\showDOI{\tempurl}


\bibitem[\protect\citeauthoryear{Maiti, Jadliwala, He, and Bilogrevic}{Maiti
  et~al\mbox{.}}{2018}]%
        {maiti2018side}
\bibfield{author}{\bibinfo{person}{Anindya Maiti}, \bibinfo{person}{Murtuza
  Jadliwala}, \bibinfo{person}{Jibo He}, {and} \bibinfo{person}{Igor
  Bilogrevic}.} \bibinfo{year}{2018}\natexlab{}.
\newblock \showarticletitle{Side-channel inference attacks on mobile keypads
  using smartwatches}.
\newblock \bibinfo{journal}{\emph{IEEE Transactions on Mobile Computing}}
  \bibinfo{volume}{17}, \bibinfo{number}{9} (\bibinfo{year}{2018}),
  \bibinfo{pages}{2180--2194}.
\newblock


\bibitem[\protect\citeauthoryear{Malhotra, Totti, Meira~Jr., Kumaraguru, and
  Almeida}{Malhotra et~al\mbox{.}}{2012}]%
        {malhotra12-ASONAM}
\bibfield{author}{\bibinfo{person}{Anshu Malhotra}, \bibinfo{person}{Luam
  Totti}, \bibinfo{person}{Wagner Meira~Jr.}, \bibinfo{person}{Ponnurangam
  Kumaraguru}, {and} \bibinfo{person}{Virgilio Almeida}.}
  \bibinfo{year}{2012}\natexlab{}.
\newblock \showarticletitle{Studying User Footprints in Different Online Social
  Networks}. In \bibinfo{booktitle}{\emph{Proceedings of the 2012 International
  Conference on Advances in Social Networks Analysis and Mining (ASONAM 2012)}}
  \emph{(\bibinfo{series}{ASONAM '12})}. \bibinfo{publisher}{IEEE Computer
  Society}, \bibinfo{address}{USA}, \bibinfo{pages}{1065–1070}.
\newblock
\showISBNx{9780769547992}
\urldef\tempurl%
\url{https://doi.org/10.1109/ASONAM.2012.184}
\showDOI{\tempurl}


\bibitem[\protect\citeauthoryear{Manning, Wong, Graham, Ranbaduge, Christen,
  Taylor, Wortley, Makkai, and Skorich}{Manning et~al\mbox{.}}{2018}]%
        {manning2018towards}
\bibfield{author}{\bibinfo{person}{Matthew Manning},
  \bibinfo{person}{Gabriel~TW Wong}, \bibinfo{person}{Timothy Graham},
  \bibinfo{person}{Thilina Ranbaduge}, \bibinfo{person}{Peter Christen},
  \bibinfo{person}{Kerry Taylor}, \bibinfo{person}{Richard Wortley},
  \bibinfo{person}{Toni Makkai}, {and} \bibinfo{person}{Pierre Skorich}.}
  \bibinfo{year}{2018}\natexlab{}.
\newblock \showarticletitle{Towards a ‘smart’cost--benefit tool: using
  machine learning to predict the costs of criminal justice policy
  interventions}.
\newblock \bibinfo{journal}{\emph{Crime Science}} \bibinfo{volume}{7},
  \bibinfo{number}{1} (\bibinfo{year}{2018}), \bibinfo{pages}{12}.
\newblock


\bibitem[\protect\citeauthoryear{Marquardt, Verma, Carter, and
  Traynor}{Marquardt et~al\mbox{.}}{2011}]%
        {marquardt2011sp}
\bibfield{author}{\bibinfo{person}{Philip Marquardt}, \bibinfo{person}{Arunabh
  Verma}, \bibinfo{person}{Henry Carter}, {and} \bibinfo{person}{Patrick
  Traynor}.} \bibinfo{year}{2011}\natexlab{}.
\newblock \showarticletitle{(sp) iphone: Decoding vibrations from nearby
  keyboards using mobile phone accelerometers}. In
  \bibinfo{booktitle}{\emph{Proceedings of the 18th ACM conference on Computer
  and communications security}}. \bibinfo{pages}{551--562}.
\newblock


\bibitem[\protect\citeauthoryear{Martorella}{Martorella}{2020}]%
        {laramies81:online}
\bibfield{author}{\bibinfo{person}{Christian Martorella}.}
  \bibinfo{year}{2020}\natexlab{}.
\newblock \bibinfo{title}{laramies/metagoofil: Metadata harvester}.
\newblock \bibinfo{howpublished}{{https://github.com/laramies/metagoofil}}.
\newblock
\newblock
\shownote{(Accessed on 10/20/2020).}


\bibitem[\protect\citeauthoryear{Matta, Cardarilli, Di~Nunzio, Fazzolari,
  Giardino, re, Silvestri, and Spanò}{Matta et~al\mbox{.}}{2019}]%
        {matta19-el}
\bibfield{author}{\bibinfo{person}{Marco Matta}, \bibinfo{person}{Gian~Carlo
  Cardarilli}, \bibinfo{person}{Luca Di~Nunzio}, \bibinfo{person}{Rocco
  Fazzolari}, \bibinfo{person}{Daniele Giardino}, \bibinfo{person}{Marco re},
  \bibinfo{person}{Francesca Silvestri}, {and} \bibinfo{person}{Sergio
  Spanò}.} \bibinfo{year}{2019}\natexlab{}.
\newblock \showarticletitle{Q-RTS: a Real-Time Swarm Intelligence based on
  Multi-Agent Q-Learning}.
\newblock \bibinfo{journal}{\emph{Electronics Letters}} (\bibinfo{date}{03}
  \bibinfo{year}{2019}).
\newblock
\urldef\tempurl%
\url{https://doi.org/10.1049/el.2019.0244}
\showDOI{\tempurl}


\bibitem[\protect\citeauthoryear{Mattei}{Mattei}{2017}]%
        {mattei17}
\bibfield{author}{\bibinfo{person}{Tobias~A Mattei}.}
  \bibinfo{year}{2017}\natexlab{}.
\newblock \showarticletitle{Privacy, Confidentiality, and Security of Health
  Care Information: Lessons from the Recent WannaCry Cyberattack}.
\newblock \bibinfo{journal}{\emph{World neurosurgery}}  \bibinfo{volume}{104}
  (\bibinfo{date}{August} \bibinfo{year}{2017}), \bibinfo{pages}{972—974}.
\newblock
\showISSN{1878-8750}
\urldef\tempurl%
\url{https://doi.org/10.1016/j.wneu.2017.06.104}
\showDOI{\tempurl}


\bibitem[\protect\citeauthoryear{Melis, Demontis, Pintor, Sotgiu, and
  Biggio}{Melis et~al\mbox{.}}{2019}]%
        {melis2019secml}
\bibfield{author}{\bibinfo{person}{Marco Melis}, \bibinfo{person}{Ambra
  Demontis}, \bibinfo{person}{Maura Pintor}, \bibinfo{person}{Angelo Sotgiu},
  {and} \bibinfo{person}{Battista Biggio}.} \bibinfo{year}{2019}\natexlab{}.
\newblock \showarticletitle{secml: A Python Library for Secure and Explainable
  Machine Learning}.
\newblock \bibinfo{journal}{\emph{arXiv preprint arXiv:1912.10013}}
  (\bibinfo{year}{2019}).
\newblock


\bibitem[\protect\citeauthoryear{Messaoud, Guennoun, Wahbi, and Sadik}{Messaoud
  et~al\mbox{.}}{2016}]%
        {messaoud2016advanced}
\bibfield{author}{\bibinfo{person}{Brahim~ID Messaoud}, \bibinfo{person}{Karim
  Guennoun}, \bibinfo{person}{Mohamed Wahbi}, {and} \bibinfo{person}{Mohamed
  Sadik}.} \bibinfo{year}{2016}\natexlab{}.
\newblock \showarticletitle{Advanced persistent threat: new analysis driven by
  life cycle phases and their challenges}. In \bibinfo{booktitle}{\emph{2016
  International Conference on Advanced Communication Systems and Information
  Security (ACOSIS)}}. IEEE, \bibinfo{pages}{1--6}.
\newblock


\bibitem[\protect\citeauthoryear{Mirsky, Doitshman, Elovici, and
  Shabtai}{Mirsky et~al\mbox{.}}{2018}]%
        {mirsky2018kitsune}
\bibfield{author}{\bibinfo{person}{Yisroel Mirsky}, \bibinfo{person}{Tomer
  Doitshman}, \bibinfo{person}{Yuval Elovici}, {and} \bibinfo{person}{Asaf
  Shabtai}.} \bibinfo{year}{2018}\natexlab{}.
\newblock \showarticletitle{Kitsune: an ensemble of autoencoders for online
  network intrusion detection}.
\newblock \bibinfo{journal}{\emph{arXiv preprint arXiv:1802.09089}}
  (\bibinfo{year}{2018}).
\newblock


\bibitem[\protect\citeauthoryear{Mirsky and Lee}{Mirsky and Lee}{2021}]%
        {mirsky2021creation}
\bibfield{author}{\bibinfo{person}{Yisroel Mirsky} {and} \bibinfo{person}{Wenke
  Lee}.} \bibinfo{year}{2021}\natexlab{}.
\newblock \showarticletitle{The creation and detection of deepfakes: A survey}.
\newblock \bibinfo{journal}{\emph{ACM Computing Surveys (CSUR)}}
  \bibinfo{volume}{54}, \bibinfo{number}{1} (\bibinfo{year}{2021}),
  \bibinfo{pages}{1--41}.
\newblock


\bibitem[\protect\citeauthoryear{Mirsky, Mahler, Shelef, and Elovici}{Mirsky
  et~al\mbox{.}}{2019}]%
        {236284}
\bibfield{author}{\bibinfo{person}{Yisroel Mirsky}, \bibinfo{person}{Tom
  Mahler}, \bibinfo{person}{Ilan Shelef}, {and} \bibinfo{person}{Yuval
  Elovici}.} \bibinfo{year}{2019}\natexlab{}.
\newblock \showarticletitle{CT-GAN: Malicious Tampering of 3D Medical Imagery
  using Deep Learning}. In \bibinfo{booktitle}{\emph{28th {USENIX} Security
  Symposium ({USENIX} Security 19)}}. \bibinfo{publisher}{{USENIX}
  Association}, \bibinfo{address}{Santa Clara, CA}, \bibinfo{pages}{461--478}.
\newblock
\showISBNx{978-1-939133-06-9}
\urldef\tempurl%
\url{https://www.usenix.org/conference/usenixsecurity19/presentation/mirsky}
\showURL{%
\tempurl}


\bibitem[\protect\citeauthoryear{Mokhov, Paquet, and Debbabi}{Mokhov
  et~al\mbox{.}}{2014}]%
        {NLP_static}
\bibfield{author}{\bibinfo{person}{Serguei~A. Mokhov}, \bibinfo{person}{Joey
  Paquet}, {and} \bibinfo{person}{Mourad Debbabi}.}
  \bibinfo{year}{2014}\natexlab{}.
\newblock \showarticletitle{The Use of NLP Techniques in Static Code Analysis
  to Detect Weaknesses and Vulnerabilities}. In
  \bibinfo{booktitle}{\emph{Advances in Artificial Intelligence}},
  \bibfield{editor}{\bibinfo{person}{Marina Sokolova} {and}
  \bibinfo{person}{Peter van Beek}} (Eds.). \bibinfo{publisher}{Springer
  International Publishing}, \bibinfo{address}{Cham},
  \bibinfo{pages}{326--332}.
\newblock
\showISBNx{978-3-319-06483-3}


\bibitem[\protect\citeauthoryear{Monaco}{Monaco}{2019}]%
        {236286}
\bibfield{author}{\bibinfo{person}{John~V. Monaco}.}
  \bibinfo{year}{2019}\natexlab{}.
\newblock \showarticletitle{What Are You Searching For? A Remote Keylogging
  Attack on Search Engine Autocomplete}. In \bibinfo{booktitle}{\emph{28th
  {USENIX} Security Symposium ({USENIX} Security 19)}}.
  \bibinfo{publisher}{{USENIX} Association}, \bibinfo{address}{Santa Clara,
  CA}, \bibinfo{pages}{959--976}.
\newblock
\showISBNx{978-1-939133-06-9}
\urldef\tempurl%
\url{https://www.usenix.org/conference/usenixsecurity19/presentation/monaco}
\showURL{%
\tempurl}


\bibitem[\protect\citeauthoryear{Mozur}{Mozur}{2018}]%
        {chinaSurv}
\bibfield{author}{\bibinfo{person}{Paul Mozur}.}
  \bibinfo{year}{2018}\natexlab{}.
\newblock \bibinfo{title}{Looking Through the Eyes of China’s Surveillance
  State}.
\newblock
\newblock
\urldef\tempurl%
\url{https://www.nytimes.com/2018/07/16/technology/china-surveillance-state.html}
\showURL{%
\tempurl}
\newblock
\shownote{accessed: June 2018.}


\bibitem[\protect\citeauthoryear{Mueller}{Mueller}{2018}]%
        {mueller2018indictment}
\bibfield{author}{\bibinfo{person}{R. Mueller}.}
  \bibinfo{year}{2018}\natexlab{}.
\newblock \bibinfo{title}{Indictment - United States of America vs. VIKTOR
  BORISOVICH NETYKSHO, et al.}
\newblock
\newblock
\urldef\tempurl%
\url{{https://www.justice.gov/file/1080281/download}}
\showURL{%
\tempurl}


\bibitem[\protect\citeauthoryear{Muñoz-González, Biggio, Demontis, Paudice,
  Wongrassamee, Lupu, and Roli}{Muñoz-González et~al\mbox{.}}{2017}]%
        {munoz-gonzalez17-aisec}
\bibfield{author}{\bibinfo{person}{Luis Muñoz-González},
  \bibinfo{person}{Battista Biggio}, \bibinfo{person}{Ambra Demontis},
  \bibinfo{person}{Andrea Paudice}, \bibinfo{person}{Vasin Wongrassamee},
  \bibinfo{person}{Emil~C. Lupu}, {and} \bibinfo{person}{Fabio Roli}.}
  \bibinfo{year}{2017}\natexlab{}.
\newblock \showarticletitle{Towards {Poisoning} of {Deep} {Learning}
  {Algorithms} with {Back}-gradient {Optimization}}. In
  \bibinfo{booktitle}{\emph{10th {ACM} {Workshop} on {Artificial}
  {Intelligence} and {Security}}} \emph{(\bibinfo{series}{{AISec} '17})},
  \bibfield{editor}{\bibinfo{person}{Bhavani~M. Thuraisingham},
  \bibinfo{person}{Battista Biggio}, \bibinfo{person}{David~Mandell Freeman},
  \bibinfo{person}{Brad Miller}, {and} \bibinfo{person}{Arunesh Sinha}} (Eds.).
  \bibinfo{publisher}{ACM}, \bibinfo{address}{New York, NY, USA},
  \bibinfo{pages}{27--38}.
\newblock


\bibitem[\protect\citeauthoryear{Nam, Jeon, Kim, and Moon}{Nam
  et~al\mbox{.}}{2020}]%
        {nam2020recurrent}
\bibfield{author}{\bibinfo{person}{Sungyup Nam}, \bibinfo{person}{Seungho
  Jeon}, \bibinfo{person}{Hongkyo Kim}, {and} \bibinfo{person}{Jongsub Moon}.}
  \bibinfo{year}{2020}\natexlab{}.
\newblock \showarticletitle{Recurrent GANs Password Cracker For IoT Password
  Security Enhancement}.
\newblock \bibinfo{journal}{\emph{Sensors}} \bibinfo{volume}{20},
  \bibinfo{number}{11} (\bibinfo{year}{2020}), \bibinfo{pages}{3106}.
\newblock


\bibitem[\protect\citeauthoryear{{Narayanan} and {Shmatikov}}{{Narayanan} and
  {Shmatikov}}{2008}]%
        {Narayanan08-sp}
\bibfield{author}{\bibinfo{person}{A. {Narayanan}} {and} \bibinfo{person}{V.
  {Shmatikov}}.} \bibinfo{year}{2008}\natexlab{}.
\newblock \showarticletitle{Robust De-anonymization of Large Sparse Datasets}.
  In \bibinfo{booktitle}{\emph{2008 IEEE Symposium on Security and Privacy (sp
  2008)}}. \bibinfo{pages}{111--125}.
\newblock
\urldef\tempurl%
\url{https://doi.org/10.1109/SP.2008.33}
\showDOI{\tempurl}


\bibitem[\protect\citeauthoryear{Nasar, Jaffry, and Malik}{Nasar
  et~al\mbox{.}}{2019}]%
        {nasar2019textual}
\bibfield{author}{\bibinfo{person}{Zara Nasar}, \bibinfo{person}{Syed~Waqar
  Jaffry}, {and} \bibinfo{person}{Muhammad~Kamran Malik}.}
  \bibinfo{year}{2019}\natexlab{}.
\newblock \showarticletitle{Textual keyword extraction and summarization:
  State-of-the-art}.
\newblock \bibinfo{journal}{\emph{Information Processing \& Management}}
  \bibinfo{volume}{56}, \bibinfo{number}{6} (\bibinfo{year}{2019}),
  \bibinfo{pages}{102088}.
\newblock


\bibitem[\protect\citeauthoryear{Nicolae, Sinn, Tran, Buesser, Rawat, Wistuba,
  Zantedeschi, Baracaldo, Chen, Ludwig, Molloy, and Edwards}{Nicolae
  et~al\mbox{.}}{2018}]%
        {art2018-arxiv}
\bibfield{author}{\bibinfo{person}{Maria-Irina Nicolae},
  \bibinfo{person}{Mathieu Sinn}, \bibinfo{person}{Minh~Ngoc Tran},
  \bibinfo{person}{Beat Buesser}, \bibinfo{person}{Ambrish Rawat},
  \bibinfo{person}{Martin Wistuba}, \bibinfo{person}{Valentina Zantedeschi},
  \bibinfo{person}{Nathalie Baracaldo}, \bibinfo{person}{Bryant Chen},
  \bibinfo{person}{Heiko Ludwig}, \bibinfo{person}{Ian Molloy}, {and}
  \bibinfo{person}{Ben Edwards}.} \bibinfo{year}{2018}\natexlab{}.
\newblock \showarticletitle{Adversarial Robustness Toolbox v1.2.0}.
\newblock \bibinfo{journal}{\emph{CoRR}}  \bibinfo{volume}{1807.01069}
  (\bibinfo{year}{2018}).
\newblock
\urldef\tempurl%
\url{https://arxiv.org/pdf/1807.01069}
\showURL{%
\tempurl}


\bibitem[\protect\citeauthoryear{Nirkin, Keller, and Hassner}{Nirkin
  et~al\mbox{.}}{2019}]%
        {nirkin2019fsgan}
\bibfield{author}{\bibinfo{person}{Yuval Nirkin}, \bibinfo{person}{Yosi
  Keller}, {and} \bibinfo{person}{Tal Hassner}.}
  \bibinfo{year}{2019}\natexlab{}.
\newblock \showarticletitle{Fsgan: Subject agnostic face swapping and
  reenactment}. In \bibinfo{booktitle}{\emph{Proceedings of the IEEE/CVF
  International Conference on Computer Vision}}. \bibinfo{pages}{7184--7193}.
\newblock


\bibitem[\protect\citeauthoryear{Novo and Morla}{Novo and Morla}{2020}]%
        {novo20-aisec}
\bibfield{author}{\bibinfo{person}{Carlos Novo} {and} \bibinfo{person}{Ricardo
  Morla}.} \bibinfo{year}{2020}\natexlab{}.
\newblock \showarticletitle{Flow-based {Detection} and {Proxy}-based {Evasion}
  of {Encrypted} {Malware} {C2} {Traffic}}. In
  \bibinfo{booktitle}{\emph{Proceedings of the 13th {ACM} {Workshop} on
  {Artificial} {Intelligence} and {Security}}}
  \emph{(\bibinfo{series}{{AISec}'20})}. \bibinfo{publisher}{Association for
  Computing Machinery}, \bibinfo{address}{New York, NY, USA},
  \bibinfo{pages}{83--91}.
\newblock
\showISBNx{978-1-4503-8094-2}
\urldef\tempurl%
\url{https://doi.org/10.1145/3411508.3421379}
\showDOI{\tempurl}


\bibitem[\protect\citeauthoryear{Otsuka, Watanabe, and Matsumoto}{Otsuka
  et~al\mbox{.}}{2015}]%
        {otsuka2015learning}
\bibfield{author}{\bibinfo{person}{Hiroshi Otsuka}, \bibinfo{person}{Yukihiro
  Watanabe}, {and} \bibinfo{person}{Yasuhide Matsumoto}.}
  \bibinfo{year}{2015}\natexlab{}.
\newblock \showarticletitle{Learning from before and after recovery to detect
  latent misconfiguration}. In \bibinfo{booktitle}{\emph{2015 IEEE 39th Annual
  Computer Software and Applications Conference}}, Vol.~\bibinfo{volume}{3}.
  IEEE, \bibinfo{pages}{141--148}.
\newblock


\bibitem[\protect\citeauthoryear{Ou, Govindavajhala, and Appel}{Ou
  et~al\mbox{.}}{2005}]%
        {ou05-usenix-mulval}
\bibfield{author}{\bibinfo{person}{X. Ou}, \bibinfo{person}{Sudhakar
  Govindavajhala}, {and} \bibinfo{person}{Andrew~W. Appel}.}
  \bibinfo{year}{2005}\natexlab{}.
\newblock \showarticletitle{{MulVAL}: {A} {Logic}-based {Network} {Security}
  {Analyzer}}. In \bibinfo{booktitle}{\emph{{USENIX} {Security} {Symposium}}}.
\newblock


\bibitem[\protect\citeauthoryear{Oxylabs}{Oxylabs}{2021}]%
        {Innovati92:online}
\bibfield{author}{\bibinfo{person}{Oxylabs}.} \bibinfo{year}{2021}\natexlab{}.
\newblock \bibinfo{title}{Innovative Proxy Service to Gather Data at Scale}.
\newblock \bibinfo{howpublished}{{https://oxylabs.io/}}.
\newblock
\newblock
\shownote{(Accessed on 04/14/2021).}


\bibitem[\protect\citeauthoryear{Panagiotou, Ghita, Shiaeles, and
  Bendiab}{Panagiotou et~al\mbox{.}}{2019}]%
        {facewall_2019_Panagiotou}
\bibfield{author}{\bibinfo{person}{Aimilia Panagiotou}, \bibinfo{person}{Bogdan
  Ghita}, \bibinfo{person}{Stavros Shiaeles}, {and} \bibinfo{person}{Keltoum
  Bendiab}.} \bibinfo{year}{2019}\natexlab{}.
\newblock \showarticletitle{FaceWallGraph: Using Machine Learning for Profiling
  User Behaviour from Facebook Wall}. In \bibinfo{booktitle}{\emph{Internet of
  Things, Smart Spaces, and Next Generation Networks and Systems}},
  \bibfield{editor}{\bibinfo{person}{Olga Galinina}, \bibinfo{person}{Sergey
  Andreev}, \bibinfo{person}{Sergey Balandin}, {and} \bibinfo{person}{Yevgeni
  Koucheryavy}} (Eds.). \bibinfo{publisher}{Springer International Publishing},
  \bibinfo{address}{Cham}, \bibinfo{pages}{125--134}.
\newblock
\showISBNx{978-3-030-30859-9}


\bibitem[\protect\citeauthoryear{Papernot, Faghri, Carlini, Goodfellow,
  Feinman, Kurakin, Xie, Sharma, Brown, Roy, Matyasko, Behzadan, Hambardzumyan,
  Zhang, Juang, Li, Sheatsley, Garg, Uesato, Gierke, Dong, Berthelot,
  Hendricks, Rauber, and Long}{Papernot et~al\mbox{.}}{2018}]%
        {papernot2018cleverhans}
\bibfield{author}{\bibinfo{person}{Nicolas Papernot}, \bibinfo{person}{Fartash
  Faghri}, \bibinfo{person}{Nicholas Carlini}, \bibinfo{person}{Ian
  Goodfellow}, \bibinfo{person}{Reuben Feinman}, \bibinfo{person}{Alexey
  Kurakin}, \bibinfo{person}{Cihang Xie}, \bibinfo{person}{Yash Sharma},
  \bibinfo{person}{Tom Brown}, \bibinfo{person}{Aurko Roy},
  \bibinfo{person}{Alexander Matyasko}, \bibinfo{person}{Vahid Behzadan},
  \bibinfo{person}{Karen Hambardzumyan}, \bibinfo{person}{Zhishuai Zhang},
  \bibinfo{person}{Yi-Lin Juang}, \bibinfo{person}{Zhi Li},
  \bibinfo{person}{Ryan Sheatsley}, \bibinfo{person}{Abhibhav Garg},
  \bibinfo{person}{Jonathan Uesato}, \bibinfo{person}{Willi Gierke},
  \bibinfo{person}{Yinpeng Dong}, \bibinfo{person}{David Berthelot},
  \bibinfo{person}{Paul Hendricks}, \bibinfo{person}{Jonas Rauber}, {and}
  \bibinfo{person}{Rujun Long}.} \bibinfo{year}{2018}\natexlab{}.
\newblock \showarticletitle{Technical Report on the CleverHans v2.1.0
  Adversarial Examples Library}.
\newblock \bibinfo{journal}{\emph{arXiv preprint arXiv:1610.00768}}
  (\bibinfo{year}{2018}).
\newblock


\bibitem[\protect\citeauthoryear{{Papernot}, {McDaniel}, {Sinha}, and
  {Wellman}}{{Papernot} et~al\mbox{.}}{2018}]%
        {Papernot18-eurosp}
\bibfield{author}{\bibinfo{person}{N. {Papernot}}, \bibinfo{person}{P.
  {McDaniel}}, \bibinfo{person}{A. {Sinha}}, {and} \bibinfo{person}{M.~P.
  {Wellman}}.} \bibinfo{year}{2018}\natexlab{}.
\newblock \showarticletitle{SoK: Security and Privacy in Machine Learning}. In
  \bibinfo{booktitle}{\emph{2018 IEEE European Symposium on Security and
  Privacy (EuroS P)}}. \bibinfo{pages}{399--414}.
\newblock
\urldef\tempurl%
\url{https://doi.org/10.1109/EuroSP.2018.00035}
\showDOI{\tempurl}


\bibitem[\protect\citeauthoryear{Pasandi, Nazarian, and Pedram}{Pasandi
  et~al\mbox{.}}{2019}]%
        {pasandi2019approximate}
\bibfield{author}{\bibinfo{person}{Ghasem Pasandi}, \bibinfo{person}{Shahin
  Nazarian}, {and} \bibinfo{person}{Massoud Pedram}.}
  \bibinfo{year}{2019}\natexlab{}.
\newblock \showarticletitle{Approximate logic synthesis: A reinforcement
  learning-based technology mapping approach}. In
  \bibinfo{booktitle}{\emph{20th International Symposium on Quality Electronic
  Design (ISQED)}}. IEEE, \bibinfo{pages}{26--32}.
\newblock


\bibitem[\protect\citeauthoryear{Patel, Suthar, and Thakar}{Patel
  et~al\mbox{.}}{2019}]%
        {patel2019survey}
\bibfield{author}{\bibinfo{person}{Manish~I Patel}, \bibinfo{person}{Sirali
  Suthar}, {and} \bibinfo{person}{Jil Thakar}.}
  \bibinfo{year}{2019}\natexlab{}.
\newblock \showarticletitle{Survey on Image Compression using Machine Learning
  and Deep Learning}. In \bibinfo{booktitle}{\emph{2019 International
  Conference on Intelligent Computing and Control Systems (ICCS)}}. IEEE,
  \bibinfo{pages}{1103--1105}.
\newblock


\bibitem[\protect\citeauthoryear{Peck, Nie, Sivaguru, Grumer, Olumofin, Yu,
  Nascimento, and De~Cock}{Peck et~al\mbox{.}}{2019}]%
        {peck2019charbot}
\bibfield{author}{\bibinfo{person}{Jonathan Peck}, \bibinfo{person}{Claire
  Nie}, \bibinfo{person}{Raaghavi Sivaguru}, \bibinfo{person}{Charles Grumer},
  \bibinfo{person}{Femi Olumofin}, \bibinfo{person}{Bin Yu},
  \bibinfo{person}{Anderson Nascimento}, {and} \bibinfo{person}{Martine
  De~Cock}.} \bibinfo{year}{2019}\natexlab{}.
\newblock \showarticletitle{CharBot: A simple and effective method for evading
  DGA classifiers}.
\newblock \bibinfo{journal}{\emph{IEEE Access}}  \bibinfo{volume}{7}
  (\bibinfo{year}{2019}), \bibinfo{pages}{91759--91771}.
\newblock


\bibitem[\protect\citeauthoryear{Pellet, Shiaeles, and Stavrou}{Pellet
  et~al\mbox{.}}{2019}]%
        {pellet2019localising}
\bibfield{author}{\bibinfo{person}{Hector Pellet}, \bibinfo{person}{Stavros
  Shiaeles}, {and} \bibinfo{person}{Stavros Stavrou}.}
  \bibinfo{year}{2019}\natexlab{}.
\newblock \showarticletitle{Localising social network users and profiling their
  movement}.
\newblock \bibinfo{journal}{\emph{Computers \& Security}}  \bibinfo{volume}{81}
  (\bibinfo{year}{2019}), \bibinfo{pages}{49--57}.
\newblock


\bibitem[\protect\citeauthoryear{Perianin, Carr{\'e}, Dyseryn, Facon, and
  Guilley}{Perianin et~al\mbox{.}}{2020}]%
        {perianin2020end}
\bibfield{author}{\bibinfo{person}{Thomas Perianin}, \bibinfo{person}{Sebastien
  Carr{\'e}}, \bibinfo{person}{Victor Dyseryn}, \bibinfo{person}{Adrien Facon},
  {and} \bibinfo{person}{Sylvain Guilley}.} \bibinfo{year}{2020}\natexlab{}.
\newblock \showarticletitle{End-to-end automated cache-timing attack driven by
  Machine Learning}.
\newblock \bibinfo{journal}{\emph{Journal of Cryptographic Engineering}}
  (\bibinfo{year}{2020}), \bibinfo{pages}{1--12}.
\newblock


\bibitem[\protect\citeauthoryear{Perin, Chmielewski, Batina, and Picek}{Perin
  et~al\mbox{.}}{2020}]%
        {Perin_Chmielewski_Batina_Picek_2020}
\bibfield{author}{\bibinfo{person}{Guilherme Perin}, \bibinfo{person}{Łukasz
  Chmielewski}, \bibinfo{person}{Lejla Batina}, {and} \bibinfo{person}{Stjepan
  Picek}.} \bibinfo{year}{2020}\natexlab{}.
\newblock \showarticletitle{Keep it Unsupervised: Horizontal Attacks Meet Deep
  Learning}.
\newblock \bibinfo{journal}{\emph{IACR Transactions on Cryptographic Hardware
  and Embedded Systems}} \bibinfo{volume}{2021}, \bibinfo{number}{1}
  (\bibinfo{date}{Dec.} \bibinfo{year}{2020}), \bibinfo{pages}{343--372}.
\newblock
\urldef\tempurl%
\url{https://doi.org/10.46586/tches.v2021.i1.343-372}
\showDOI{\tempurl}


\bibitem[\protect\citeauthoryear{Picek, Heuser, Jovic, Bhasin, and
  Regazzoni}{Picek et~al\mbox{.}}{2019}]%
        {picek2019curse}
\bibfield{author}{\bibinfo{person}{Stjepan Picek}, \bibinfo{person}{Annelie
  Heuser}, \bibinfo{person}{Alan Jovic}, \bibinfo{person}{Shivam Bhasin}, {and}
  \bibinfo{person}{Francesco Regazzoni}.} \bibinfo{year}{2019}\natexlab{}.
\newblock \showarticletitle{The curse of class imbalance and conflicting
  metrics with machine learning for side-channel evaluations}.
\newblock \bibinfo{journal}{\emph{IACR Transactions on Cryptographic Hardware
  and Embedded Systems}} \bibinfo{volume}{2019}, \bibinfo{number}{1}
  (\bibinfo{year}{2019}), \bibinfo{pages}{1--29}.
\newblock


\bibitem[\protect\citeauthoryear{Picek, Samiotis, Kim, Heuser, Bhasin, and
  Legay}{Picek et~al\mbox{.}}{2018}]%
        {picek2018performance}
\bibfield{author}{\bibinfo{person}{Stjepan Picek},
  \bibinfo{person}{Ioannis~Petros Samiotis}, \bibinfo{person}{Jaehun Kim},
  \bibinfo{person}{Annelie Heuser}, \bibinfo{person}{Shivam Bhasin}, {and}
  \bibinfo{person}{Axel Legay}.} \bibinfo{year}{2018}\natexlab{}.
\newblock \showarticletitle{On the performance of convolutional neural networks
  for side-channel analysis}. In \bibinfo{booktitle}{\emph{International
  Conference on Security, Privacy, and Applied Cryptography Engineering}}.
  Springer, \bibinfo{pages}{157--176}.
\newblock


\bibitem[\protect\citeauthoryear{{Pierazzi}, {Pendlebury}, {Cortellazzi}, and
  {Cavallaro}}{{Pierazzi} et~al\mbox{.}}{2020}]%
        {pierazzi20-sp}
\bibfield{author}{\bibinfo{person}{F. {Pierazzi}}, \bibinfo{person}{F.
  {Pendlebury}}, \bibinfo{person}{J. {Cortellazzi}}, {and} \bibinfo{person}{L.
  {Cavallaro}}.} \bibinfo{year}{2020}\natexlab{}.
\newblock \showarticletitle{Intriguing Properties of Adversarial ML Attacks in
  the Problem Space}. In \bibinfo{booktitle}{\emph{2020 IEEE Symposium on
  Security and Privacy (SP)}}. \bibinfo{pages}{1332--1349}.
\newblock
\showISSN{2375-1207}
\urldef\tempurl%
\url{https://doi.org/10.1109/SP40000.2020.00073}
\showDOI{\tempurl}


\bibitem[\protect\citeauthoryear{{Rahman}, {Rochan}, and {Wang}}{{Rahman}
  et~al\mbox{.}}{2019}]%
        {Rahman19-avss}
\bibfield{author}{\bibinfo{person}{T. {Rahman}}, \bibinfo{person}{M. {Rochan}},
  {and} \bibinfo{person}{Y. {Wang}}.} \bibinfo{year}{2019}\natexlab{}.
\newblock \showarticletitle{Video-Based Person Re-Identification using Refined
  Attention Networks}. In \bibinfo{booktitle}{\emph{2019 16th IEEE
  International Conference on Advanced Video and Signal Based Surveillance
  (AVSS)}}. \bibinfo{pages}{1--8}.
\newblock
\urldef\tempurl%
\url{https://doi.org/10.1109/AVSS.2019.8909869}
\showDOI{\tempurl}


\bibitem[\protect\citeauthoryear{{Rathi}, {Malik}, {Varshney}, {Sharma}, and
  {Mendiratta}}{{Rathi} et~al\mbox{.}}{2018}]%
        {8530517}
\bibfield{author}{\bibinfo{person}{M. {Rathi}}, \bibinfo{person}{A. {Malik}},
  \bibinfo{person}{D. {Varshney}}, \bibinfo{person}{R. {Sharma}}, {and}
  \bibinfo{person}{S. {Mendiratta}}.} \bibinfo{year}{2018}\natexlab{}.
\newblock \showarticletitle{Sentiment Analysis of Tweets Using Machine Learning
  Approach}. In \bibinfo{booktitle}{\emph{2018 Eleventh International
  Conference on Contemporary Computing (IC3)}}. \bibinfo{pages}{1--3}.
\newblock
\urldef\tempurl%
\url{https://doi.org/10.1109/IC3.2018.8530517}
\showDOI{\tempurl}


\bibitem[\protect\citeauthoryear{Rebryk and Beliaev}{Rebryk and
  Beliaev}{2020}]%
        {rebryk2020convoice}
\bibfield{author}{\bibinfo{person}{Yurii Rebryk} {and}
  \bibinfo{person}{Stanislav Beliaev}.} \bibinfo{year}{2020}\natexlab{}.
\newblock \showarticletitle{ConVoice: Real-Time Zero-Shot Voice Style Transfer
  with Convolutional Network}.
\newblock \bibinfo{journal}{\emph{arXiv preprint arXiv:2005.07815}}
  (\bibinfo{year}{2020}).
\newblock


\bibitem[\protect\citeauthoryear{Ren, Tan, Qin, Zhao, Zhao, and Liu}{Ren
  et~al\mbox{.}}{2019}]%
        {ren2019almost}
\bibfield{author}{\bibinfo{person}{Yi Ren}, \bibinfo{person}{Xu Tan},
  \bibinfo{person}{Tao Qin}, \bibinfo{person}{Sheng Zhao},
  \bibinfo{person}{Zhou Zhao}, {and} \bibinfo{person}{Tie-Yan Liu}.}
  \bibinfo{year}{2019}\natexlab{}.
\newblock \showarticletitle{Almost unsupervised text to speech and automatic
  speech recognition}.
\newblock \bibinfo{journal}{\emph{arXiv preprint arXiv:1905.06791}}
  (\bibinfo{year}{2019}).
\newblock


\bibitem[\protect\citeauthoryear{Ribeiro, Singh, and Guestrin}{Ribeiro
  et~al\mbox{.}}{2016}]%
        {ribeiro16-kdd}
\bibfield{author}{\bibinfo{person}{Marco~Tulio Ribeiro},
  \bibinfo{person}{Sameer Singh}, {and} \bibinfo{person}{Carlos Guestrin}.}
  \bibinfo{year}{2016}\natexlab{}.
\newblock \showarticletitle{“{Why} {Should} {I} {Trust} {You}?”:
  {Explaining} the {Predictions} of {Any} {Classifier}}. In
  \bibinfo{booktitle}{\emph{22nd {ACM} {SIGKDD} {Int}'l {Conf}. {Knowl}.
  {Disc}. {Data} {Mining}}} \emph{(\bibinfo{series}{{KDD} '16})}.
  \bibinfo{publisher}{ACM}, \bibinfo{address}{New York, NY, USA},
  \bibinfo{pages}{1135--1144}.
\newblock


\bibitem[\protect\citeauthoryear{{Rigaki} and {Garcia}}{{Rigaki} and
  {Garcia}}{2018}]%
        {rigaki2018bringing}
\bibfield{author}{\bibinfo{person}{M. {Rigaki}} {and} \bibinfo{person}{S.
  {Garcia}}.} \bibinfo{year}{2018}\natexlab{}.
\newblock \showarticletitle{Bringing a GAN to a Knife-Fight: Adapting Malware
  Communication to Avoid Detection}. In \bibinfo{booktitle}{\emph{2018 IEEE
  Security and Privacy Workshops (SPW)}}. \bibinfo{pages}{70--75}.
\newblock
\urldef\tempurl%
\url{https://doi.org/10.1109/SPW.2018.00019}
\showDOI{\tempurl}


\bibitem[\protect\citeauthoryear{Roller, Dinan, Goyal, Ju, Williamson, Liu, Xu,
  Ott, Shuster, Smith, Boureau, and Weston}{Roller et~al\mbox{.}}{2020}]%
        {roller20-arxiv}
\bibfield{author}{\bibinfo{person}{Stephen Roller}, \bibinfo{person}{Emily
  Dinan}, \bibinfo{person}{Naman Goyal}, \bibinfo{person}{Da Ju},
  \bibinfo{person}{Mary Williamson}, \bibinfo{person}{Yinhan Liu},
  \bibinfo{person}{Jing Xu}, \bibinfo{person}{Myle Ott}, \bibinfo{person}{Kurt
  Shuster}, \bibinfo{person}{Eric~M. Smith}, \bibinfo{person}{Y.-Lan Boureau},
  {and} \bibinfo{person}{Jason Weston}.} \bibinfo{year}{2020}\natexlab{}.
\newblock \showarticletitle{Recipes for building an open-domain chatbot}.
\newblock \bibinfo{journal}{\emph{arXiv:2004.13637 [cs]}}
  (\bibinfo{date}{April} \bibinfo{year}{2020}).
\newblock
\urldef\tempurl%
\url{http://arxiv.org/abs/2004.13637}
\showURL{%
\tempurl}
\newblock
\shownote{arXiv: 2004.13637.}


\bibitem[\protect\citeauthoryear{Salminen, Jung, and Jansen}{Salminen
  et~al\mbox{.}}{2019}]%
        {salminen2019future}
\bibfield{author}{\bibinfo{person}{Joni Salminen}, \bibinfo{person}{Soon-gyo
  Jung}, {and} \bibinfo{person}{Bernard~J Jansen}.}
  \bibinfo{year}{2019}\natexlab{}.
\newblock \showarticletitle{The Future of Data-driven Personas: A Marriage of
  Online Analytics Numbers and Human Attributes.}. In
  \bibinfo{booktitle}{\emph{ICEIS (1)}}. \bibinfo{pages}{608--615}.
\newblock


\bibitem[\protect\citeauthoryear{Salminen, Rao, Jung, Chowdhury, and
  Jansen}{Salminen et~al\mbox{.}}{2020}]%
        {salminen2020enriching}
\bibfield{author}{\bibinfo{person}{Joni Salminen},
  \bibinfo{person}{Rohan~Gurunandan Rao}, \bibinfo{person}{Soon-gyo Jung},
  \bibinfo{person}{Shammur~A Chowdhury}, {and} \bibinfo{person}{Bernard~J
  Jansen}.} \bibinfo{year}{2020}\natexlab{}.
\newblock \showarticletitle{Enriching Social Media Personas with Personality
  Traits: A Deep Learning Approach Using the Big Five Classes}. In
  \bibinfo{booktitle}{\emph{International Conference on Human-Computer
  Interaction}}. Springer, \bibinfo{pages}{101--120}.
\newblock


\bibitem[\protect\citeauthoryear{Samulowitz and Memisevic}{Samulowitz and
  Memisevic}{2007}]%
        {samulowitz2007learning}
\bibfield{author}{\bibinfo{person}{Horst Samulowitz} {and}
  \bibinfo{person}{Roland Memisevic}.} \bibinfo{year}{2007}\natexlab{}.
\newblock \showarticletitle{Learning to solve QBF}. In
  \bibinfo{booktitle}{\emph{AAAI}}, Vol.~\bibinfo{volume}{7}.
  \bibinfo{pages}{255--260}.
\newblock


\bibitem[\protect\citeauthoryear{Schreyer, Sattarov, Reimer, and
  Borth}{Schreyer et~al\mbox{.}}{2019}]%
        {schreyer2019adversarial}
\bibfield{author}{\bibinfo{person}{Marco Schreyer}, \bibinfo{person}{Timur
  Sattarov}, \bibinfo{person}{Bernd Reimer}, {and} \bibinfo{person}{Damian
  Borth}.} \bibinfo{year}{2019}\natexlab{}.
\newblock \bibinfo{title}{Adversarial Learning of Deepfakes in Accounting}.
\newblock
\newblock
\showeprint[arxiv]{1910.03810}~[cs.LG]


\bibitem[\protect\citeauthoryear{Schwartz and Kurniawati}{Schwartz and
  Kurniawati}{2019}]%
        {schwartz2019autonomous}
\bibfield{author}{\bibinfo{person}{Jonathon Schwartz} {and}
  \bibinfo{person}{Hanna Kurniawati}.} \bibinfo{year}{2019}\natexlab{}.
\newblock \showarticletitle{Autonomous penetration testing using reinforcement
  learning}.
\newblock \bibinfo{journal}{\emph{arXiv preprint arXiv:1905.05965}}
  (\bibinfo{year}{2019}).
\newblock


\bibitem[\protect\citeauthoryear{Seymour and Tully}{Seymour and Tully}{2016}]%
        {seymour2016weaponizing}
\bibfield{author}{\bibinfo{person}{John Seymour} {and} \bibinfo{person}{Philip
  Tully}.} \bibinfo{year}{2016}\natexlab{}.
\newblock \showarticletitle{Weaponizing data science for social engineering:
  Automated E2E spear phishing on Twitter}.
\newblock \bibinfo{journal}{\emph{Black Hat USA}}  \bibinfo{volume}{37}
  (\bibinfo{year}{2016}), \bibinfo{pages}{1--39}.
\newblock


\bibitem[\protect\citeauthoryear{Seymour and Tully}{Seymour and Tully}{2018}]%
        {seymour2018generative}
\bibfield{author}{\bibinfo{person}{John Seymour} {and} \bibinfo{person}{Philip
  Tully}.} \bibinfo{year}{2018}\natexlab{}.
\newblock \showarticletitle{Generative models for spear phishing posts on
  social media}.
\newblock \bibinfo{journal}{\emph{arXiv preprint arXiv:1802.05196}}
  (\bibinfo{year}{2018}).
\newblock


\bibitem[\protect\citeauthoryear{Shafahi, Huang, Najibi, Suciu, Studer,
  Dumitras, and Goldstein}{Shafahi et~al\mbox{.}}{2018}]%
        {Shafahi18-nips}
\bibfield{author}{\bibinfo{person}{Ali Shafahi}, \bibinfo{person}{W.~Ronny
  Huang}, \bibinfo{person}{Mahyar Najibi}, \bibinfo{person}{Octavian Suciu},
  \bibinfo{person}{Christoph Studer}, \bibinfo{person}{Tudor Dumitras}, {and}
  \bibinfo{person}{Tom Goldstein}.} \bibinfo{year}{2018}\natexlab{}.
\newblock \showarticletitle{Poison Frogs! Targeted Clean-Label Poisoning
  Attacks on Neural Networks}. In \bibinfo{booktitle}{\emph{Proceedings of the
  32nd International Conference on Neural Information Processing Systems}}
  (Montr\'{e}al, Canada) \emph{(\bibinfo{series}{NIPS'18})}.
  \bibinfo{publisher}{Curran Associates Inc.}, \bibinfo{address}{Red Hook, NY,
  USA}, \bibinfo{pages}{6106–6116}.
\newblock


\bibitem[\protect\citeauthoryear{Shan, Wenger, Zhang, Li, Zheng, and Zhao}{Shan
  et~al\mbox{.}}{2020}]%
        {shan2020fawkes}
\bibfield{author}{\bibinfo{person}{Shawn Shan}, \bibinfo{person}{Emily Wenger},
  \bibinfo{person}{Jiayun Zhang}, \bibinfo{person}{Huiying Li},
  \bibinfo{person}{Haitao Zheng}, {and} \bibinfo{person}{Ben~Y Zhao}.}
  \bibinfo{year}{2020}\natexlab{}.
\newblock \showarticletitle{Fawkes: Protecting Privacy against Unauthorized
  Deep Learning Models}. In \bibinfo{booktitle}{\emph{29th $\{$USENIX$\}$
  Security Symposium ($\{$USENIX$\}$ Security 20)}}.
  \bibinfo{pages}{1589--1604}.
\newblock


\bibitem[\protect\citeauthoryear{shaoanlu}{shaoanlu}{2020}]%
        {shaoanlu87:online}
\bibfield{author}{\bibinfo{person}{shaoanlu}.} \bibinfo{year}{2020}\natexlab{}.
\newblock \bibinfo{title}{shaoanlu/faceswap-GAN: A denoising autoencoder +
  adversarial losses and attention mechanisms for face swapping.}
\newblock \bibinfo{howpublished}{{https://github.com/shaoanlu/faceswap-GAN}}.
\newblock
\newblock
\shownote{(Accessed on 10/19/2020).}


\bibitem[\protect\citeauthoryear{Sharif, Bhagavatula, Bauer, and Reiter}{Sharif
  et~al\mbox{.}}{2016}]%
        {sharif2016accessorize}
\bibfield{author}{\bibinfo{person}{Mahmood Sharif}, \bibinfo{person}{Sruti
  Bhagavatula}, \bibinfo{person}{Lujo Bauer}, {and} \bibinfo{person}{Michael~K
  Reiter}.} \bibinfo{year}{2016}\natexlab{}.
\newblock \showarticletitle{Accessorize to a crime: Real and stealthy attacks
  on state-of-the-art face recognition}. In
  \bibinfo{booktitle}{\emph{Proceedings of the 2016 ACM SIGSAC Conference on
  Computer and Communications Security}}. ACM, \bibinfo{pages}{1528--1540}.
\newblock


\bibitem[\protect\citeauthoryear{Sharon, Berend, Liu, Shabtai, and
  Elovici}{Sharon et~al\mbox{.}}{2021}]%
        {sharon2021tantra}
\bibfield{author}{\bibinfo{person}{Yam Sharon}, \bibinfo{person}{David Berend},
  \bibinfo{person}{Yang Liu}, \bibinfo{person}{Asaf Shabtai}, {and}
  \bibinfo{person}{Yuval Elovici}.} \bibinfo{year}{2021}\natexlab{}.
\newblock \showarticletitle{TANTRA: Timing-Based Adversarial Network Traffic
  Reshaping Attack}.
\newblock \bibinfo{journal}{\emph{arXiv preprint arXiv:2103.06297}}
  (\bibinfo{year}{2021}).
\newblock


\bibitem[\protect\citeauthoryear{She, Krishna, Yan, Jana, and Ray}{She
  et~al\mbox{.}}{2020}]%
        {she2020mtfuzz}
\bibfield{author}{\bibinfo{person}{Dongdong She}, \bibinfo{person}{Rahul
  Krishna}, \bibinfo{person}{Lu Yan}, \bibinfo{person}{Suman Jana}, {and}
  \bibinfo{person}{Baishakhi Ray}.} \bibinfo{year}{2020}\natexlab{}.
\newblock \showarticletitle{MTFuzz: Fuzzing with a Multi-Task Neural Network}.
\newblock \bibinfo{journal}{\emph{arXiv preprint arXiv:2005.12392}}
  (\bibinfo{year}{2020}).
\newblock


\bibitem[\protect\citeauthoryear{She, Pei, Epstein, Yang, Ray, and Jana}{She
  et~al\mbox{.}}{2019}]%
        {she2019neuzz}
\bibfield{author}{\bibinfo{person}{Dongdong She}, \bibinfo{person}{Kexin Pei},
  \bibinfo{person}{Dave Epstein}, \bibinfo{person}{Junfeng Yang},
  \bibinfo{person}{Baishakhi Ray}, {and} \bibinfo{person}{Suman Jana}.}
  \bibinfo{year}{2019}\natexlab{}.
\newblock \showarticletitle{NEUZZ: Efficient fuzzing with neural program
  smoothing}. In \bibinfo{booktitle}{\emph{2019 IEEE Symposium on Security and
  Privacy (SP)}}. IEEE, \bibinfo{pages}{803--817}.
\newblock


\bibitem[\protect\citeauthoryear{Shin, Song, and Moazzezi}{Shin
  et~al\mbox{.}}{2015}]%
        {shin2015recognizing}
\bibfield{author}{\bibinfo{person}{Eui Chul~Richard Shin},
  \bibinfo{person}{Dawn Song}, {and} \bibinfo{person}{Reza Moazzezi}.}
  \bibinfo{year}{2015}\natexlab{}.
\newblock \showarticletitle{Recognizing functions in binaries with neural
  networks}. In \bibinfo{booktitle}{\emph{24th $\{$USENIX$\}$ Security
  Symposium ($\{$USENIX$\}$ Security 15)}}. \bibinfo{pages}{611--626}.
\newblock


\bibitem[\protect\citeauthoryear{Shokri, Stronati, Song, and Shmatikov}{Shokri
  et~al\mbox{.}}{2017}]%
        {shokri17-sp}
\bibfield{author}{\bibinfo{person}{R. Shokri}, \bibinfo{person}{M. Stronati},
  \bibinfo{person}{C. Song}, {and} \bibinfo{person}{V. Shmatikov}.}
  \bibinfo{year}{2017}\natexlab{}.
\newblock \showarticletitle{Membership {Inference} {Attacks} {Against}
  {Machine} {Learning} {Models}}. In \bibinfo{booktitle}{\emph{2017 {IEEE}
  {Symposium} on {Security} and {Privacy} ({SP})}}. \bibinfo{pages}{3--18}.
\newblock


\bibitem[\protect\citeauthoryear{Shumailov, Simon, Yan, and Anderson}{Shumailov
  et~al\mbox{.}}{2019}]%
        {shumailov2019hearing}
\bibfield{author}{\bibinfo{person}{Ilia Shumailov}, \bibinfo{person}{Laurent
  Simon}, \bibinfo{person}{Jeff Yan}, {and} \bibinfo{person}{Ross Anderson}.}
  \bibinfo{year}{2019}\natexlab{}.
\newblock \showarticletitle{Hearing your touch: A new acoustic side channel on
  smartphones}.
\newblock \bibinfo{journal}{\emph{arXiv preprint arXiv:1903.11137}}
  (\bibinfo{year}{2019}).
\newblock


\bibitem[\protect\citeauthoryear{Shumailov, Zhao, Bates, Papernot, Mullins, and
  Anderson}{Shumailov et~al\mbox{.}}{2020}]%
        {shumailov2020sponge}
\bibfield{author}{\bibinfo{person}{Ilia Shumailov}, \bibinfo{person}{Yiren
  Zhao}, \bibinfo{person}{Daniel Bates}, \bibinfo{person}{Nicolas Papernot},
  \bibinfo{person}{Robert Mullins}, {and} \bibinfo{person}{Ross Anderson}.}
  \bibinfo{year}{2020}\natexlab{}.
\newblock \showarticletitle{Sponge Examples: Energy-Latency Attacks on Neural
  Networks}.
\newblock \bibinfo{journal}{\emph{arXiv preprint arXiv:2006.03463}}
  (\bibinfo{year}{2020}).
\newblock


\bibitem[\protect\citeauthoryear{Siarohin, Lathuilière, Tulyakov, Ricci, and
  Sebe}{Siarohin et~al\mbox{.}}{2019}]%
        {Siarohin_2019_NeurIPS}
\bibfield{author}{\bibinfo{person}{Aliaksandr Siarohin},
  \bibinfo{person}{Stéphane Lathuilière}, \bibinfo{person}{Sergey Tulyakov},
  \bibinfo{person}{Elisa Ricci}, {and} \bibinfo{person}{Nicu Sebe}.}
  \bibinfo{year}{2019}\natexlab{}.
\newblock \showarticletitle{First Order Motion Model for Image Animation}. In
  \bibinfo{booktitle}{\emph{Conference on Neural Information Processing Systems
  (NeurIPS)}}.
\newblock


\bibitem[\protect\citeauthoryear{Sidi, Nadler, and Shabtai}{Sidi
  et~al\mbox{.}}{2020}]%
        {sidi2020maskdga}
\bibfield{author}{\bibinfo{person}{Lior Sidi}, \bibinfo{person}{Asaf Nadler},
  {and} \bibinfo{person}{Asaf Shabtai}.} \bibinfo{year}{2020}\natexlab{}.
\newblock \showarticletitle{MaskDGA: An Evasion Attack Against DGA Classifiers
  and Adversarial Defenses}.
\newblock \bibinfo{journal}{\emph{IEEE Access}}  \bibinfo{volume}{8}
  (\bibinfo{year}{2020}), \bibinfo{pages}{161580--161592}.
\newblock


\bibitem[\protect\citeauthoryear{Singh and Thakur}{Singh and Thakur}{2020}]%
        {singh2020survey}
\bibfield{author}{\bibinfo{person}{Siddhant Singh} {and}
  \bibinfo{person}{Hardeo~K Thakur}.} \bibinfo{year}{2020}\natexlab{}.
\newblock \showarticletitle{Survey of Various AI Chatbots Based on Technology
  Used}. In \bibinfo{booktitle}{\emph{2020 8th International Conference on
  Reliability, Infocom Technologies and Optimization (Trends and Future
  Directions)(ICRITO)}}. IEEE, \bibinfo{pages}{1074--1079}.
\newblock


\bibitem[\protect\citeauthoryear{Song, Wagner, and Tian}{Song
  et~al\mbox{.}}{2001}]%
        {song2001timing}
\bibfield{author}{\bibinfo{person}{Dawn~Xiaodong Song},
  \bibinfo{person}{David~A Wagner}, {and} \bibinfo{person}{Xuqing Tian}.}
  \bibinfo{year}{2001}\natexlab{}.
\newblock \showarticletitle{Timing analysis of keystrokes and timing attacks on
  ssh.}. In \bibinfo{booktitle}{\emph{USENIX Security Symposium}},
  Vol.~\bibinfo{volume}{2001}.
\newblock


\bibitem[\protect\citeauthoryear{Spiliotopoulos, Margaris, and
  Vassilakis}{Spiliotopoulos et~al\mbox{.}}{2020}]%
        {spiliotopoulos2020data}
\bibfield{author}{\bibinfo{person}{Dimitris Spiliotopoulos},
  \bibinfo{person}{Dionisis Margaris}, {and} \bibinfo{person}{Costas
  Vassilakis}.} \bibinfo{year}{2020}\natexlab{}.
\newblock \showarticletitle{Data-Assisted Persona Construction Using Social
  Media Data}.
\newblock \bibinfo{journal}{\emph{Big Data and Cognitive Computing}}
  \bibinfo{volume}{4}, \bibinfo{number}{3} (\bibinfo{year}{2020}),
  \bibinfo{pages}{21}.
\newblock


\bibitem[\protect\citeauthoryear{Stupp}{Stupp}{[n.d.]}]%
        {fraudsters_mimic:online}
\bibfield{author}{\bibinfo{person}{Catherine Stupp}.}
  \bibinfo{year}{[n.d.]}\natexlab{}.
\newblock \bibinfo{title}{Fraudsters Used AI to Mimic CEO’s Voice in Unusual
  Cybercrime Case}.
\newblock
  \bibinfo{howpublished}{https://www.wsj.com/articles/fraudsters-use-ai-to-mimic-ceos-voice-in-unusual-cybercrime-case-11567157402}.
\newblock
\newblock
\shownote{(Accessed on 10/14/2020).}


\bibitem[\protect\citeauthoryear{{Suciu}, {Coull}, and {Johns}}{{Suciu}
  et~al\mbox{.}}{2019}]%
        {suciu19-spw}
\bibfield{author}{\bibinfo{person}{O. {Suciu}}, \bibinfo{person}{S.~E.
  {Coull}}, {and} \bibinfo{person}{J. {Johns}}.}
  \bibinfo{year}{2019}\natexlab{}.
\newblock \showarticletitle{Exploring Adversarial Examples in Malware
  Detection}. In \bibinfo{booktitle}{\emph{2019 IEEE Security and Privacy
  Workshops (SPW)}}. \bibinfo{pages}{8--14}.
\newblock
\urldef\tempurl%
\url{https://doi.org/10.1109/SPW.2019.00015}
\showDOI{\tempurl}


\bibitem[\protect\citeauthoryear{Sun, Jin, Chen, Zhang, Zhang, and Zhang}{Sun
  et~al\mbox{.}}{2016}]%
        {sun2016visible}
\bibfield{author}{\bibinfo{person}{Jingchao Sun}, \bibinfo{person}{Xiaocong
  Jin}, \bibinfo{person}{Yimin Chen}, \bibinfo{person}{Jinxue Zhang},
  \bibinfo{person}{Yanchao Zhang}, {and} \bibinfo{person}{Rui Zhang}.}
  \bibinfo{year}{2016}\natexlab{}.
\newblock \showarticletitle{VISIBLE: Video-Assisted Keystroke Inference from
  Tablet Backside Motion.}. In \bibinfo{booktitle}{\emph{NDSS}}.
\newblock


\bibitem[\protect\citeauthoryear{Sun, Tewari, Xu, Fritz, Theobalt, and
  Schiele}{Sun et~al\mbox{.}}{2018}]%
        {sun2018hybrid}
\bibfield{author}{\bibinfo{person}{Qianru Sun}, \bibinfo{person}{Ayush Tewari},
  \bibinfo{person}{Weipeng Xu}, \bibinfo{person}{Mario Fritz},
  \bibinfo{person}{Christian Theobalt}, {and} \bibinfo{person}{Bernt Schiele}.}
  \bibinfo{year}{2018}\natexlab{}.
\newblock \showarticletitle{A hybrid model for identity obfuscation by face
  replacement}. In \bibinfo{booktitle}{\emph{Proceedings of the European
  Conference on Computer Vision (ECCV)}}. \bibinfo{pages}{553--569}.
\newblock


\bibitem[\protect\citeauthoryear{Sutro}{Sutro}{2020}]%
        {sutro20}
\bibfield{author}{\bibinfo{person}{Alejo~Grigera Sutro}.}
  \bibinfo{year}{2020}\natexlab{}.
\newblock \bibinfo{title}{Machine-{Learning} {Based} {Evaluation} of {Access}
  {Control} {Lists} to {Identify} {Anomalies}}.
\newblock
\newblock
\urldef\tempurl%
\url{https://www.tdcommons.org/dpubs\_series/2870}
\showURL{%
\tempurl}


\bibitem[\protect\citeauthoryear{Szegedy, Zaremba, Sutskever, Bruna, Erhan,
  Goodfellow, and Fergus}{Szegedy et~al\mbox{.}}{2014}]%
        {szegedy14-iclr}
\bibfield{author}{\bibinfo{person}{Christian Szegedy},
  \bibinfo{person}{Wojciech Zaremba}, \bibinfo{person}{Ilya Sutskever},
  \bibinfo{person}{Joan Bruna}, \bibinfo{person}{Dumitru Erhan},
  \bibinfo{person}{Ian Goodfellow}, {and} \bibinfo{person}{Rob Fergus}.}
  \bibinfo{year}{2014}\natexlab{}.
\newblock \showarticletitle{Intriguing properties of neural networks}. In
  \bibinfo{booktitle}{\emph{International Conference on Learning
  Representations}}.
\newblock
\urldef\tempurl%
\url{http://arxiv.org/abs/1312.6199}
\showURL{%
\tempurl}


\bibitem[\protect\citeauthoryear{Tariq}{Tariq}{2018}]%
        {Tariq2018IMPACTOC}
\bibfield{author}{\bibinfo{person}{N. Tariq}.} \bibinfo{year}{2018}\natexlab{}.
\newblock \showarticletitle{IMPACT OF CYBERATTACKS ON FINANCIAL INSTITUTIONS}.
\newblock \bibinfo{journal}{\emph{The Journal of Internet Banking and
  Commerce}}  \bibinfo{volume}{23} (\bibinfo{year}{2018}),
  \bibinfo{pages}{1--11}.
\newblock


\bibitem[\protect\citeauthoryear{Truong, Zelinka, and Senkerik}{Truong
  et~al\mbox{.}}{2019}]%
        {truong2019neural}
\bibfield{author}{\bibinfo{person}{Thanh~Cong Truong}, \bibinfo{person}{Ivan
  Zelinka}, {and} \bibinfo{person}{Roman Senkerik}.}
  \bibinfo{year}{2019}\natexlab{}.
\newblock \showarticletitle{Neural swarm virus}.
\newblock In \bibinfo{booktitle}{\emph{Swarm, Evolutionary, and Memetic
  Computing and Fuzzy and Neural Computing}}. \bibinfo{publisher}{Springer},
  \bibinfo{pages}{122--134}.
\newblock


\bibitem[\protect\citeauthoryear{Ucci, Aniello, and Baldoni}{Ucci
  et~al\mbox{.}}{2019}]%
        {ucci2019survey}
\bibfield{author}{\bibinfo{person}{Daniele Ucci}, \bibinfo{person}{Leonardo
  Aniello}, {and} \bibinfo{person}{Roberto Baldoni}.}
  \bibinfo{year}{2019}\natexlab{}.
\newblock \showarticletitle{Survey of machine learning techniques for malware
  analysis}.
\newblock \bibinfo{journal}{\emph{Computers \& Security}}  \bibinfo{volume}{81}
  (\bibinfo{year}{2019}), \bibinfo{pages}{123--147}.
\newblock


\bibitem[\protect\citeauthoryear{Wang and Gong}{Wang and Gong}{2018}]%
        {wang18-sp}
\bibfield{author}{\bibinfo{person}{Binghui Wang} {and}
  \bibinfo{person}{Neil~Zhenqiang Gong}.} \bibinfo{year}{2018}\natexlab{}.
\newblock \showarticletitle{Stealing hyperparameters in machine learning}. In
  \bibinfo{booktitle}{\emph{2018 IEEE Symposium on Security and Privacy (SP)}}.
  IEEE, \bibinfo{pages}{36--52}.
\newblock


\bibitem[\protect\citeauthoryear{Wang, Neupane, Qian, Abu-Ghazaleh,
  Krishnamurthy, Colbert, and Yu}{Wang et~al\mbox{.}}{2019b}]%
        {wang2019unveiling}
\bibfield{author}{\bibinfo{person}{Daimeng Wang}, \bibinfo{person}{Ajaya
  Neupane}, \bibinfo{person}{Zhiyun Qian}, \bibinfo{person}{Nael~B
  Abu-Ghazaleh}, \bibinfo{person}{Srikanth~V Krishnamurthy},
  \bibinfo{person}{Edward~JM Colbert}, {and} \bibinfo{person}{Paul Yu}.}
  \bibinfo{year}{2019}\natexlab{b}.
\newblock \showarticletitle{Unveiling your keystrokes: A Cache-based
  Side-channel Attack on Graphics Libraries.}. In
  \bibinfo{booktitle}{\emph{NDSS}}.
\newblock


\bibitem[\protect\citeauthoryear{Wang, Zhao, Yang, Wu, Hu, and Xing}{Wang
  et~al\mbox{.}}{2019d}]%
        {wang2019deeptrust}
\bibfield{author}{\bibinfo{person}{Qi Wang}, \bibinfo{person}{Weiliang Zhao},
  \bibinfo{person}{Jian Yang}, \bibinfo{person}{Jia Wu},
  \bibinfo{person}{Wenbin Hu}, {and} \bibinfo{person}{Qianli Xing}.}
  \bibinfo{year}{2019}\natexlab{d}.
\newblock \showarticletitle{DeepTrust: A Deep User Model of Homophily Effect
  for Trust Prediction}. In \bibinfo{booktitle}{\emph{2019 IEEE International
  Conference on Data Mining (ICDM)}}. IEEE, \bibinfo{pages}{618--627}.
\newblock


\bibitem[\protect\citeauthoryear{{Wang}, {Nepal}, {Rudolph}, {Grobler}, {Chen},
  and {Chen}}{{Wang} et~al\mbox{.}}{2020}]%
        {wang20}
\bibfield{author}{\bibinfo{person}{S. {Wang}}, \bibinfo{person}{S. {Nepal}},
  \bibinfo{person}{C. {Rudolph}}, \bibinfo{person}{M. {Grobler}},
  \bibinfo{person}{S. {Chen}}, {and} \bibinfo{person}{T. {Chen}}.}
  \bibinfo{year}{2020}\natexlab{}.
\newblock \showarticletitle{Backdoor Attacks against Transfer Learning with
  Pre-trained Deep Learning Models}.
\newblock \bibinfo{journal}{\emph{IEEE Transactions on Services Computing}}
  (\bibinfo{year}{2020}), \bibinfo{pages}{1--1}.
\newblock
\urldef\tempurl%
\url{https://doi.org/10.1109/TSC.2020.3000900}
\showDOI{\tempurl}


\bibitem[\protect\citeauthoryear{Wang, Yamagishi, Todisco, Delgado, Nautsch,
  Evans, Sahidullah, Vestman, Kinnunen, Lee, et~al\mbox{.}}{Wang
  et~al\mbox{.}}{2019c}]%
        {wang2019asvspoof}
\bibfield{author}{\bibinfo{person}{Xin Wang}, \bibinfo{person}{Junichi
  Yamagishi}, \bibinfo{person}{Massimiliano Todisco}, \bibinfo{person}{Hector
  Delgado}, \bibinfo{person}{Andreas Nautsch}, \bibinfo{person}{Nicholas
  Evans}, \bibinfo{person}{Md Sahidullah}, \bibinfo{person}{Ville Vestman},
  \bibinfo{person}{Tomi Kinnunen}, \bibinfo{person}{Kong~Aik Lee},
  {et~al\mbox{.}}} \bibinfo{year}{2019}\natexlab{c}.
\newblock \showarticletitle{The ASVspoof 2019 database}.
\newblock \bibinfo{journal}{\emph{arXiv preprint arXiv:1911.01601}}
  (\bibinfo{year}{2019}).
\newblock


\bibitem[\protect\citeauthoryear{Wang, Bao, Ding, and Zhu}{Wang
  et~al\mbox{.}}{2017}]%
        {wang2017face}
\bibfield{author}{\bibinfo{person}{Ya Wang}, \bibinfo{person}{Tianlong Bao},
  \bibinfo{person}{Chunhui Ding}, {and} \bibinfo{person}{Ming Zhu}.}
  \bibinfo{year}{2017}\natexlab{}.
\newblock \showarticletitle{Face recognition in real-world surveillance videos
  with deep learning method}. In \bibinfo{booktitle}{\emph{Image, Vision and
  Computing (ICIVC), 2017 2nd International Conference on}}. IEEE,
  \bibinfo{pages}{239--243}.
\newblock


\bibitem[\protect\citeauthoryear{Wang, Cai, Gu, and Shao}{Wang
  et~al\mbox{.}}{2019a}]%
        {wang2019your}
\bibfield{author}{\bibinfo{person}{Yao Wang}, \bibinfo{person}{Wandong Cai},
  \bibinfo{person}{Tao Gu}, {and} \bibinfo{person}{Wei Shao}.}
  \bibinfo{year}{2019}\natexlab{a}.
\newblock \showarticletitle{Your eyes reveal your secrets: an eye movement
  based password inference on smartphone}.
\newblock \bibinfo{journal}{\emph{IEEE transactions on mobile computing}}
  (\bibinfo{year}{2019}).
\newblock


\bibitem[\protect\citeauthoryear{Wang, Cai, Gu, Shao, Khalil, and Xu}{Wang
  et~al\mbox{.}}{2018}]%
        {wang2018gazerevealer}
\bibfield{author}{\bibinfo{person}{Yao Wang}, \bibinfo{person}{Wandong Cai},
  \bibinfo{person}{Tao Gu}, \bibinfo{person}{Wei Shao},
  \bibinfo{person}{Ibrahim Khalil}, {and} \bibinfo{person}{Xianghua Xu}.}
  \bibinfo{year}{2018}\natexlab{}.
\newblock \showarticletitle{GazeRevealer: Inferring password using smartphone
  front camera}. In \bibinfo{booktitle}{\emph{Proceedings of the 15th EAI
  International Conference on Mobile and Ubiquitous Systems: Computing,
  Networking and Services}}. \bibinfo{pages}{254--263}.
\newblock


\bibitem[\protect\citeauthoryear{Wang, Jia, Liu, Huang, and Liu}{Wang
  et~al\mbox{.}}{2020}]%
        {wang2020systematic}
\bibfield{author}{\bibinfo{person}{Yan Wang}, \bibinfo{person}{Peng Jia},
  \bibinfo{person}{Luping Liu}, \bibinfo{person}{Cheng Huang}, {and}
  \bibinfo{person}{Zhonglin Liu}.} \bibinfo{year}{2020}\natexlab{}.
\newblock \showarticletitle{A systematic review of fuzzing based on machine
  learning techniques}.
\newblock \bibinfo{journal}{\emph{PloS one}} \bibinfo{volume}{15},
  \bibinfo{number}{8} (\bibinfo{year}{2020}), \bibinfo{pages}{e0237749}.
\newblock


\bibitem[\protect\citeauthoryear{Wang, Yao, Kwok, and Ni}{Wang
  et~al\mbox{.}}{2020}]%
        {wang20-acmcs}
\bibfield{author}{\bibinfo{person}{Yaqing Wang}, \bibinfo{person}{Quanming
  Yao}, \bibinfo{person}{James~T. Kwok}, {and} \bibinfo{person}{Lionel~M. Ni}.}
  \bibinfo{year}{2020}\natexlab{}.
\newblock \showarticletitle{Generalizing from a Few Examples: A Survey on
  Few-Shot Learning}.
\newblock \bibinfo{journal}{\emph{ACM Comput. Surv.}} \bibinfo{volume}{53},
  \bibinfo{number}{3}, Article \bibinfo{articleno}{63} (\bibinfo{date}{June}
  \bibinfo{year}{2020}), \bibinfo{numpages}{34}~pages.
\newblock
\showISSN{0360-0300}
\urldef\tempurl%
\url{https://doi.org/10.1145/3386252}
\showDOI{\tempurl}


\bibitem[\protect\citeauthoryear{Wang, Zhang, Yao, Qu, and Guo}{Wang
  et~al\mbox{.}}{2011}]%
        {wang2011inferring}
\bibfield{author}{\bibinfo{person}{Yipeng Wang}, \bibinfo{person}{Zhibin
  Zhang}, \bibinfo{person}{Danfeng~Daphne Yao}, \bibinfo{person}{Buyun Qu},
  {and} \bibinfo{person}{Li Guo}.} \bibinfo{year}{2011}\natexlab{}.
\newblock \showarticletitle{Inferring protocol state machine from network
  traces: a probabilistic approach}. In \bibinfo{booktitle}{\emph{International
  Conference on Applied Cryptography and Network Security}}. Springer,
  \bibinfo{pages}{1--18}.
\newblock


\bibitem[\protect\citeauthoryear{Weissbart, Picek, and Batina}{Weissbart
  et~al\mbox{.}}{2019}]%
        {10.1007/978-3-030-35869-3_8}
\bibfield{author}{\bibinfo{person}{L{\'e}o Weissbart}, \bibinfo{person}{Stjepan
  Picek}, {and} \bibinfo{person}{Lejla Batina}.}
  \bibinfo{year}{2019}\natexlab{}.
\newblock \showarticletitle{One Trace Is All It Takes: Machine Learning-Based
  Side-Channel Attack on EdDSA}. In \bibinfo{booktitle}{\emph{Security,
  Privacy, and Applied Cryptography Engineering}},
  \bibfield{editor}{\bibinfo{person}{Shivam Bhasin}, \bibinfo{person}{Avi
  Mendelson}, {and} \bibinfo{person}{Mridul Nandi}} (Eds.).
  \bibinfo{publisher}{Springer International Publishing},
  \bibinfo{address}{Cham}, \bibinfo{pages}{86--105}.
\newblock
\showISBNx{978-3-030-35869-3}


\bibitem[\protect\citeauthoryear{White, Matthews, Snow, and Monrose}{White
  et~al\mbox{.}}{2011}]%
        {white11-sp}
\bibfield{author}{\bibinfo{person}{Andrew White}, \bibinfo{person}{Austin
  Matthews}, \bibinfo{person}{Kevin Snow}, {and} \bibinfo{person}{Fabian
  Monrose}.} \bibinfo{year}{2011}\natexlab{}.
\newblock \showarticletitle{Phonotactic Reconstruction of Encrypted VoIP
  Conversations: Hookt on Fon-iks}.
\newblock \bibinfo{journal}{\emph{Proceedings - IEEE Symposium on Security and
  Privacy}}, \bibinfo{pages}{3 -- 18}.
\newblock
\urldef\tempurl%
\url{https://doi.org/10.1109/SP.2011.34}
\showDOI{\tempurl}


\bibitem[\protect\citeauthoryear{{Woh} and {Lee}}{{Woh} and {Lee}}{2018}]%
        {woh18-scis}
\bibfield{author}{\bibinfo{person}{S. {Woh}} {and} \bibinfo{person}{J. {Lee}}.}
  \bibinfo{year}{2018}\natexlab{}.
\newblock \showarticletitle{Game State Prediction with Ensemble of Machine
  Learning Techniques}. In \bibinfo{booktitle}{\emph{2018 Joint 10th
  International Conference on Soft Computing and Intelligent Systems (SCIS) and
  19th International Symposium on Advanced Intelligent Systems (ISIS)}}.
  \bibinfo{pages}{89--92}.
\newblock
\urldef\tempurl%
\url{https://doi.org/10.1109/SCIS-ISIS.2018.00025}
\showDOI{\tempurl}


\bibitem[\protect\citeauthoryear{Workman}{Workman}{2008}]%
        {workman2008wisecrackers}
\bibfield{author}{\bibinfo{person}{Michael Workman}.}
  \bibinfo{year}{2008}\natexlab{}.
\newblock \showarticletitle{Wisecrackers: A theory-grounded investigation of
  phishing and pretext social engineering threats to information security}.
\newblock \bibinfo{journal}{\emph{Journal of the American Society for
  Information Science and Technology}} \bibinfo{volume}{59},
  \bibinfo{number}{4} (\bibinfo{year}{2008}), \bibinfo{pages}{662--674}.
\newblock


\bibitem[\protect\citeauthoryear{Wu, Gong, Tong, and Fan}{Wu
  et~al\mbox{.}}{2021}]%
        {wu21}
\bibfield{author}{\bibinfo{person}{Runze Wu}, \bibinfo{person}{Jinxin Gong},
  \bibinfo{person}{Weiyue Tong}, {and} \bibinfo{person}{Bing Fan}.}
  \bibinfo{year}{2021}\natexlab{}.
\newblock \showarticletitle{Network Attack Path Selection and Evaluation Based
  on Q-Learning}.
\newblock \bibinfo{journal}{\emph{Applied Sciences}} \bibinfo{volume}{11},
  \bibinfo{number}{1} (\bibinfo{year}{2021}).
\newblock
\showISSN{2076-3417}
\urldef\tempurl%
\url{https://doi.org/10.3390/app11010285}
\showDOI{\tempurl}


\bibitem[\protect\citeauthoryear{Xu, Liu, Feng, Yin, Song, and Song}{Xu
  et~al\mbox{.}}{2017}]%
        {Xu_2017}
\bibfield{author}{\bibinfo{person}{Xiaojun Xu}, \bibinfo{person}{Chang Liu},
  \bibinfo{person}{Qian Feng}, \bibinfo{person}{Heng Yin}, \bibinfo{person}{Le
  Song}, {and} \bibinfo{person}{Dawn Song}.} \bibinfo{year}{2017}\natexlab{}.
\newblock \showarticletitle{Neural Network-based Graph Embedding for
  Cross-Platform Binary Code Similarity Detection}.
\newblock \bibinfo{journal}{\emph{Proceedings of the 2017 ACM SIGSAC Conference
  on Computer and Communications Security}} (\bibinfo{date}{Oct}
  \bibinfo{year}{2017}).
\newblock
\showISBNx{9781450349468}
\urldef\tempurl%
\url{https://doi.org/10.1145/3133956.3134018}
\showDOI{\tempurl}


\bibitem[\protect\citeauthoryear{{Yager}}{{Yager}}{1984}]%
        {Yager84}
\bibfield{author}{\bibinfo{person}{R.~R. {Yager}}.}
  \bibinfo{year}{1984}\natexlab{}.
\newblock \showarticletitle{Approximate reasoning as a basis for rule-based
  expert systems}.
\newblock \bibinfo{journal}{\emph{IEEE Transactions on Systems, Man, and
  Cybernetics}} \bibinfo{volume}{SMC-14}, \bibinfo{number}{4}
  (\bibinfo{year}{1984}), \bibinfo{pages}{636--643}.
\newblock
\urldef\tempurl%
\url{https://doi.org/10.1109/TSMC.1984.6313337}
\showDOI{\tempurl}


\bibitem[\protect\citeauthoryear{Yang, Hu, Dyer, Xing, and
  Berg-Kirkpatrick}{Yang et~al\mbox{.}}{2018}]%
        {yang2018unsupervised}
\bibfield{author}{\bibinfo{person}{Zichao Yang}, \bibinfo{person}{Zhiting Hu},
  \bibinfo{person}{Chris Dyer}, \bibinfo{person}{Eric~P Xing}, {and}
  \bibinfo{person}{Taylor Berg-Kirkpatrick}.} \bibinfo{year}{2018}\natexlab{}.
\newblock \showarticletitle{Unsupervised text style transfer using language
  models as discriminators}. In \bibinfo{booktitle}{\emph{Advances in Neural
  Information Processing Systems}}. \bibinfo{pages}{7287--7298}.
\newblock


\bibitem[\protect\citeauthoryear{Yao, Li, Zheng, and Zhao}{Yao
  et~al\mbox{.}}{2019}]%
        {yao19-ccs}
\bibfield{author}{\bibinfo{person}{Yuanshun Yao}, \bibinfo{person}{Huiying Li},
  \bibinfo{person}{Haitao Zheng}, {and} \bibinfo{person}{Ben~Y. Zhao}.}
  \bibinfo{year}{2019}\natexlab{}.
\newblock \showarticletitle{Latent Backdoor Attacks on Deep Neural Networks}.
  In \bibinfo{booktitle}{\emph{Proceedings of the 2019 ACM SIGSAC Conference on
  Computer and Communications Security}} (London, United Kingdom)
  \emph{(\bibinfo{series}{CCS '19})}. \bibinfo{publisher}{Association for
  Computing Machinery}, \bibinfo{address}{New York, NY, USA},
  \bibinfo{pages}{2041–2055}.
\newblock
\showISBNx{9781450367479}
\urldef\tempurl%
\url{https://doi.org/10.1145/3319535.3354209}
\showDOI{\tempurl}


\bibitem[\protect\citeauthoryear{Ye, Zhou, Venkat, Marucs, Tatbul, Tithi,
  Petersen, Mattson, Kraska, Dubey, et~al\mbox{.}}{Ye et~al\mbox{.}}{2020}]%
        {ye2020misim}
\bibfield{author}{\bibinfo{person}{Fangke Ye}, \bibinfo{person}{Shengtian
  Zhou}, \bibinfo{person}{Anand Venkat}, \bibinfo{person}{Ryan Marucs},
  \bibinfo{person}{Nesime Tatbul}, \bibinfo{person}{Jesmin~Jahan Tithi},
  \bibinfo{person}{Paul Petersen}, \bibinfo{person}{Timothy Mattson},
  \bibinfo{person}{Tim Kraska}, \bibinfo{person}{Pradeep Dubey},
  {et~al\mbox{.}}} \bibinfo{year}{2020}\natexlab{}.
\newblock \showarticletitle{MISIM: An End-to-End Neural Code Similarity
  System}.
\newblock \bibinfo{journal}{\emph{arXiv preprint arXiv:2006.05265}}
  (\bibinfo{year}{2020}).
\newblock


\bibitem[\protect\citeauthoryear{{Yousefi}, {Mtetwa}, {Zhang}, and
  {Tianfield}}{{Yousefi} et~al\mbox{.}}{2018}]%
        {Yousefi18-ICTSP}
\bibfield{author}{\bibinfo{person}{M. {Yousefi}}, \bibinfo{person}{N.
  {Mtetwa}}, \bibinfo{person}{Y. {Zhang}}, {and} \bibinfo{person}{H.
  {Tianfield}}.} \bibinfo{year}{2018}\natexlab{}.
\newblock \showarticletitle{A Reinforcement Learning Approach for Attack Graph
  Analysis}. In \bibinfo{booktitle}{\emph{2018 17th IEEE International
  Conference On Trust, Security And Privacy In Computing And Communications/
  12th IEEE International Conference On Big Data Science And Engineering
  (TrustCom/BigDataSE)}}. \bibinfo{pages}{212--217}.
\newblock
\urldef\tempurl%
\url{https://doi.org/10.1109/TrustCom/BigDataSE.2018.00041}
\showDOI{\tempurl}


\bibitem[\protect\citeauthoryear{Yu, Lu, Chen, Zhu, and Kong}{Yu
  et~al\mbox{.}}{2019}]%
        {yu2019indirect}
\bibfield{author}{\bibinfo{person}{Jiadi Yu}, \bibinfo{person}{Li Lu},
  \bibinfo{person}{Yingying Chen}, \bibinfo{person}{Yanmin Zhu}, {and}
  \bibinfo{person}{Linghe Kong}.} \bibinfo{year}{2019}\natexlab{}.
\newblock \showarticletitle{An indirect eavesdropping attack of keystrokes on
  touch screen through acoustic sensing}.
\newblock \bibinfo{journal}{\emph{IEEE Transactions on Mobile Computing}}
  (\bibinfo{year}{2019}).
\newblock


\bibitem[\protect\citeauthoryear{Yun, Jeong, Kim, Kang, and Kim}{Yun
  et~al\mbox{.}}{2019}]%
        {yun2019graph}
\bibfield{author}{\bibinfo{person}{Seongjun Yun}, \bibinfo{person}{Minbyul
  Jeong}, \bibinfo{person}{Raehyun Kim}, \bibinfo{person}{Jaewoo Kang}, {and}
  \bibinfo{person}{Hyunwoo~J Kim}.} \bibinfo{year}{2019}\natexlab{}.
\newblock \showarticletitle{Graph transformer networks}.
\newblock \bibinfo{journal}{\emph{arXiv preprint arXiv:1911.06455}}
  (\bibinfo{year}{2019}).
\newblock


\bibitem[\protect\citeauthoryear{Zelinka, Das, Sikora, and
  {\v{S}}enke{\v{r}}{\'\i}k}{Zelinka et~al\mbox{.}}{2018}]%
        {zelinka2018swarm}
\bibfield{author}{\bibinfo{person}{Ivan Zelinka}, \bibinfo{person}{Swagatam
  Das}, \bibinfo{person}{Lubomir Sikora}, {and} \bibinfo{person}{Roman
  {\v{S}}enke{\v{r}}{\'\i}k}.} \bibinfo{year}{2018}\natexlab{}.
\newblock \showarticletitle{Swarm virus-Next-generation virus and antivirus
  paradigm?}
\newblock \bibinfo{journal}{\emph{Swarm and Evolutionary Computation}}
  \bibinfo{volume}{43} (\bibinfo{year}{2018}), \bibinfo{pages}{207--224}.
\newblock


\bibitem[\protect\citeauthoryear{Zeng and Church}{Zeng and Church}{2009}]%
        {zheng09}
\bibfield{author}{\bibinfo{person}{W. Zeng} {and} \bibinfo{person}{R.~L.
  Church}.} \bibinfo{year}{2009}\natexlab{}.
\newblock \showarticletitle{Finding Shortest Paths on Real Road Networks: The
  Case for A*}.
\newblock \bibinfo{journal}{\emph{Int. J. Geogr. Inf. Sci.}}
  \bibinfo{volume}{23}, \bibinfo{number}{4} (\bibinfo{date}{April}
  \bibinfo{year}{2009}), \bibinfo{pages}{531–543}.
\newblock
\showISSN{1365-8816}
\urldef\tempurl%
\url{https://doi.org/10.1080/13658810801949850}
\showDOI{\tempurl}


\bibitem[\protect\citeauthoryear{zerofox}{zerofox}{2020}]%
        {zerofoxo16:online}
\bibfield{author}{\bibinfo{person}{zerofox}.} \bibinfo{year}{2020}\natexlab{}.
\newblock \bibinfo{title}{zerofox-oss/SNAP\_R: A machine learning based social
  media pen\-testing tool}.
\newblock \bibinfo{howpublished}{{https://github.com/zerofox-oss/SNAP\_R}}.
\newblock
\newblock
\shownote{(Accessed on 10/21/2020).}


\bibitem[\protect\citeauthoryear{Zhang, Chen, Song, Boning, Dhillon, and
  Hsieh}{Zhang et~al\mbox{.}}{2019}]%
        {zhang19-iclr}
\bibfield{author}{\bibinfo{person}{Huan Zhang}, \bibinfo{person}{Hongge Chen},
  \bibinfo{person}{Zhao Song}, \bibinfo{person}{Duane~S. Boning},
  \bibinfo{person}{Inderjit~S. Dhillon}, {and} \bibinfo{person}{Cho{-}Jui
  Hsieh}.} \bibinfo{year}{2019}\natexlab{}.
\newblock \showarticletitle{The Limitations of Adversarial Training and the
  Blind-Spot Attack}. In \bibinfo{booktitle}{\emph{7th International Conference
  on Learning Representations, {ICLR} 2019, New Orleans, LA, USA, May 6-9,
  2019}}. \bibinfo{publisher}{OpenReview.net}.
\newblock
\urldef\tempurl%
\url{https://openreview.net/forum?id=HylTBhA5tQ}
\showURL{%
\tempurl}


\bibitem[\protect\citeauthoryear{Zhang and Chen}{Zhang and Chen}{2018}]%
        {zhang2018link}
\bibfield{author}{\bibinfo{person}{Muhan Zhang} {and} \bibinfo{person}{Yixin
  Chen}.} \bibinfo{year}{2018}\natexlab{}.
\newblock \showarticletitle{Link prediction based on graph neural networks}. In
  \bibinfo{booktitle}{\emph{Advances in Neural Information Processing
  Systems}}. \bibinfo{pages}{5165--5175}.
\newblock


\bibitem[\protect\citeauthoryear{Zhang, Lau, Liao, and Kwok}{Zhang
  et~al\mbox{.}}{2012}]%
        {zhang12-ICIS}
\bibfield{author}{\bibinfo{person}{W. Zhang}, \bibinfo{person}{R.Y.K. Lau},
  \bibinfo{person}{S.S.Y. Liao}, {and} \bibinfo{person}{R.C.-W Kwok}.}
  \bibinfo{year}{2012}\natexlab{}.
\newblock \showarticletitle{A probabilistic generative model for latent
  business networks mining}.
\newblock \bibinfo{journal}{\emph{International Conference on Information
  Systems, ICIS 2012}}  \bibinfo{volume}{2} (\bibinfo{date}{01}
  \bibinfo{year}{2012}), \bibinfo{pages}{1102--1118}.
\newblock


\bibitem[\protect\citeauthoryear{Zhang}{Zhang}{2018}]%
        {zhang2018analysis}
\bibfield{author}{\bibinfo{person}{X. Zhang}.} \bibinfo{year}{2018}\natexlab{}.
\newblock \bibinfo{title}{Analysis of New Agent Tesla Spyware Variant}.
\newblock
\newblock
\urldef\tempurl%
\url{{https://www.fortinet.com/blog/threat-research/analysis-of-new-agent-tesla-spyware-variant.html}}
\showURL{%
\tempurl}


\bibitem[\protect\citeauthoryear{{Zhang}, {Meng}, and {Pratama}}{{Zhang}
  et~al\mbox{.}}{2016}]%
        {zhang16-iecon}
\bibfield{author}{\bibinfo{person}{Y. {Zhang}}, \bibinfo{person}{J.~E. {Meng}},
  {and} \bibinfo{person}{M. {Pratama}}.} \bibinfo{year}{2016}\natexlab{}.
\newblock \showarticletitle{Extractive document summarization based on
  convolutional neural networks}. In \bibinfo{booktitle}{\emph{IECON 2016 -
  42nd Annual Conference of the IEEE Industrial Electronics Society}}.
  \bibinfo{pages}{918--922}.
\newblock
\urldef\tempurl%
\url{https://doi.org/10.1109/IECON.2016.7793761}
\showDOI{\tempurl}


\bibitem[\protect\citeauthoryear{Zhiyang, Wang, Li, Wu, Zhou, and
  Huang}{Zhiyang et~al\mbox{.}}{2019}]%
        {Zhiyang19-ieeeaccess}
\bibfield{author}{\bibinfo{person}{Fang Zhiyang}, \bibinfo{person}{Junfeng
  Wang}, \bibinfo{person}{Boya Li}, \bibinfo{person}{Siqi Wu},
  \bibinfo{person}{Yingjie Zhou}, {and} \bibinfo{person}{Haiying Huang}.}
  \bibinfo{year}{2019}\natexlab{}.
\newblock \showarticletitle{Evading Anti-Malware Engines With Deep
  Reinforcement Learning}.
\newblock \bibinfo{journal}{\emph{IEEE Access}}  \bibinfo{volume}{PP}
  (\bibinfo{date}{03} \bibinfo{year}{2019}), \bibinfo{pages}{1--1}.
\newblock
\urldef\tempurl%
\url{https://doi.org/10.1109/ACCESS.2019.2908033}
\showDOI{\tempurl}


\bibitem[\protect\citeauthoryear{Zhou, Elbadry, Gao, and Ye}{Zhou
  et~al\mbox{.}}{2017}]%
        {zhou2017batmapper}
\bibfield{author}{\bibinfo{person}{Bing Zhou}, \bibinfo{person}{Mohammed
  Elbadry}, \bibinfo{person}{Ruipeng Gao}, {and} \bibinfo{person}{Fan Ye}.}
  \bibinfo{year}{2017}\natexlab{}.
\newblock \showarticletitle{BatMapper: Acoustic sensing based indoor floor plan
  construction using smartphones}. In \bibinfo{booktitle}{\emph{Proceedings of
  the 15th Annual International Conference on Mobile Systems, Applications, and
  Services}}. \bibinfo{pages}{42--55}.
\newblock


\bibitem[\protect\citeauthoryear{{Zhu}, {Jing}, {You}, {Zhang}, and
  {Zhang}}{{Zhu} et~al\mbox{.}}{2018}]%
        {zhu18-tip}
\bibfield{author}{\bibinfo{person}{X. {Zhu}}, \bibinfo{person}{X. {Jing}},
  \bibinfo{person}{X. {You}}, \bibinfo{person}{X. {Zhang}}, {and}
  \bibinfo{person}{T. {Zhang}}.} \bibinfo{year}{2018}\natexlab{}.
\newblock \showarticletitle{Video-Based Person Re-Identification by
  Simultaneously Learning Intra-Video and Inter-Video Distance Metrics}.
\newblock \bibinfo{journal}{\emph{IEEE Transactions on Image Processing}}
  \bibinfo{volume}{27}, \bibinfo{number}{11} (\bibinfo{year}{2018}),
  \bibinfo{pages}{5683--5695}.
\newblock
\urldef\tempurl%
\url{https://doi.org/10.1109/TIP.2018.2861366}
\showDOI{\tempurl}


\end{thebibliography}

\end{document}